\documentclass{bmvc2k}

\usepackage{graphicx}
\usepackage{booktabs}
\usepackage{multirow}
\usepackage{colortbl}
\usepackage{arydshln}
\usepackage{overpic}
\usepackage{wrapfig}
\usepackage[export]{adjustbox}
\usepackage{caption}
\usepackage{amssymb}
\usepackage{amsmath}
\usepackage{bm}
\usepackage[dvipsnames]{xcolor}
\title{FlowIt: Global Matching via Hierarchical Transformers and Optimal Transport for Optical Flow}

\addauthor{Sadra Safadoust}{https://sadrasafa.github.io}{1}
\addauthor{Fabio Tosi}{https://fabiotosi92.github.io}{2}
\addauthor{Matteo Poggi}{https://mattpoggi.github.io}{2}
\addauthor{Fatma Güney}{https://mysite.ku.edu.tr/fguney}{1}

\addinstitution{
 Department of Computer Engineering \\
 and KUIS AI Center, 
 Koç University, \\
 Istanbul, Turkey
}
\addinstitution{
 Department of Computer Science and Engineering (DISI),\\
 University of Bologna, 
 Italy \\
}

\runninghead{S. Safadoust et al.}{FlowIt}

\def\eg{\emph{e.g}\bmvaOneDot}
\def\Eg{\emph{E.g}\bmvaOneDot}
\def\etal{\emph{et al}\bmvaOneDot}

\begin{document}

\maketitle
\begin{center} 
    \centering
    \vspace{-0.65cm}
    \begin{overpic}[width=0.9\linewidth]
    {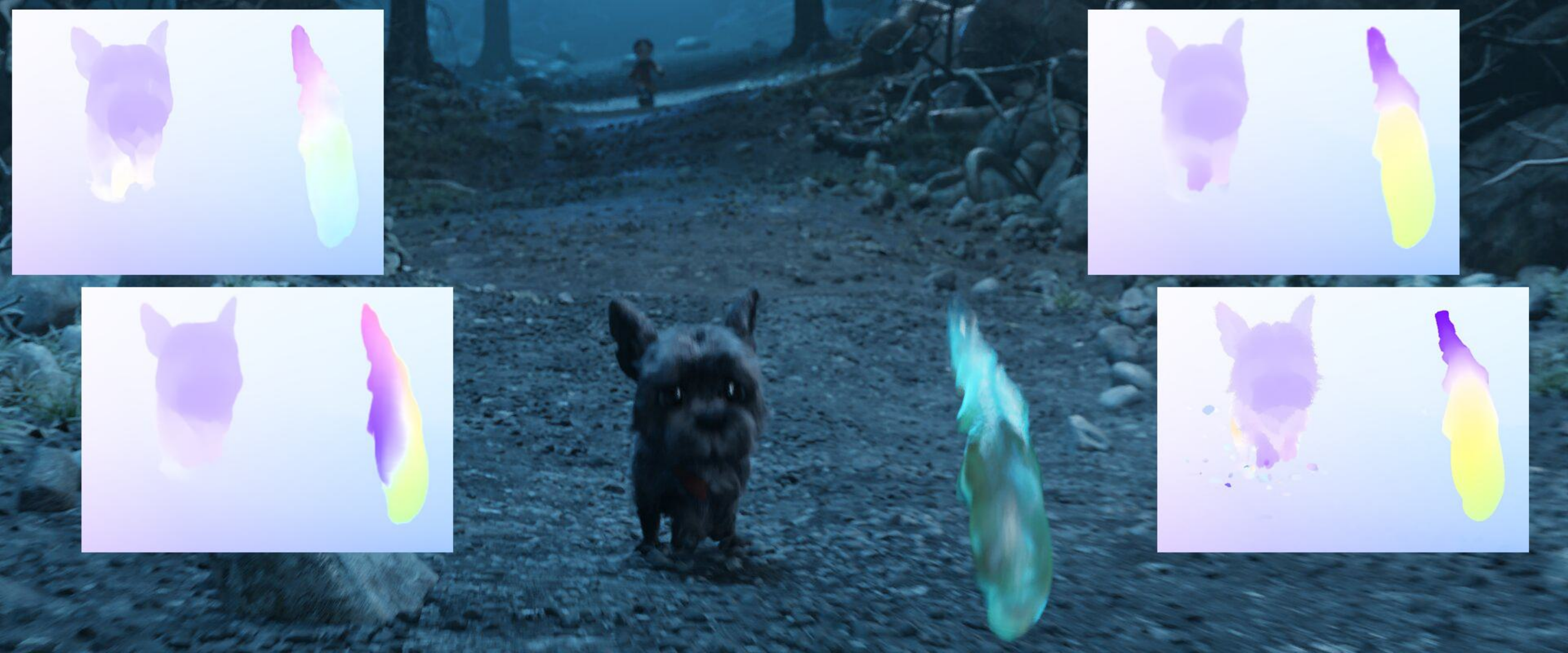}
    \put (2,25) {\textcolor{purple}{\textbf{\texttt{WAFT}}}}
    \put (70,25) {\small\textcolor{purple}{\textbf{\texttt{FlowIt}}}}
    \put (7,7) {\textcolor{purple}{\textbf{\texttt{SEA-RAFT}}}}
    \put (75,7) {\small\textcolor{purple}{\textbf{\texttt{Ground Truth}}}}
    \end{overpic}\vspace{0.18cm}
    \captionof{figure}{\textbf{Qualitative comparison between WAFT \cite{wang2025waft}, SEA-RAFT \cite{wang2024sea}, and FlowIt.} Every model is trained on multiple synthetic datasets and evaluated zero-shot on the Spring training set. Our FlowIt can generalize much better to this unseen domain. }
    \label{fig:teaser}
\end{center}

%
%
\newcommand{\red}[1]{{\color{red}#1}}
\newcommand{\todo}[1]{{\color{red}#1}}
\newcommand{\TODO}[1]{\textbf{\color{red}[TODO: #1]}}

\newcommand{\Perp}{\perp\!\!\! \perp}
\newcommand{\bK}{\mathbf{K}}
\newcommand{\bX}{\mathbf{X}}
\newcommand{\bY}{\mathbf{Y}}
\newcommand{\bk}{\mathbf{k}}
\newcommand{\bx}{\mathbf{x}}
\newcommand{\by}{\mathbf{y}}
\newcommand{\bhy}{\hat{\mathbf{y}}}
\newcommand{\bty}{\tilde{\mathbf{y}}}
\newcommand{\bG}{\mathbf{G}}
\newcommand{\bI}{\mathbf{I}}
\newcommand{\bg}{\mathbf{g}}
\newcommand{\bS}{\mathbf{S}}
\newcommand{\bs}{\mathbf{s}}
\newcommand{\bM}{\mathbf{M}}
\newcommand{\bw}{\mathbf{w}}
\newcommand{\eye}{\mathbf{I}}
\newcommand{\bU}{\mathbf{U}}
\newcommand{\bV}{\mathbf{V}}
\newcommand{\bW}{\mathbf{W}}
\newcommand{\bn}{\mathbf{n}}
\newcommand{\bv}{\mathbf{v}}
\newcommand{\bwv}{\mathbf{wv}}
\newcommand{\bq}{\mathbf{q}}
\newcommand{\bR}{\mathbf{R}}
\newcommand{\bi}{\mathbf{i}}
\newcommand{\bj}{\mathbf{j}}
\newcommand{\bp}{\mathbf{p}}
\newcommand{\bt}{\mathbf{t}}
\newcommand{\bJ}{\mathbf{J}}
\newcommand{\bu}{\mathbf{u}}
\newcommand{\bB}{\mathbf{B}}
\newcommand{\bD}{\mathbf{D}}
\newcommand{\bz}{\mathbf{z}}
\newcommand{\bP}{\mathbf{P}}
\newcommand{\bC}{\mathbf{C}}
\newcommand{\bA}{\mathbf{A}}
\newcommand{\bZ}{\mathbf{Z}}
\newcommand{\bff}{\mathbf{f}}
\newcommand{\bF}{\mathbf{F}}
\newcommand{\bo}{\mathbf{o}}
\newcommand{\bO}{\mathbf{O}}
\newcommand{\bc}{\mathbf{c}}
\newcommand{\bT}{\mathbf{T}}
\newcommand{\bQ}{\mathbf{Q}}
\newcommand{\bL}{\mathbf{L}}
\newcommand{\bl}{\mathbf{l}}
\newcommand{\ba}{\mathbf{a}}
\newcommand{\bE}{\mathbf{E}}
\newcommand{\bH}{\mathbf{H}}
\newcommand{\bd}{\mathbf{d}}
\newcommand{\br}{\mathbf{r}}
\newcommand{\be}{\mathbf{e}}
\newcommand{\bb}{\mathbf{b}}
\newcommand{\bh}{\mathbf{h}}
\newcommand{\bhh}{\hat{\mathbf{h}}}
\newcommand{\btheta}{\boldsymbol{\theta}}
\newcommand{\bTheta}{\boldsymbol{\Theta}}
\newcommand{\bpi}{\boldsymbol{\pi}}
\newcommand{\bphi}{\boldsymbol{\phi}}
\newcommand{\bpsi}{\boldsymbol{\psi}}
\newcommand{\bPhi}{\boldsymbol{\Phi}}
\newcommand{\bmu}{\boldsymbol{\mu}}
\newcommand{\bsigma}{\boldsymbol{\sigma}}
\newcommand{\bSigma}{\boldsymbol{\Sigma}}
\newcommand{\bGamma}{\boldsymbol{\Gamma}}
\newcommand{\bbeta}{\boldsymbol{\beta}}
\newcommand{\bomega}{\boldsymbol{\omega}}
\newcommand{\blambda}{\boldsymbol{\lambda}}
\newcommand{\bLambda}{\boldsymbol{\Lambda}}
\newcommand{\bkappa}{\boldsymbol{\kappa}}
\newcommand{\btau}{\boldsymbol{\tau}}
\newcommand{\balpha}{\boldsymbol{\alpha}}
\newcommand{\nR}{\mathbb{R}}
\newcommand{\nN}{\mathbb{N}}
\newcommand{\nL}{\mathbb{L}}
\newcommand{\nE}{\mathbb{E}}
\newcommand{\cN}{\mathcal{N}}
\newcommand{\cM}{\mathcal{M}}
\newcommand{\cR}{\mathcal{R}}
\newcommand{\cB}{\mathcal{B}}
\newcommand{\cL}{\mathcal{L}}
\newcommand{\cH}{\mathcal{H}}
\newcommand{\cS}{\mathcal{S}}
\newcommand{\cT}{\mathcal{T}}
\newcommand{\cO}{\mathcal{O}}
\newcommand{\cC}{\mathcal{C}}
\newcommand{\cP}{\mathcal{P}}
\newcommand{\cE}{\mathcal{E}}
\newcommand{\cI}{\mathcal{I}}
\newcommand{\cF}{\mathcal{F}}
\newcommand{\cK}{\mathcal{K}}
\newcommand{\cW}{\mathcal{W}}
\newcommand{\cY}{\mathcal{Y}}
\newcommand{\cX}{\mathcal{X}}
\newcommand{\cZ}{\mathcal{Z}}
\def\bgamma{\boldsymbol\gamma}

\newcommand{\specialcell}[2][c]{%
  \begin{tabular}[#1]{@{}c@{}}#2\end{tabular}}

\newcommand{\figref}[1]{\Fig~\ref{#1}}
\newcommand{\secref}[1]{Section~\ref{#1}}
\newcommand{\algref}[1]{Algorithm~\ref{#1}}
\newcommand{\eqnref}[1]{Eq.~\eqref{#1}}
\newcommand{\tabref}[1]{Table~\ref{#1}}

\newcommand{\rulesep}{\unskip\ \vrule\ }



\newcommand{\KLD}[2]{D_{\mathrm{KL}} \Big(#1 \mid\mid #2 \Big)}

\renewcommand{\b}{\ensuremath{\mathbf}}

\def\mc{\mathcal}
\def\mb{\mathbf}

\newcommand{\T}{^{\raisemath{-1pt}{\mathsf{T}}}}

\makeatletter
\DeclareRobustCommand\onedot{\futurelet\@let@token\@onedot}
\def\@onedot{\ifx\@let@token.\else.\null\fi\xspace}
\def\eg{e.g\onedot} \def\Eg{E.g\onedot}
\def\ie{i.e\onedot} \def\Ie{I.e\onedot}
\def\cf{cf\onedot} \def\Cf{Cf\onedot}
\def\etc{etc\onedot} \def\vs{vs\onedot}
\def\wrt{wrt\onedot}
\def\dof{d.o.f\onedot}
\def\etal{et~al\onedot} \def\iid{i.i.d\onedot}
\def\Fig{Fig\onedot} \def\Eqn{Eqn\onedot} \def\Sec{Sec\onedot} \def\Alg{Alg\onedot}
\makeatother

\newcommand{\xdownarrow}[1]{%
  {\left\downarrow\vbox to #1{}\right.\kern-\nulldelimiterspace}
}

\newcommand{\xuparrow}[1]{%
  {\left\uparrow\vbox to #1{}\right.\kern-\nulldelimiterspace}
}


\newcommand*\rot{\rotatebox{90}}
\newcommand{\boldparagraph}[1]{\vspace{0.15cm}\noindent{\bf #1.} }
\newcommand{\boldquestion}[1]{\vspace{0.2cm}\noindent{\bf #1} }

\newcommand{\matt}[1]{ \noindent {\color{orange} {#1}} } 
\newcommand{\sh}[1]{ \noindent {\color{blue} {#1}} } 
\newcommand{\ftm}[1]{ \noindent {\color{cyan} {#1}}}

\newcommand{\cmark}{\checkmark}%
\newcommand{\xmark}{\ding{53}}%


\definecolor{First}{HTML}{BDE6CD}
\definecolor{Second}{HTML}{E2EEBC}
\definecolor{Third}{HTML}{FFF8C5}


\newcommand{\fst}[1]{\bf\cellcolor{First}#1}
\newcommand{\snd}[1]{\cellcolor{Second}#1}
\newcommand{\trd}[1]{\cellcolor{Third}#1}

\newcommand{\matteo}[1]{{\color{orange}#1}}
\newcommand{\fabio}[1]{{\color{red}#1}}
\newcommand{\sadra}[1]{{\color{blue}#1}}

\newcommand{\net}{FlowIt}
\def\multirowcenter{-0.5\dimexpr \aboverulesep + \belowrulesep + \cmidrulewidth}
\newsavebox\CBox
\def\textBF#1{\sbox\CBox{#1}\resizebox{\wd\CBox}{\ht\CBox}{\textbf{#1}}}

\begin{abstract}
We present FlowIt, a novel architecture for optical flow estimation 
that combines global matching with confidence and occlusion-guided refinement. At its core, FlowIt leverages a hierarchical transformer architecture that captures extensive global context, enabling the model to effectively model long-range correspondences. To overcome the limitations of localized matching, we formulate the flow initialization as an optimal transport problem. This formulation yields a highly robust initial flow field, alongside explicitly derived occlusion and confidence maps. These cues are then seamlessly integrated into a guided refinement stage, where the network actively propagates reliable motion estimates from high-confidence regions into ambiguous, low-confidence areas. Extensive experiments across the Sintel, KITTI, Spring, and LayeredFlow datasets validate the effectiveness of our approach. FlowIt achieves state-of-the-art results on the competitive Sintel benchmark and establishes new state-of-the-art cross-dataset zero-shot generalization performance on Sintel, Spring, and LayeredFlow, while also delivering competitive performance on both the KITTI benchmark and KITTI zero-shot generalization settings.
\end{abstract}

\section{Introduction}
Optical flow estimation, the task of determining the 2D motion of pixels between consecutive frames, is a cornerstone of dynamic scene understanding, crucial for many downstream 
applications such as video action recognition~\cite{sun2018optical, piergiovanni2019representation, zhao2020improved}, autonomous driving~\cite{kitti, menze2015object, janai2020computer}, 
frame interpolation~\cite{liu2020video, zhao2020improved, huang2022real, xu2019quadratic}, 
and video in-painting~\cite{xu2019deep, gao2020flow, kim2019deep}. In recent years, the landscape of deep learning-based optical flow has been heavily dominated by iterative local search architectures \cite{teed2020raft, jiang2021learning2, wang2024sea, morimitsu2025dpflow}.
While these recurrent refinement models achieve impressive scores on constrained benchmarks by iteratively updating a flow field using local correlation lookups, their fundamental reliance on local search mechanisms inherently limits their ability to enforce long-range global consistency. This limitation often leads to sub-optimal generalization, particularly in scenarios involving large displacements, severe occlusions, or textureless regions.

Conversely, global matching architectures~\cite{xu2022gmflow, zhao2022global, jiang2021learning} provide a theoretically more robust alternative by explicitly capturing long-range dependencies and utilizing comprehensive context to resolve local ambiguities. However, the prohibitive computational and memory costs associated with dense 4D correlation volumes restrict these models to heavily downsampled feature maps, often at $\frac{1}{8}$ of the original resolution. Compounding this issue, the subsequent iterative refinement modules typically rely on localized search, leaving them susceptible to local minima and limiting their ability to correct large-displacement errors.

In response to these challenges, we propose \textbf{\net{}}, a new global matching network for optical flow estimation. 
The backbone of our framework relies on a hierarchical transformer design that facilitates rich context sharing, both horizontally within individual resolution stages and vertically across multiple scales. This feature exchange ensures that the model retains global coherence and effectively maps distant correspondences, critically enabling operations at a higher spatial scale. Drawing inspiration from recent advances in stereo matching that demonstrate the efficacy of Optimal Transport for disparity initialization~\cite{min2024confidence, min2025s2m2}, we formulate our initial flow extraction from the cost volume as an Optimal Transport problem~\cite{cuturi2013sinkhorn}, which yields a highly stable global flow prior, as well as confidence and occlusion maps, for subsequent refinement. 
Beyond architectural enhancements, we introduce a refinement module guided by confidence and occlusion maps.
These auxiliary predictions are leveraged to actively steer the refinement process, encouraging reliable motion cues to propagate from regions of high certainty into ambiguous, low-confidence zones. Supervised by a dedicated loss formulation, this guided propagation significantly improves the accuracy of the predicted flow field, particularly in visually challenging areas. We demonstrate the critical advantage of this approach on challenging scenes, where our model exhibits robust performance and substantial accuracy gains even in the presence of challenging visual phenomena such as transparent and reflective surfaces~\cite{wen2024layeredflow}.

We comprehensively validate the performance of \net{} across standard optical flow datasets, namely Sintel~\cite{sintel}, KITTI~\cite{kitti}, Spring~\cite{mehl2023spring}, and LayeredFlow~\cite{wen2024layeredflow}. Our model achieves state-of-the-art results on established benchmarks, while also demonstrating very competitive zero-shot generalization capabilities. %

In summary, our contributions are as follows:
\begin{itemize}
\item We introduce \net{}, a novel transformer-based global matching architecture for optical flow estimation. By leveraging a hierarchical transformer design and an Optimal Transport-based flow initialization, the proposed framework effectively captures long-range correspondences while remaining robust across challenging motion scenarios.
To accommodate diverse computational constraints, we offer multiple model variants that optimize the trade-off between parameter size and performance.
    \item We propose a specialized loss formulation that significantly enhances flow accuracy. Coupled with the joint estimation of confidence and occlusion maps, this approach guarantees the high reliability of the final motion predictions.
\item We establish new state-of-the-art results on the Sintel benchmark, as well as zero-shot generalization performance across the Sintel, Spring, and LayeredFlow datasets, while simultaneously demonstrating competitive performance on the KITTI benchmark.
\end{itemize}

Our project webpage is available at: \url{https://kuis-ai.github.io/FlowIt/}.
\section{Related Work}

The evolution of optical flow estimation has transitioned from traditional energy minimization frameworks \cite{horn1981determining, black1993framework, zach2007duality} to the current dominance of deep neural networks. Among classical approaches, some introduced the coarse-to-fine spatial pyramid \cite{anandan1989computational} to address large displacements where gradient-based optimization typically fails, which estimates motion across multiple resolutions. This is coupled with a warping mechanism \cite{brox2004high} that iteratively aligns the source image to the target to resolve residual motion.
Additionally, the use of discrete cost volumes storing matching costs across a search window provided robust global optimization, often enhanced by high-dimensional feature descriptors \cite{liu2010sift, besse2014pmbp}. 

These design choices now form the structural blueprint for modern deep learning architectures. The shift to end-to-end deep models started with FlowNet \cite{chairs}, introducing a correlation layer used to compute a cost volume, followed by FlowNet 2 \cite{ilg2017flownet}, which added multiple refinement stages making use of warping.
Similarly, another pillar of traditional methods, i.e., coarse-to-fine spatial pyramid, has been adopted by SpyNet \cite{ranjan2017optical}, while PWC-Net \cite{sun2018pwc} combined spatial pyramid with warping and cost volumes. Further advancements still built on these three pillars to optimize volumetric efficiency \cite{yang2019volumetric}, achieve lightweight footprints \cite{hui18liteflownet, hui20liteflownet2, hui20liteflownet3}, while also progressively improving the training strategies to attain optimal performance from existing architectures \cite{sun2019models}.

In the twenties, a paradigm shift in network design occurred with RAFT \cite{teed2020raft}, casting flow estimation into an optimization-inspired iterative process guided by an all-pairs correlation volume, followed by further works \cite{sun2022disentangling} that assessed the separate contributions of architecture and training. This approach has become the cornerstone for subsequent architectures \cite{morimitsu2024recurrent, deng2023explicit}, some of which focused on high-resolution processing \cite{zheng2022dip, jahedi2024ms, jahedi2024ccmr}, or on simplifying the iterative process to reduce complexity, at first with SEA-RAFT \cite{wang2024sea} running fewer iterations and, lately, with WAFT \cite{wang2025waft} by replacing the need for a cost volume with a warping-based iterative refinement stage. 

On a parallel track, the increasing popularity of vision transformers led to the development of transformer-based architectures for optical flow, such as  FlowFormer~\cite{huang2022flowformer}, FlowFormer++~\cite{shi2023flowformer++}, CRAFT~\cite{sui2022craft}, GMFlowNet~\cite{zhao2022global}, and efficient high-resolution approaches~\cite{leroy2023win} leveraging various forms of attention for global context. WAFT~\cite{wang2025waft} exploits a vision transformer to guide the warping-based iterative refinement.
Built on these two new design pillars, more architectures focused on specific challenges typical of optical flow, such as handling occlusions~\cite{jiang2021learning, luo2022learning, sun2022skflow}, exploiting the affinities with other matching tasks, such as stereo~\cite{xu2022gmflow, xu2023unifying}, or matching pretraining \cite{dong2023rethinking}, as well as the power of pre-trained depth foundation models \cite{Poggi_2025_ICCV}.  On a side track, the use of diffusion models~\cite{saxena2023surprising} also unveiled surprising effectiveness even in the absence of task-specific designs.

Finally, other works explored directions orthogonal to the more classical, supervised optical flow setting, considering unsupervised approaches~\cite{jonschkowski2020matters, liu2020learning} as a possible solution to data scarcity, sometimes in synergy with stereo matching~\cite{jiao2021effiscene, liu2020flow2stereo}, or developing alternatives for automatically generating training data~\cite{sun2021autoflow, han2022realflow,aleotti2021learning,jeong2023distractflow}. Beyond the two-frame setting, multi-frame approaches have also been explored,
by focusing on consecutive tri-frame flow estimations \cite{shi2023videoflow}, introducing a reusable memory buffer \cite{dong2024memflow} or reduced correlation volumes \cite{Bargatin_2025_ICCV}, or in an auto-regressive fashion \cite{liu2026arflow}, yet at the cost of higher computational complexity.
\section{Method}
\label{sec:method}

\begin{figure*}[t]
    \centering
    \begin{overpic}[clip,trim=0cm 0cm 0cm 0cm,width=1\linewidth]{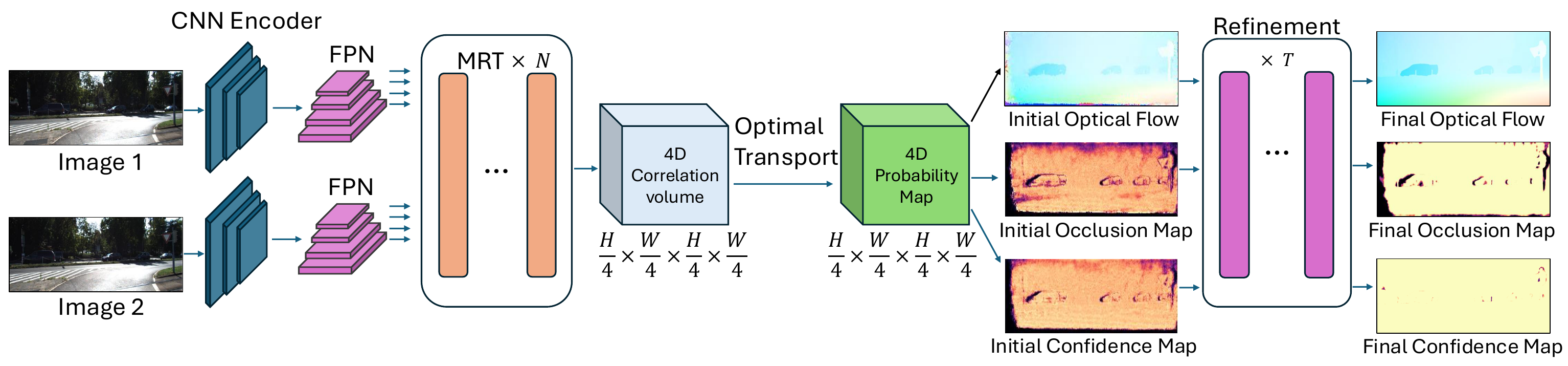}
    \end{overpic}
    \vspace{0.01cm}
    \caption{\textbf{Architecture Overview.} \net{} extracts multi-scale features from images using a CNN encoder followed by a Feature Pyramid Network (FPN). These features are processed with one or more Multi-Resolution Transformer (MRT) blocks. A 4D correlation volume is constructed using the $\frac{1}{4}$ resolution features, and optimal transport is applied to produce a 4D probability map. Initial flow, occlusion, and confidence maps are derived using the probability map. These predictions are refined through three refinement iterations to obtain the final outputs.}
    \label{fig:method}
\end{figure*}

Given a pair of input images $I_1, I_2 \in \mathbb{R}^{H \times W \times 3}$, our objective is to estimate a dense optical flow field $\mathbf{F} \in \mathbb{R}^{H \times W \times 2}$ that represents the pixel-wise displacement between the two frames. 
Our approach is structured into four sequential stages, as illustrated in \figref{fig:method}. In the following, we describe each stage in detail.

\subsection{Feature Extraction}
\label{sec:feature_extraction}

Recent state-of-the-art methods~\cite{wang2025waft} employ large-scale Vision Transformers for feature extraction. While effective, such models are computationally expensive for dense optical flow estimation, particularly at high resolutions. %

To achieve a more efficient design, we adopt a CNN-based backbone to extract low-level visual representations. These features are then processed by a Feature Pyramid Network (FPN) to construct a multi-scale feature pyramid. Formally, for each image $I_i$ in the input pair, where $i \in \{1,2\}$, we define the feature extraction pipeline  as $\mathcal{F}(I_i) = \{\mathbf{f}_i^{\,4}, \mathbf{f}_i^{\,8}, \mathbf{f}_i^{\,16}, \mathbf{f}_i^{\,32} \}$ where $\mathbf{f}_i^{\,s}$ represents the feature maps of image $i$ at $\frac{1}{s}$ of the original resolution.
 The resulting multi-scale features are subsequently fed into the Multi-Resolution Transformer (MRT) for further contextual refinement.

The MRT follows a design similar to \cite{min2025s2m2}, processing all pyramid levels in parallel while enabling cross-scale interaction through an Adaptive Gated Fusion mechanism \cite{min2025s2m2}, which promotes effective information exchange across resolutions. Ultimately, the MRT aggregates these hierarchical inputs into a unified feature representation at $\frac{1}{4}$ resolution, denoted as $\mathcal{G}(\mathbf{f}_i^{\,4}, \mathbf{f}_i^{\,8}, \mathbf{f}_i^{\,16}, \mathbf{f}_i^{\,32})= \mathbf{g}_i$,  where $\mathbf{g}_{i}$ represents the final, contextually enriched feature map for image $I_i$.%

\subsection{Initial Matching}
\label{sec:flow_init}

\begin{figure*}[t]
    \centering
    \begin{tabular}{ccc}
    \begin{overpic}[clip,trim=1.7cm 1cm 0.5cm 2cm,width=0.3\linewidth]{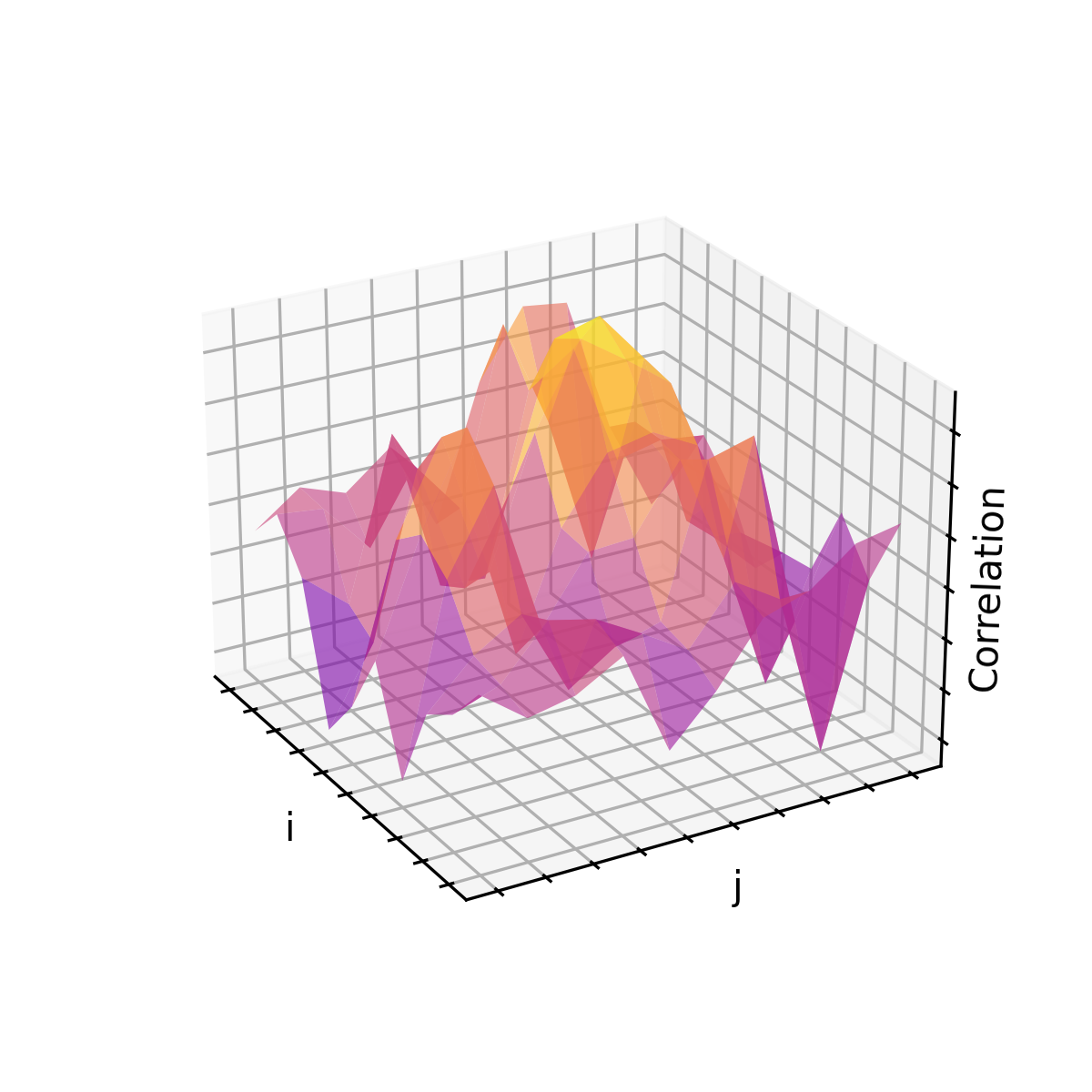}
    \end{overpic} &
    \begin{overpic}[clip,trim=1.7cm 1cm 0.5cm 2cm,width=0.3\linewidth]{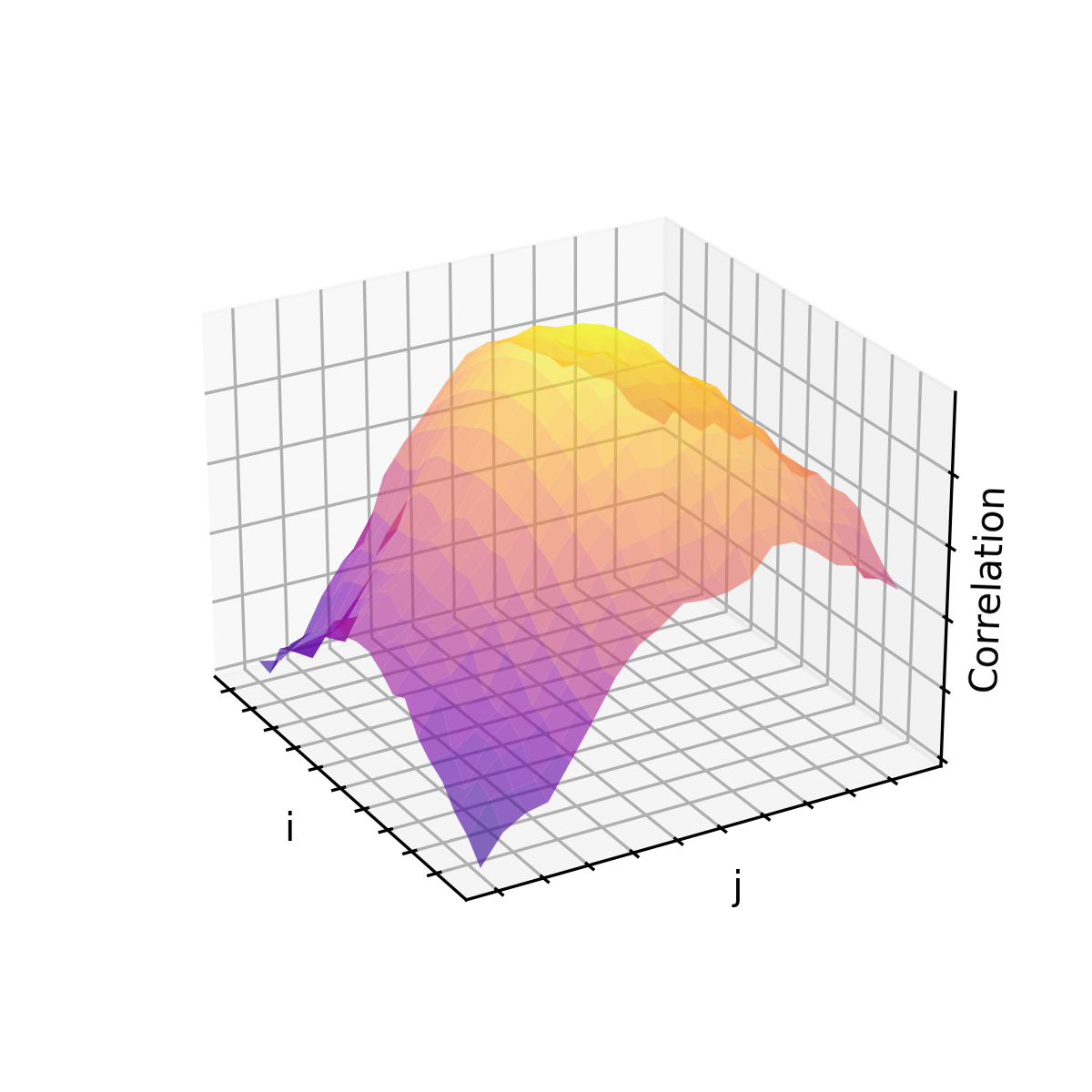}
    \end{overpic} & 
    \begin{overpic}[clip,trim=1.7cm 1cm 0.5cm 2cm,width=0.3\linewidth]{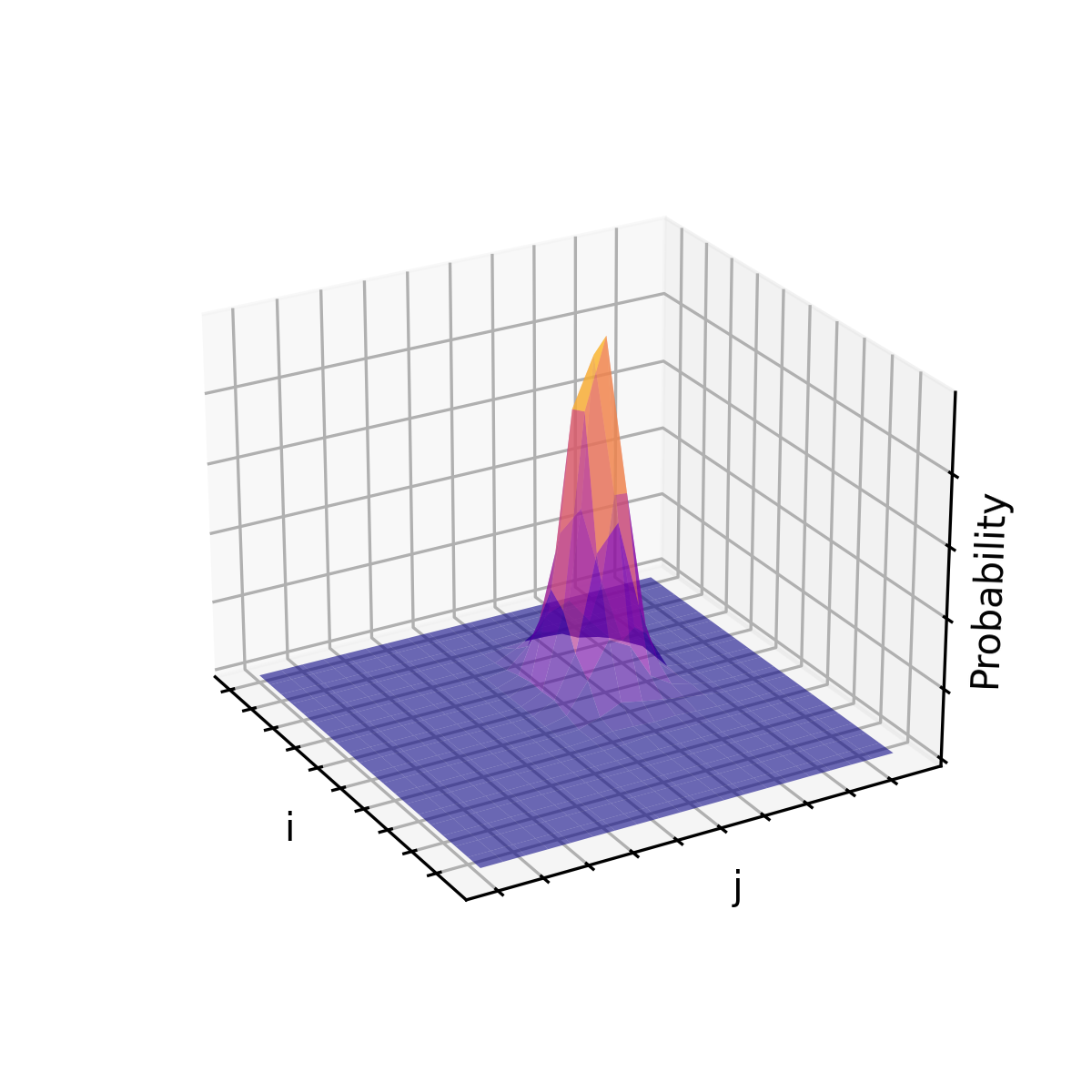}
    \end{overpic} \\ [-0.3cm]
    \textbf{(a)} & \textbf{(b)} & \textbf{(c)}
    \end{tabular}
    \vspace{+0.2cm}
        \caption{\textbf{Correlation and probability visualizations for a single pixel.} \textbf{(a)} and \textbf{(b)} show the correlation values in the pixel's neighborhood for SEA-RAFT and \net{}, respectively. \textbf{(c)} shows the corresponding probability map obtained by applying optimal transport to \net{}'s correlation volume.}

    \label{fig:manifold}
\end{figure*}

To obtain the initial flow estimate,  we utilize the $\frac{1}{4}$-resolution feature maps, $\mathbf{g}_1$ and $\mathbf{g}_2$, to construct an all-pairs correlation volume
$
C \in \mathbb{R}^{\frac{H}{4} \times \frac{W}{4} \times \frac{H}{4} \times \frac{W}{4}}.
$
We cast correspondence estimation as a global matching problem over this correlation volume. %
Specifically, we employ entropy-regularized optimal transport and solve it using the Sinkhorn algorithm~\cite{cuturi2013sinkhorn}. 
This formulation enforces soft mutual-consistency constraints between the two images and produces a globally consistent transport plan that is robust to matching ambiguities, where a locally optimal match from one image may not be optimal from the other. We adopt the standard dustbin strategy~\cite{min2024confidence, sarlin2020superglue} to account for occlusions and unmatched regions. 
The optimal transport solver produces a probabilistic matching tensor
$
\mathbf{P} \in \mathbb{R}^{\frac{H}{4} \times \frac{W}{4} \times \frac{H}{4} \times \frac{W}{4}},
$
where $\mathbf{P}(u, v, u', v')$ denotes the probability of matching pixel $(u, v)$ in $\mathbf{g}_1$ to pixel $(u', v')$ in $\mathbf{g}_2$. Figure~\ref{fig:manifold} illustrates examples of correlation and probability volumes. Compared to SEA-RAFT~\cite{wang2024sea}, the correlation volume is notably smoother, highlighting the effect of global matching. 

The initial flow field is computed using a localized soft expectation around the most confident match. For a source pixel $(u, v)$, let $(u^*, v^*)$ denote the target coordinate with the highest matching probability. We define a local window $\mathcal{W}$ of radius $r$ centered at this peak:
\begin{equation}
\mathcal{W} = \{ (u^* + i, v^* + j) \mid -r \le i \le r, \; -r \le j \le r \}.
\end{equation}

We then compute the  target coordinate $(\hat{u}, \hat{v})$ as the probability-weighted centroid within this window:
\begin{equation}
\hat{u} =
\frac{\sum_{(u', v') \in \mathcal{W}} \mathbf{P}(u, v, u', v') \, u'}
{\sum_{(u', v') \in \mathcal{W}} \mathbf{P}(u, v, u', v') + \epsilon},
\quad
\hat{v} =
\frac{\sum_{(u', v') \in \mathcal{W}} \mathbf{P}(u, v, u', v') \, v'}
{\sum_{(u', v') \in \mathcal{W}} \mathbf{P}(u, v, u', v') + \epsilon}.
\end{equation}
where $\epsilon$ is a small positive number to prevent division by zero.
Finally, the initial optical flow $\mathbf{F}_0$ at location $(u, v)$ is defined as the displacement between the expected target coordinate and the source coordinate:
\begin{equation}
\mathbf{F}_0({u,v}) =
\begin{bmatrix}
\hat{u} - u \\
\hat{v} - v
\end{bmatrix}.
\end{equation}

In addition to the initial flow, we derive an initial occlusion map and initial confidence map from $\mathbf{P}$. The confidence of the flow prediction for a given pixel is calculated as the sum of the matching probabilities within the local window surrounding the peak. Specifically, the initial confidence map $\bGamma_0$ is derived as:
\begin{equation}
    \bGamma_0({u,v}) = \sum_{(u',v') \in \mathcal{W}}\mathbf{P}(u,v,u',v')
\end{equation}
where high matching probabilities in the window lead to a high confidence.

The initial occlusion map $\mathbf{O}_0$ is derived as  marginal sum over all target candidates:
\begin{equation}
    \mathbf{O}_0({u,v}) = \sum_{u'=1}^{\frac{W}{4}}\sum_{v'=1}^{\frac{H}{4}}\mathbf{P}(u,v,u',v')
\end{equation}
This quantity measures the total probability mass assigned to valid correspondences for pixel 
$(u,v)$. When the pixel has a clear match in the second image, the transport plan concentrates significant probability mass on the corresponding locations, resulting in a high marginal sum. Conversely, if the pixel is occluded or unmatched, most of the mass is assigned to the dustbin~\cite{sarlin2020superglue}, producing a lower marginal value.

Together, the initial flow, occlusion, and confidence maps form a robust initialization that will be subsequently refined.

\subsection{Refinement}
\label{sec:refine}
The flow field undergoes iterative local refinements, similar to RAFT~\cite{teed2020raft}, to further sharpen the estimates. During each iteration $t$, a refinement network predicts residual updates for the optical flow $\Delta \mathbf{F}^{\,t}$, the confidence map $\Delta \bGamma^{\,t}$, and the occlusion mask $\Delta \mathbf{O}^{\,t}$. The refinements are conditioned on the predicted flow, occlusion, confidence, and local feature correlations. To guarantee numerical stability and enforce the valid $[0, 1]$ bounds for occlusion and confidence maps, their respective residual updates are accumulated in logit space:
\begin{align}\mathbf{F}^{\,t+1} &= \mathbf{F}^{\,t} + \Delta \mathbf{F}^{\,t}\\
\bGamma^{\,t+1} &= \sigma\left(\sigma^{-1}\left(\bGamma^{\,t}\right) + \Delta \bGamma^{\,t}\right) \\
\mathbf{O}^{\,t+1} &= \sigma\left(\sigma^{-1}\left(\mathbf{O}^{\,t}\right) + \Delta \mathbf{O}^{\,t}\right)
\end{align}
where $\sigma$ denotes the sigmoid function and $\sigma^{-1}$ represents its inverse (the logit function).
In practice, we estimate the optical flow residuals $\Delta \mathbf{F}^{\,t}$ independently for the horizontal ($u$) and vertical ($v$) components with two separate networks. As demonstrated in  Sec. \ref{sec:ablation}, this axis-wise refinement strategy is critical for achieving high performance. %

Up to this point, all predicted flow, occlusion, and confidence maps are computed at $\frac{1}{4}$ resolution. Following prior work~\cite{teed2020raft, min2025s2m2}, we employ convex upsampling to obtain full-resolution predictions. Specifically, each high-resolution flow vector is expressed as a convex combination of its corresponding $3 \times 3$ neighborhood in the coarse-resolution grid, where the combination weights are learned feature maps. The same upsampling strategy is applied to the occlusion and confidence maps. 

\subsection{Supervision}
\label{sec:loss}
%

%

The predicted initial flow $\mathbf{F}^{\,0}$, occlusion map $\mathbf{O}^{\,0}$, and confidence map $\bGamma^{\,0}$, along with their refinements $\{\mathbf{F}^{\,t}, \mathbf{O}^{\,t}, \bGamma^{\,t}\}_{t=1}^{T}$, where $t$ denotes the refinement iteration, are supervised as follows. 

\boldparagraph{Occlusion Loss}
The occlusion maps are trained using an $\cL_1$ loss over all $N$ pixels:
\begin{equation}
\mathcal{L}_{\mathbf{O}} =
\frac{1}{N} \sum_{t=0}^{T}
\left|\left|
\mathbf{O}^{\,t}- \mathbf{O}_{gt}^{\,t}
\right|\right|_1,
\end{equation}
The ground-truth occlusion map $\mathbf{O}_{gt}$ is obtained via a forward–backward consistency check: a pixel is considered non-occluded if the forward–backward warping error is below 2 pixels.

\boldparagraph{Confidence Loss}
For the confidence map, the initial prediction is supervised using an $\cL_1$ loss over non-occluded pixels only, while subsequent refinements are supervised over the full image:
\begin{equation}
\mathcal{L}_{\bGamma} =
\frac{1}{|\mathbf{O}_{gt}|}
\left|\left|
\mathbf{O}_{gt} \odot (\bGamma^{\,0} - \bGamma_{gt}^{\,0})
\right|\right|_1
+
\frac{1}{N} \sum_{t=1}^{T}
\left|\left|
\bGamma^{\,t} - \bGamma_{gt}^{\,t}
\right|\right|_1.
\end{equation}
 Note that $\mathbf{O}_{gt}$ equals 1 for non-occluded pixels and 0 otherwise, and $|\mathbf{O}_{gt}|$ denotes the number of non-occluded pixels. The ground-truth confidence map $\bGamma_{gt}$ is defined as an indicator function selecting pixels whose endpoint error is below 4 pixels at full resolution (equivalent to 1 pixel at $\frac{1}{4}$ resolution).

\boldparagraph{Optical Flow Loss}
For the flow, the initial estimate is supervised using a Smooth-$\cL_1$ loss restricted to non-occluded pixels, while refinement stages employ a standard $\cL_1$ loss across all pixels:
\begin{equation}
\cL_\bF =
\frac{1}{|\mathbf{O}_{gt}|}
\text{Smooth-}L_1
\left(
\mathbf{O}_{gt} \odot (\mathbf{F}^{\,0}- \mathbf{F}_{gt})
\right)
+
\frac{1}{N} \sum_{t=1}^{T}
\left|\left|
\mathbf{F}^{\,t} - \mathbf{F}_{gt}
\right|\right|_1.
\end{equation}

%
The overall training objective is defined as a weighted combination of the flow, confidence, and occlusion losses:
\begin{equation}
\mathcal{L}
=
\lambda_{\mathbf{F}} \mathcal{L}_{\mathbf{F}}
+
\lambda_{\bGamma} \mathcal{L}_{\bGamma}
+
\lambda_{\mathbf{O}} \mathcal{L}_{\mathbf{O}}.
\end{equation}
In all experiments, we set the hyperparameters to
$\lambda_{\mathbf{F}} = 1$, and
$\lambda_{\bGamma} =
\lambda_{\mathbf{O}} = 0.1$.
\section{Experiments}
\label{sec:exp}
We evaluate \net{}  on Sintel \cite{sintel}, KITTI \cite{kitti}, Spring \cite{mehl2023spring}, and LayeredFlow \cite{wen2024layeredflow}. Following prior work~\cite{wang2025waft, wang2024sea}, TartanAir \cite{wang2020tartanair}, FlyingChairs \cite{chairs},  FlyingThings \cite{things}, and HD1K \cite{hd1k} are also used for training.

\subsection{Architecture Details}
\label{sec:arch}
Input images are first encoded using a CNN backbone, and multi-scale features are extracted via a U-Net–style feature pyramid network (FPN). These pyramid features are processed by a series of multi-resolution transformer blocks. By varying the number of transformer blocks and the channel dimensions in the encoder, we define different model variants: (S) with 128 channels and 1 transformer block, (M) with 192 channels and 2 blocks, (L) with 256 channels and 3 blocks, and (XL) with 384 channels and 3 blocks. The local refinement module employs U-Net–style network to further process features and %
applies a convolutional GRU, similar to RAFT \cite{teed2020raft}, to iteratively update flow, confidence, and occlusion maps.

\setlength\tabcolsep{15pt}
\begin{table*}[t]
    \centering
    \resizebox{\linewidth}{!}{
    \begin{tabular}{llcccccc}
    \toprule
    & \multirow{2}{*}[\multirowcenter]{Method} & \multicolumn{2}{c}{Sintel (test)} & \multicolumn{4}{c}{KITTI (test)}\\
    \cmidrule(l{0.5ex}r{0.5ex}){3-4}
    \cmidrule(l{0.5ex}r{0.5ex}){5-8}
    & & Clean$\downarrow$ & Final$\downarrow$ & Fl-all$\downarrow$ & Non-Occ$\downarrow$ & Fl-bg$\downarrow$ & Fl-fg$\downarrow$\\ 
        \midrule
        \multirow{5}{*}[\multirowcenter]{(A)}
        & MemFlow \cite{dong2024memflow} & 1.05 & 1.91 & 4.10 & 2.56 & 3.67 & 6.27 \\ 
        & StreamFlow \cite{sun2024streamflow} & 1.04 & 1.87 & 4.24 & - & - &- \\
        & ARFlow \cite{liu2026arflow} & 0.96 & 1.79 & 2.85 & 1.91 & 2.48 & 4.69 \\
        & VideoFlow \cite{shi2023videoflow} & 0.99 & 1.65 & 3.65 & - & - & - \\
        & MEM-FOF \cite{Bargatin_2025_ICCV} & 0.93 & 1.89 & 2.94 & 1.97 & 2.60 & 4.66 \\
        \midrule
        \midrule
        \multirow{17}{*}[\multirowcenter]{(B)} & SpyNet~\cite{ranjan2017optical} & 6.64 & 8.36 &  35.07 & 26.71 & 33.36 & 43.62\\ 
         & FlowNet2~\cite{ilg2017flownet} & 4.16 & 5.74 & 10.41 & 6.94 & 10.75 & 8.75\\
        & GMFlow~\cite{xu2022gmflow} & 1.74 & 2.90 & 9.32 & 3.80 & 9.67 & 7.57\\
         & PWC-Net+~\cite{sun2019models} & 3.45 & 4.60 & 7.72 & 4.91 & 7.69 & 7.88\\
        & RAFT~\cite{teed2020raft} & 1.61 & 2.86 & 5.10 & 3.07 & 4.74 & 6.87\\
        & GMA~\cite{jiang2021learning} & 1.39 & 2.47 & 5.15 & - & - & - \\
        & DIP~\cite{zheng2022dip} &1.44 & 2.83 & 4.21 & 2.43 & 3.86 & 5.96 \\
        & GMFlowNet~\cite{zhao2022global} & 1.39 & 2.65 & 4.79 & 2.75 & 4.39 & 6.87\\
        & CRAFT~\cite{sui2022craft} & 1.45 & 2.42 & 4.79 & 3.02 & 4.58 & 5.85\\
        & FlowFormer~\cite{huang2022flowformer} & 1.16 &  2.09   & 4.68 & 2.69 & 4.37 & 6.18\\
        & SKFlow~\cite{sun2022skflow} & 1.28 & 2.23 & 4.85 & - & 4.55 & 6.39\\
         & GMFlow+~\cite{xu2023unifying} & 1.03 & 2.37 & 4.49 & 2.40 & 4.27 & 5.60\\
        & EMD-L~\cite{deng2023explicit} & 1.32 & 2.51 & 4.49 & - & 4.16 & 6.15\\
        & RPKNet~\cite{morimitsu2024recurrent} & 1.31 & 2.65 & 4.64 & 2.71 & 4.63 & \fst {4.69}\\ 
         & AnyFlow~\cite{jung2023anyflow} & 1.23 & 2.44 & 4.41 & 2.69 & 4.15 & 5.76 \\
         & SAMFlow~\cite{zhou2024samflow} & 1.00 &  2.08 & 4.49 & - & - & -\\
         & DPFlow~\cite{morimitsu2025dpflow} &  1.04 & \trd 1.97 & 3.56 & 2.12 &  3.29 & \snd 4.93\\
         \midrule
         \multirow{6}{*}[\multirowcenter]{(C)} & CCMR+~\cite{jahedi2024ccmr} & 1.07 & 2.10 & 3.86 & 2.07 & 3.39 & 6.21\\ 
        & MatchFlow(G)~\cite{dong2023rethinking} & 1.16 & 2.37 & 4.63 & 2.77 & 4.33 & 6.11\\
        & Flowformer++\cite{shi2023flowformer++} & 1.07 & \snd {1.94} & 4.52 & - & - & -\\ 
        & CroCoFlow~\cite{weinzaepfel2023croco} & 1.09 & 2.44 & 3.64 & 2.40 &  3.18 & 5.94\\
       & DDVM~\cite{saxena2023surprising} &1.75 &2.48 &\fst  {3.26} & 2.24 & \fst {2.90} & 5.05\\
        & FlowDiffuser~\cite{luo2024flowdiffuser} & 1.02 & 2.03 & 4.17 & 2.82 & 3.68 & 6.64\\
         \midrule
         \multirow{8}{*}[\multirowcenter]{(D)} & SEA-RAFT(L)~\cite{wang2024sea} &1.31 & 2.60 & 4.30 & - & 4.08 & 5.37\\
        & WAFT-DAv2-a1~\cite{wang2025waft} & 1.09 & 2.34 & \trd 3.42 & \trd 2.04 & - & -\\
        & WAFT-Twins-a2~\cite{wang2025waft} & 1.02 & 2.39 & 3.53 & 2.12 &  3.18 & 5.28\\
        & WAFT-DAv2-a2~\cite{wang2025waft} &  0.95 & 2.33 & \snd 3.31 & \snd 2.03 & \snd 2.98 & \trd 4.94\\
        & WAFT-DINOv3-a2~\cite{wang2025waft} &  \trd 0.94 & 2.02 & 3.56 & 2.13 & \trd 3.16 & 5.56\\
        \hdashline
        & \textBF{\net{} (S)} & 1.01 & 2.45 & 4.50 & 2.59 & 4.31 & 5.45\\
        & \textBF{\net{} (M)} &  0.95 & 2.10 & 4.41 & 2.35 & 3.88 & 5.43\\
        & \textBF{\net{} (L)} & \snd 0.91 & 2.19 & 3.81 & 2.10 & 3.55 &  5.08\\
        & \textBF{\net{} (XL)} & \fst \textBF{0.85} & \fst 1.84 & 3.59 & \fst \textBF{1.95} &  3.29 & 5.09\\
        
    \bottomrule
    \end{tabular}
    }\vspace{0.2cm}
    \caption{\textbf{Results on Sintel~\cite{sintel} and KITTI~\cite{kitti} benchmarks.} \net{} (XL) ranks first on both Clean and Final splits of Sintel. On KITTI, it establishes a new state-of-the-art on non-occluded pixels, while remaining competitive with top-performing methods on other metrics. }
    \label{tab:sintel_kitti_benchmark}       
\end{table*}

\subsection{Implementation Details}
\label{sec:impl}
\net{} is implemented in PyTorch \cite{paszke2017automatic} and trained with the AdamW optimizer \cite{loshchilov2017decoupled} using a one-cycle learning rate scheduler \cite{smith2019super}, peaking at $10^{-5}$, and with gradients clipped to $[-1, 1]$. All experiments perform three local refinement iterations during both training and inference, unless otherwise stated. For the localized soft expectation, we use a window radius of $r=2$. The small constant $\epsilon$ is set to $10^{-6}$. A key advantage of our architecture is its compatibility with the pretrained weights from \cite{min2025s2m2}; while the tasks are different and the architectures are not identical, we initialize compatible layers using their pretrained weights. %

\boldparagraph{Training Schedule}
We train \net{} on four NVIDIA H100 GPUs with 64GB memory each. Our training schedule follows SEA-RAFT \cite{wang2024sea}. We first pretrain \net{} on TartanAir \cite{wang2020tartanair} for 300K iterations at a resolution of $480\times640$, using a batch size of 8 for the (S) model and 4 for the larger variants. We then train on FlyingChairs \cite{chairs} at $352\times480$ resolution, with batch sizes of 16 for the (S) and (M) models and 8 for the (L) and (XL) models for 100K iterations. Next, we train on FlyingThings \cite{things} (denoted as “C+T” following prior work) with a batch size of 4 for 120K iterations. 
Subsequently, we train on a mixture of FlyingThings \cite{things}, Sintel \cite{sintel}, KITTI 2015 \cite{kitti}, and HD1K \cite{hd1k} (denoted as “C+T+S+K+H”) with a batch size of 4 for 300K iterations, and use this model for submission to the Sintel benchmark. Finally, for submissions to the KITTI 2015 \cite{kitti} benchmark, we further fine-tune the “C+T+S+K+H” models on the KITTI training set for an additional 10K iterations with a batch size of 4. %

\begin{figure*}[t]
    \centering
    \renewcommand{\tabcolsep}{0.5pt}
    \begin{tabular}{cccc}
    \rotatebox[origin=l]{90}{\tiny{\quad Image 1}} &
    \hspace{0.1cm}\includegraphics[width=0.32\textwidth]{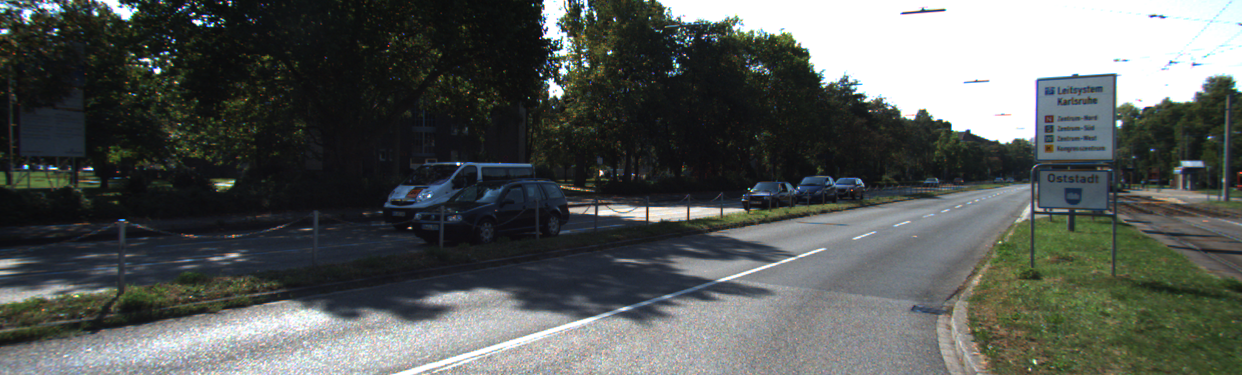} &
     \hspace{0.1cm}\includegraphics[width=0.32\textwidth]{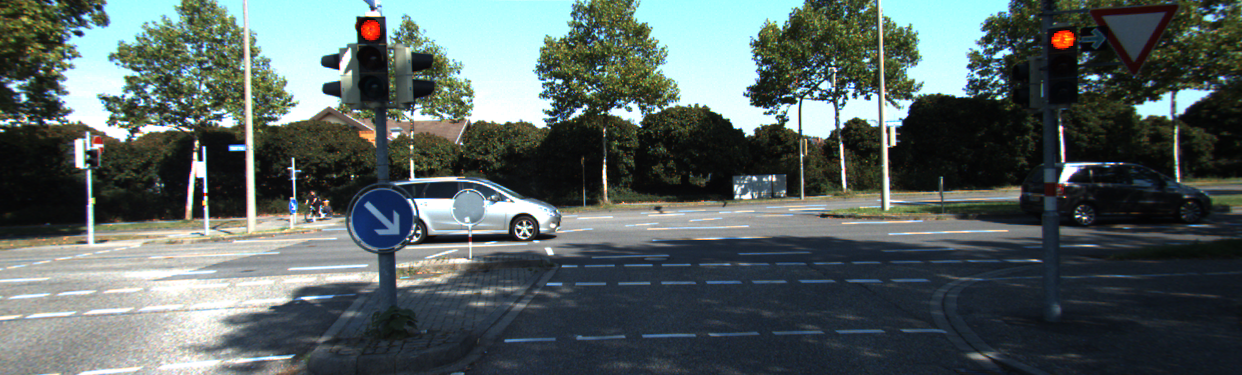}
     &
     \hspace{0.1cm}\includegraphics[width=0.32\textwidth]{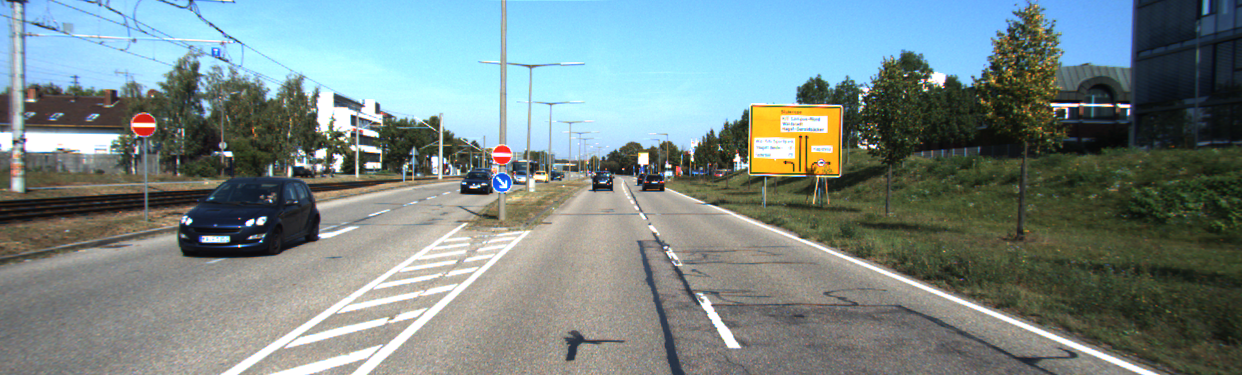}
    \\ 
    \rotatebox[origin=l]{90}{\tiny{AnyFlow \cite{jung2023anyflow}}} &
    \begin{overpic}[width=0.32\textwidth]{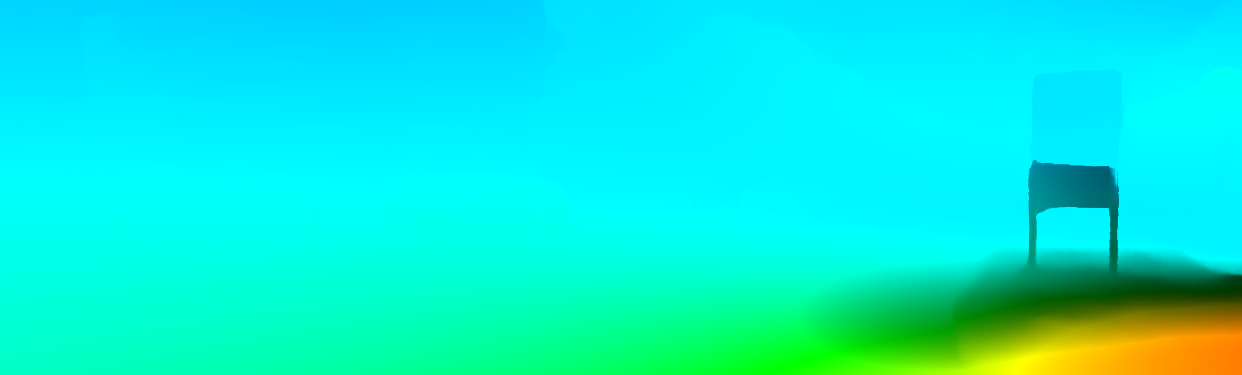} \put (3,24) {\textcolor{purple}{\textbf{\texttt{Fl-all: 2.34}}}} \end{overpic} &
     \begin{overpic}[width=0.32\textwidth]{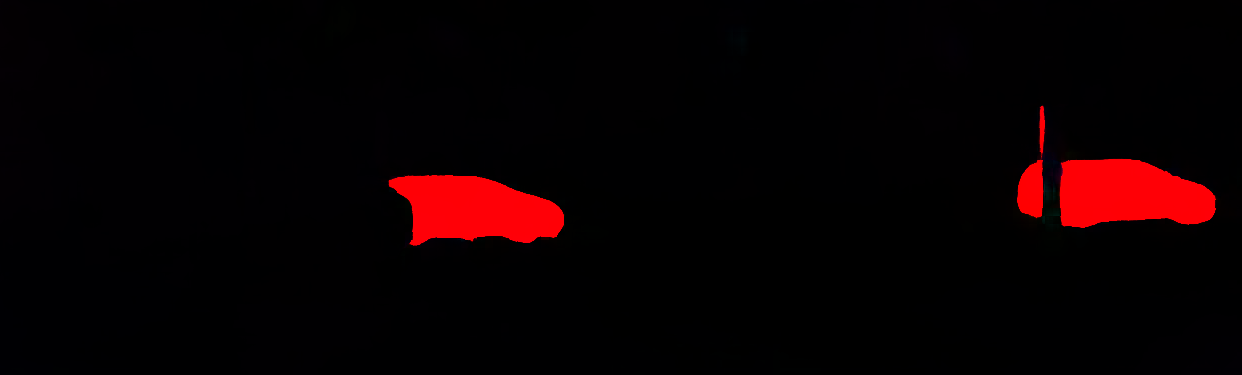}  \put (3,24) {\textcolor{purple}{\textbf{\texttt{Fl-all: 1.53}}}} \end{overpic}
     &
     \begin{overpic}[width=0.32\textwidth]{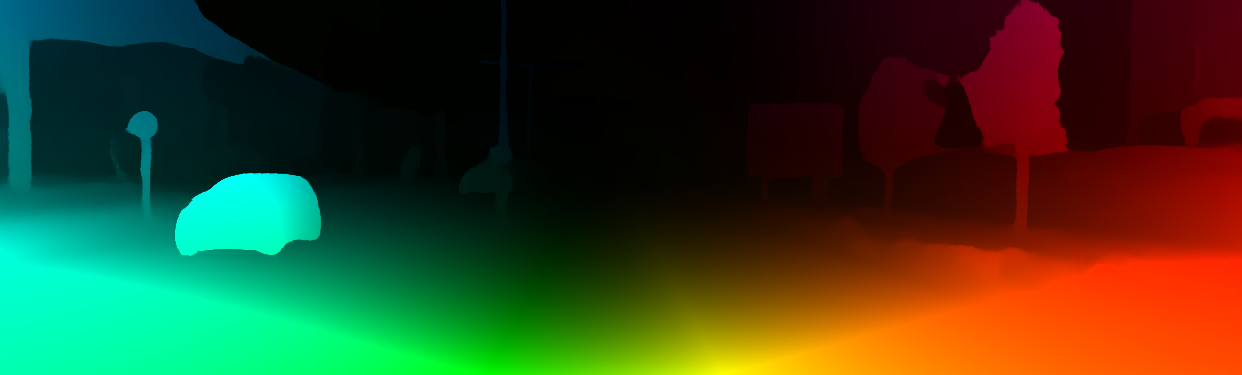}  \put (3,24) {\textcolor{purple}{\textbf{\texttt{Fl-all: 3.63}}}} \end{overpic}
    \\
    \rotatebox[origin=l]{90}{\tiny{RPKNet \cite{morimitsu2024recurrent}}} &
    \begin{overpic}[width=0.32\textwidth]{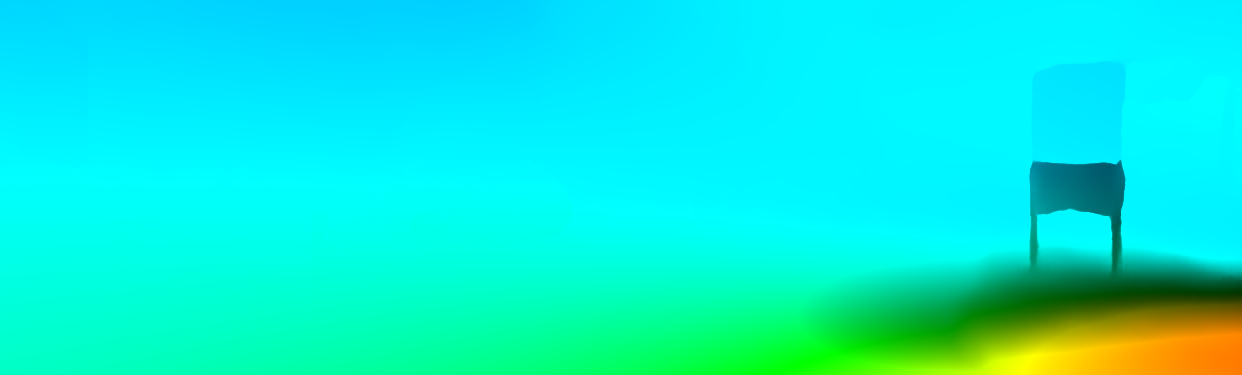} \put (3,24) {\textcolor{purple}{\textbf{\texttt{Fl-all: 2.30}}}} \end{overpic} &
     \begin{overpic}[width=0.32\textwidth]{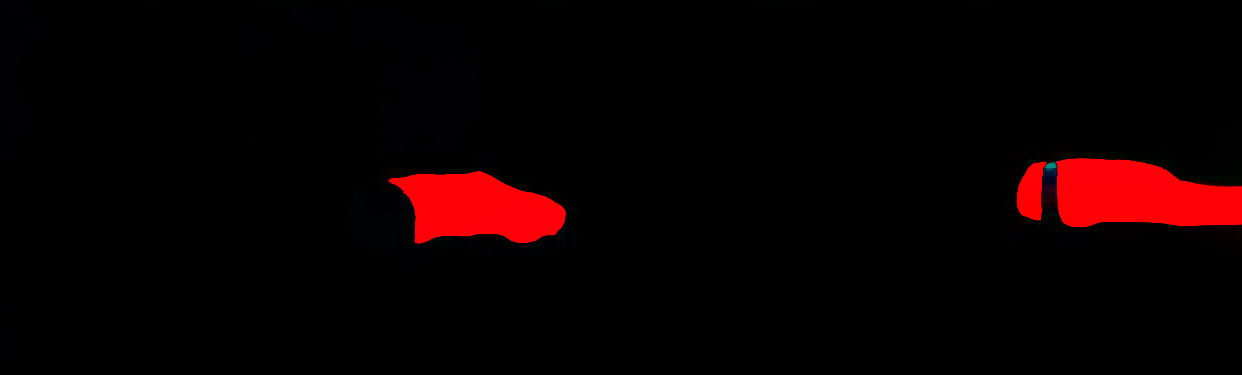}  \put (3,24) {\textcolor{purple}{\textbf{\texttt{Fl-all: 1.70}}}} \end{overpic}
     &
     \begin{overpic}[width=0.32\textwidth]{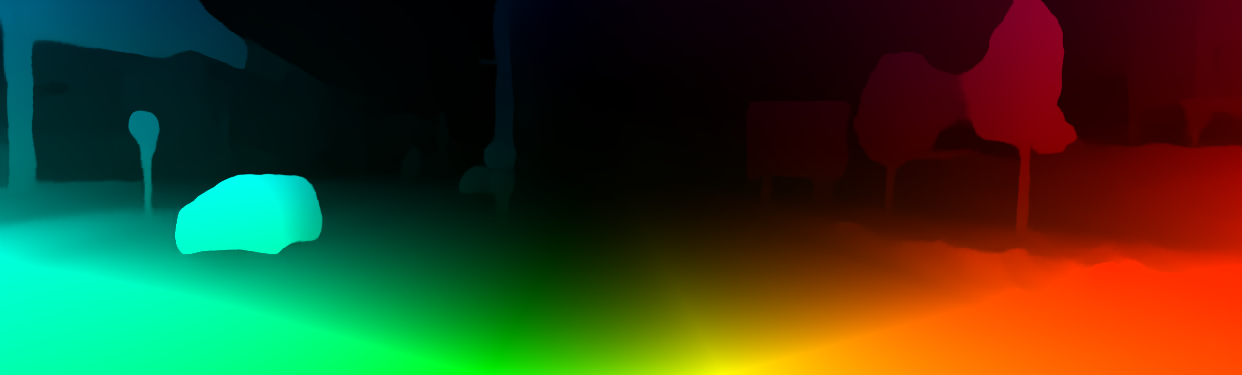}  \put (3,24) {\textcolor{purple}{\textbf{\texttt{Fl-all: 3.55}}}} \end{overpic}
    \\
    \rotatebox[origin=l]{90}{\tiny{ DPFlow~\cite{morimitsu2025dpflow}}} &
    \begin{overpic}[width=0.32\textwidth]{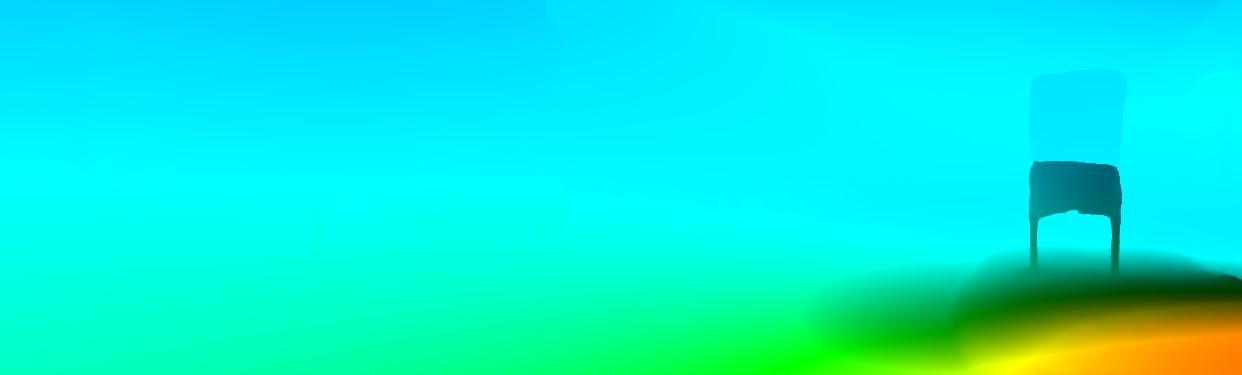} \put (3,24) {\textcolor{purple}{\textbf{\texttt{Fl-all: 2.36}}}}
    \end{overpic} &
     \begin{overpic}[width=0.32\textwidth]{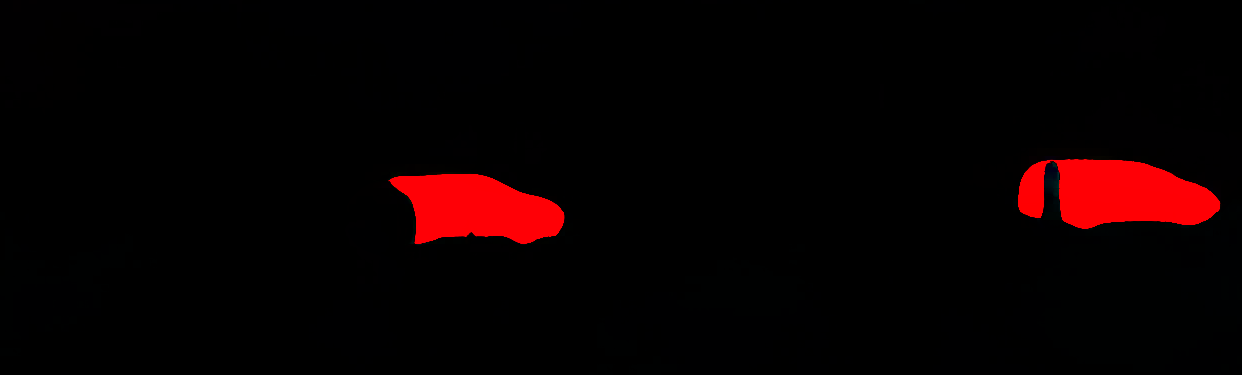} \put (3,24) {\textcolor{purple}{\textbf{\texttt{Fl-all: 1.36}}}}
     \end{overpic}
     &
     \begin{overpic}[width=0.32\textwidth]{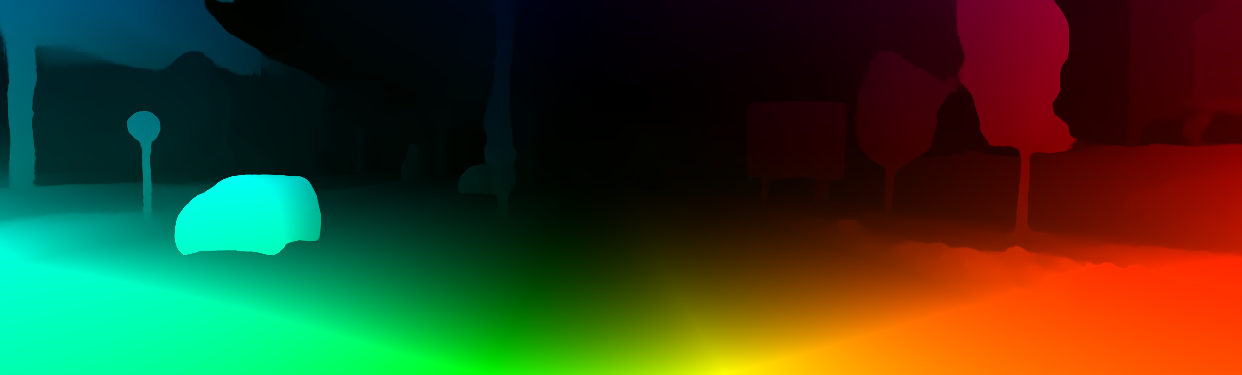} \put (3,24) {\textcolor{purple}{\textbf{\texttt{Fl-all: 2.75}}}}
     \end{overpic}
    \\
    \rotatebox[origin=l]{90}{\tiny{WAFT~\cite{wang2025waft}}} &
    \begin{overpic}[width=0.32\textwidth]{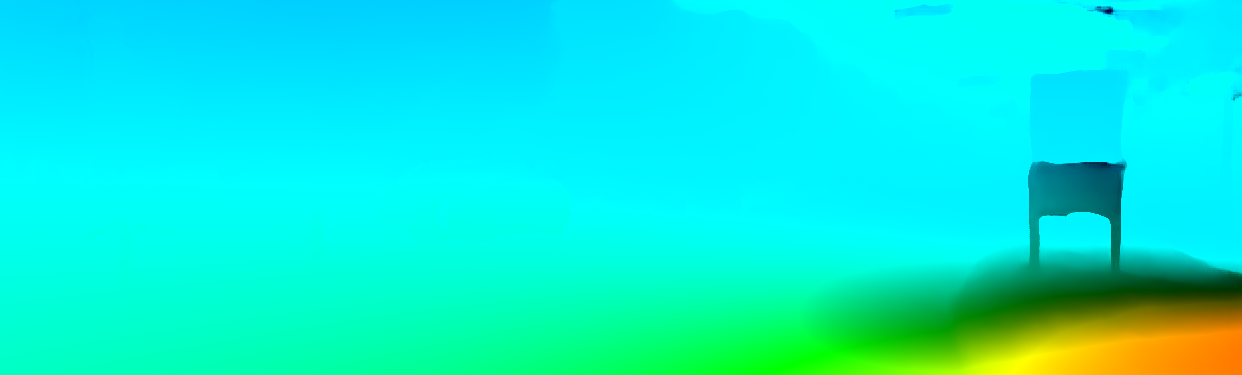} 
    \put (3,24) {\textcolor{purple}{\textbf{\texttt{Fl-all: 2.69}}}}
    \end{overpic} &
     \begin{overpic}[width=0.32\textwidth]{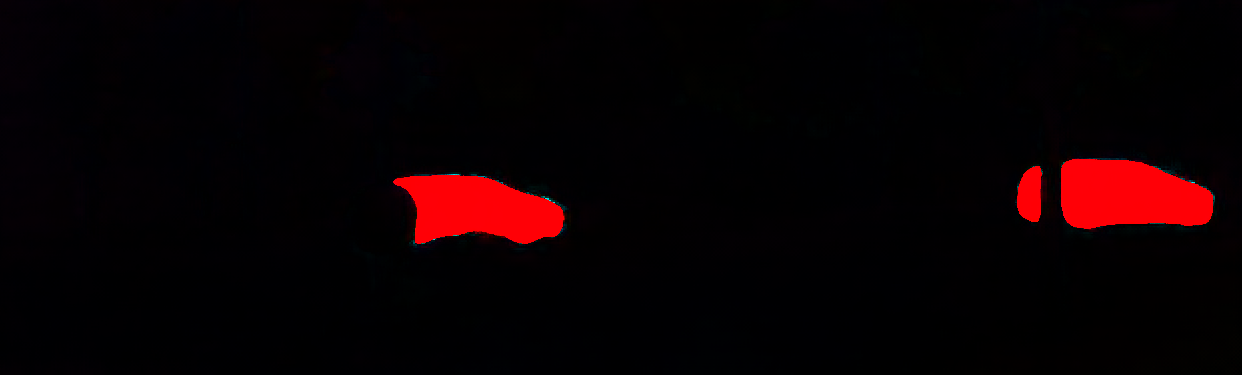} \put (3,24) {\textcolor{purple}{\textbf{\texttt{Fl-all: 1.48}}}}\end{overpic}
     &
     \begin{overpic}[width=0.32\textwidth]{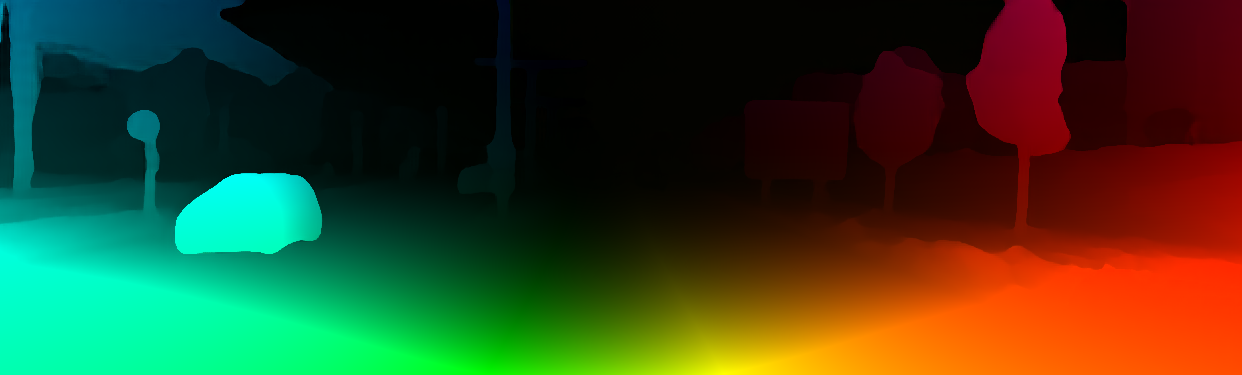} \put (3,24) {\textcolor{purple}{\textbf{\texttt{Fl-all: 2.70}}}}\end{overpic}
    \\
    \rotatebox[origin=l]{90}{\tiny{\net{} (XL)}} &
    \begin{overpic}[width=0.32\textwidth]{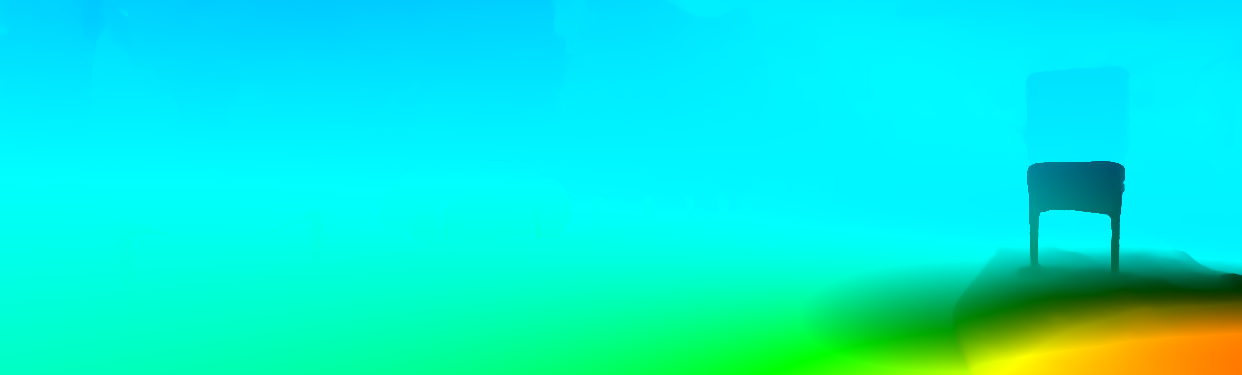} \put (3,24) {\textcolor{purple}{\textbf{\texttt{Fl-all: 1.76}}}} \end{overpic} &
     \begin{overpic}[width=0.32\textwidth]{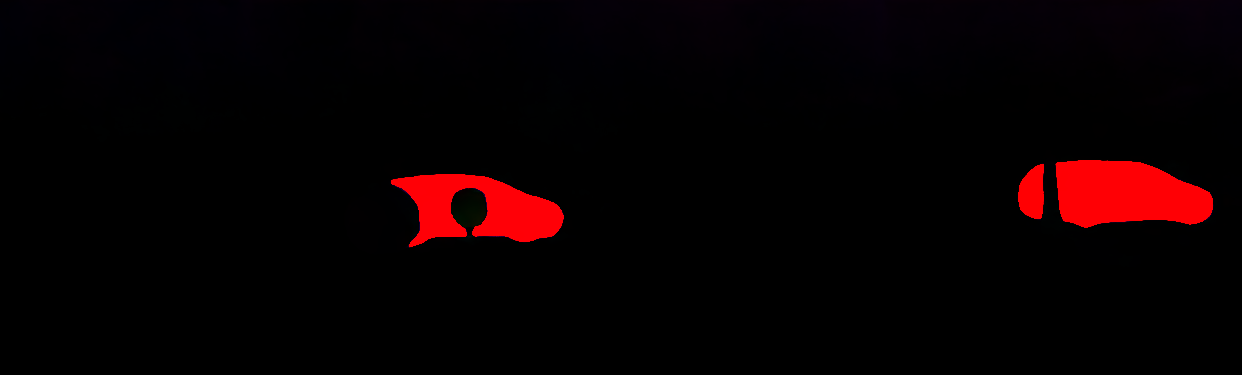}  \put (3,24) {\textcolor{purple}{\textbf{\texttt{Fl-all: 0.78}}}} \end{overpic}
     &
     \begin{overpic}[width=0.32\textwidth]{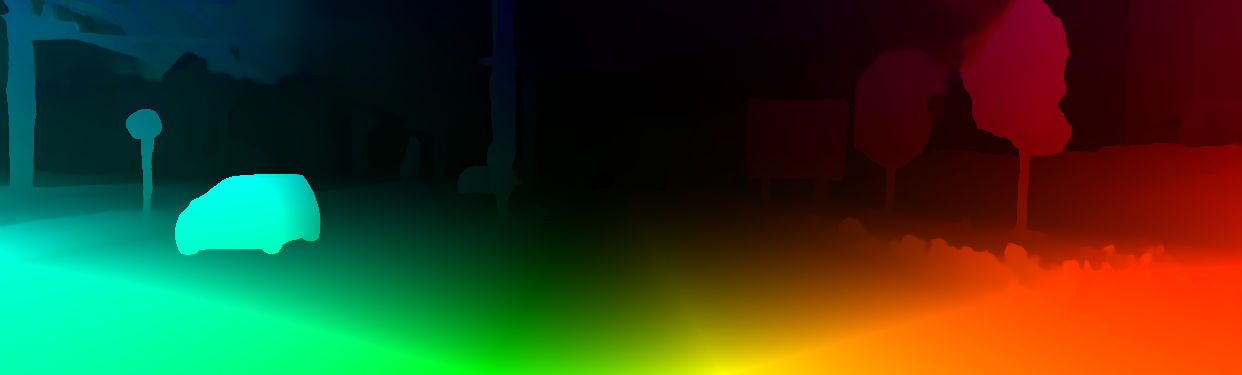}  \put (3,24) {\textcolor{purple}{\textbf{\texttt{Fl-all: 2.34}}}} \end{overpic}
    \\
    \end{tabular}
    \vspace{0.1cm}
    \caption{\textbf{Qualitative Results on KITTI~\cite{kitti} Test Set.} Visualizations and error metrics are obtained directly from the official evaluation server. For WAFT, we report results using the DAv2-a2 variant, which achieves the best performance among WAFT models on the KITTI benchmark.}
    \label{fig:kitti}
\end{figure*}

\subsection{Benchmark Results}
\label{sec:benchmark}
We report results on the Sintel and KITTI benchmarks in Table~\ref{tab:sintel_kitti_benchmark}. 
At the very top, the table reports results achieved by models processing more than 2 consecutive frames (A), usually 3 or 5, while the remaining rows concern 2-frame flow methods, either trained following the mainstream datasets protocol (B), using supplementary training datasets (C), or a simple TartanAir pre-training stage (D). A more comprehensive version of this table, specifying the exact additional data used by each model, is presented in the Supplementary Material.
We highlight the \colorbox{First}{\textbf{best}}, \colorbox{Second}{second-best}, and \colorbox{Third}{third-best} results among 2-frame models. 

Among these latter models, \net{} (XL) achieves the best performance on the Sintel benchmark, on both Clean and Final splits. Furthermore, on Sintel Clean split, it also outperforms multi-frame models \cite{liu2026arflow,shi2023videoflow,Bargatin_2025_ICCV, sun2024streamflow, dong2024memflow} while being very competitive on Final split, with only a small gap with respect to VideoFlow \cite{shi2023videoflow} and ARFlow~\cite{liu2026arflow}.
Our smaller variants (L), (M), and (S) remain competitive with WAFT and other state-of-the-art methods on this dataset, with increasing gaps with respect to our flagship model, yet with consistent gains in terms of efficiency, as we will discuss in detail later.  
Qualitative comparisons of flow predictions on the Sintel benchmark are provided in the Supplementary Material %

On KITTI, our (XL) model achieves the best performance on non-occluded pixels, while still being competitive in terms of overall Fl-all score. %
Here, the difference with multi-frame models is larger, mainly because of the camera egomotion induced by the car being dominant with respect to independently moving objects, making the use of multiple frames even more effective.
Figure~\ref{fig:kitti} shows qualitative results on the KITTI benchmark, comparing the predictions by representative methods -- WAFT \cite{wang2025waft}, DPFlow \cite{morimitsu2025dpflow}, AnyFlow \cite{jung2023anyflow}, and RPKNet \cite{morimitsu2024recurrent} -- with those by \net{}. 
Over each flow map, we report the Fl-all score, confirming the very competitive performance by our model.

\subsection{Zero-Shot Generalization Results}
\label{sec:zeroshot}

\setlength\tabcolsep{5pt}
\begin{wraptable}{r}{0.5\textwidth}\vspace{-0.5cm}
\centering
\resizebox{0.5\textwidth}{!}{
\begin{tabular}{lcccc}
\toprule
\multirow{2}{*}[\multirowcenter]{Method} & \multicolumn{2}{c}{Sintel (train)} & \multicolumn{2}{c}{KITTI (train)}\\
\cmidrule(l{0.5ex}r{0.5ex}){2-3}\cmidrule(l{0.5ex}r{0.5ex}){4-5}
    & Clean$\downarrow$ & Final$\downarrow$ & Fl-epe$\downarrow$ & Fl-all$\downarrow$ \\ 
\midrule
    PWC-Net~\cite{sun2018pwc} & 2.55 & 3.93 & 10.4 & 33.7\\
    FlowNet2~\cite{ilg2017flownet} & 2.02 & 3.14 & 10.1 & 30.4 \\
    RAFT~\cite{teed2020raft} & 1.43 & 2.71 & 5.04 & 17.4\\
    GMA~\cite{jiang2021learning} & 1.30 & 2.74 & 4.69 & 17.1\\
    SKFlow~\cite{sun2022skflow} & 1.22 & 2.46 & 4.27 & 15.5\\
    DIP~\cite{zheng2022dip} & 1.30 & 2.82 & 4.29 & 13.7\\
    EMD-L~\cite{deng2023explicit} &  0.88 & 2.55 & 4.12 & 13.5\\
    CRAFT~\cite{sui2022craft} & 1.27 & 2.79 & 4.88 & 17.5\\
    RPKNet~\cite{morimitsu2024recurrent} & 1.12 & 2.45 & - & 13.0\\
    GMFlowNet~\cite{zhao2022global} & 1.14 & 2.71 & 4.24 & 15.4\\
    FlowFormer~\cite{huang2022flowformer} & 1.01 & 2.40 & 4.09 & 14.7\\
    Flowformer++~\cite{shi2023flowformer++} & 0.90 & 2.30 & 3.93 & 14.2\\ 
    CCMR+~\cite{jahedi2024ccmr} & 0.98 & 2.36 & - & 12.9\\
    MatchFlow(G)~\cite{dong2023rethinking} &1.03& 2.45 & 4.08 & 15.6 \\
    SEA-RAFT(L)~\cite{wang2024sea} &1.19 & 4.11 & 3.62 & 12.9\\
    AnyFlow~\cite{jung2023anyflow} & 1.10 & 2.52 & 3.76 & 12.4 \\
    SAMFlow~\cite{zhou2024samflow} &  0.87 & \trd {2.11} & 3.44 & 12.3 \\
    FlowDiffuser~\cite{luo2024flowdiffuser} & \trd 0.86 &  2.19 & 3.61 & 11.8 \\
    DPFlow~\cite{morimitsu2025dpflow} & 1.02 & 2.26 & 3.37 & 11.1 \\
    FlowSeek (T) \cite{Poggi_2025_ICCV} & 1.13 & 2.48 & 4.06 & 12.2 \\
    FlowSeek (S) \cite{Poggi_2025_ICCV} & 1.05 & 2.37 & 3.32 & 11.0 \\
    FlowSeek (M) \cite{Poggi_2025_ICCV} & 1.10 & 2.31 & 3.99 & 12.1 \\
    FlowSeek (L) \cite{Poggi_2025_ICCV} & 1.03 & 2.18 & 3.31 & 11.2 \\
    {WAFT-DAv2-a1} \cite{wang2025waft} & 1.00 &  2.15 & \trd {3.10} & \snd {10.3} \\
    {WAFT-Twins-a2} \cite{wang2025waft} & 1.02 & 2.46 & \snd {2.98} & \fst {9.9} \\
    {WAFT-DAv2-a2} \cite{wang2025waft} & 1.01 & 2.49 &  {3.28} & \trd {10.9} \\
    {WAFT-DINOv3-a2} \cite{wang2025waft} & 1.28 & 2.56 & 3.49 & 12.9 \\ 
    \hdashline
    \bf \net{} (S) & 0.90 &  2.18  & 4.81 & 17.3 \\
    \bf \net{} (M) & 0.91 & 2.17 & 3.98 & 14.7 \\
    \bf \net{} (L) &  \snd 0.83 & \fst 1.99 & 3.50 & 13.2 \\
    \bf \net{} (XL) &  \fst 0.77 &  \snd 2.04 & \fst 2.87 & 11.2 \\
\bottomrule
\end{tabular}
}\vspace{0.2cm}
\caption{\textbf{Zero-Shot Generalization on Sintel (train) and KITTI (train).}}
\label{tab:kitti_sintel_zero_shot}
\end{wraptable}
\textbf{Sintel.} To assess the generalization capability of our models, %
we evaluate the zero-shot performance of \net{} on the training splits of Sintel \cite{sintel}, using the “C+T” models for all methods, following prior work~\cite{teed2020raft, huang2022flowformer, wang2024sea, wang2025waft}. The results are summarized in Table~\ref{tab:kitti_sintel_zero_shot}.
\net{} (XL) and (L) achieve the best and second-best results on both splits of the Sintel training set.
Specifically, on the Final split of Sintel (train), \net{} (L) achieves state-of-the-art performance, closely followed by \net{} (XL), while our other variants outperform all other methods except SAMFlow \cite{zhou2024samflow} and WAFT-DAv2-a1. On the Clean split of Sintel, all \net{} variants outperform the WAFT variants, with the (XL) model achieving the absolute best result -- i.e., 0.77. 
The superiority of \net{} against WAFT is confirmed qualitatively in Fig.~\ref{fig:sintel}. In particular, we can appreciate the much finer details in the flow fields predicted by our model, or some very subtle motion patterns that WAFT fails to perceive, whereas \net{} successfully models -- i.e., the snow on the ground shown in the second row, being moved by the mallet.

\textbf{KITTI.} On the KITTI dataset, our (XL) model achieves the absolute best Fl-epe score, with a consistent gain over both WAFT and FlowSeek, while lagging a little behind in terms of Fl-all.

\textbf{Spring.} Following \cite{wang2024sea}, we evaluate generalization performance on the Spring (train) dataset using the “C+T+S+K+H” models, as shown in \tabref{tab:spring_layered_zero_shot} (left). For WAFT, we use the publicly available checkpoints released for zero-shot evaluations.
Our \net{} (XL) and (L) variants outperform all SEA-RAFT, FlowSeek, and WAFT models, as well as all other methods, including MS-RAFT+, whose training protocol deviates from ours and involves additional training data. 
Our smallest variant, \net{} (S), also surpasses all other approaches on the EPE metric, while achieving performance competitive with those that use the same training data on the 1px metric. %
This further confirms the good generalization capability achieved by \net{}. We provide qualitative comparisons of optical flow predictions from different models on this dataset in the Supplementary Material. %

\textbf{LayeredFlow.} We conclude our evaluation by conducting a further zero-shot generalization experiment on the recent
LayeredFlow dataset \cite{wen2024layeredflow}, which features challenging transparent and reflective surfaces. 
Following \cite{Poggi_2025_ICCV}, similar to the evaluation on the Spring, we use the “C+T+S+K+H” models, except for WAFT, where we again use the models specific for zero-shot evaluation. Evaluation is conducted on the validation split, with input images downsampled to $\frac{1}{8}$ of their original resolution. We report performance on the first layer (i.e., the surfaces closest to the camera). Results are presented in Table~\ref{tab:spring_layered_zero_shot} (right).
Our (XL) model achieves the best results across all metrics. Furthermore, our (L) variant achieves 2nd-best results on the EPE and 1px metrics. 
Fig.~\ref{fig:layered} shows qualitative comparisons between SEA-RAFT~\cite{wang2024sea}, WAFT~\cite{wang2025waft}, FlowSeek~\cite{Poggi_2025_ICCV}, and \net{}, highlighting finer details and more accurate flow in \net{}.

\begin{figure*}[t]
    \centering
    \renewcommand{\tabcolsep}{0pt}
    \begin{tabular}{cccc}
    
    \includegraphics[width=0.24\textwidth, frame]{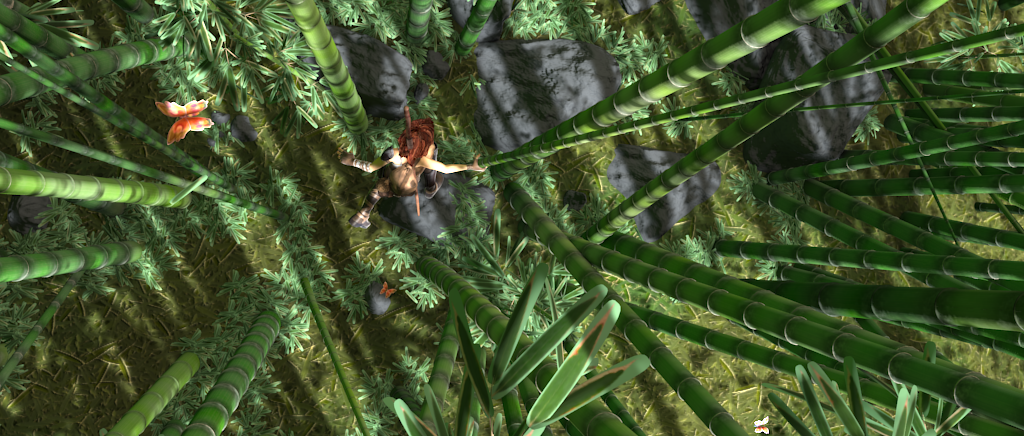} &
    \begin{overpic}[width=0.24\textwidth, frame]{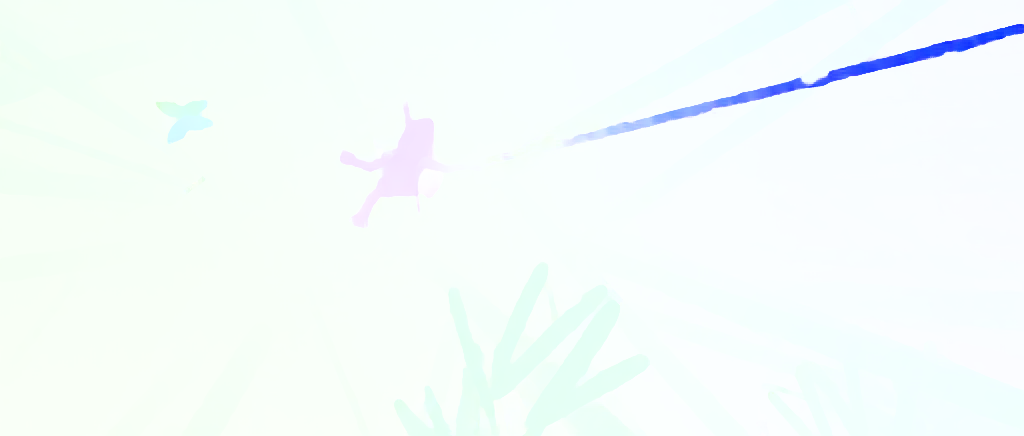}
    \put (2,35) {\textcolor{purple}{\textbf{\texttt{EPE: 0.708}}}}
    \end{overpic} &
    \begin{overpic}[width=0.24\textwidth, frame]{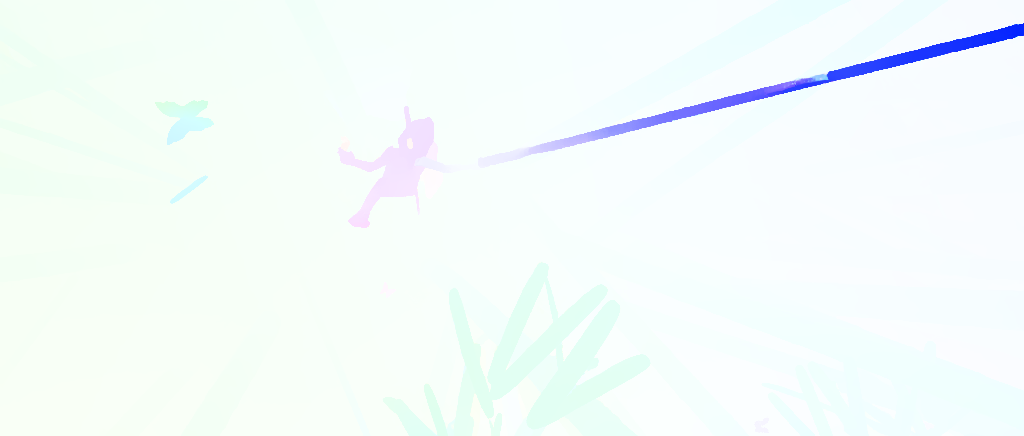}
    \put (2,35) {\textcolor{purple}{\textbf{\texttt{EPE: 0.540}}}}
    \end{overpic} &
    \begin{overpic}[width=0.24\textwidth, frame]{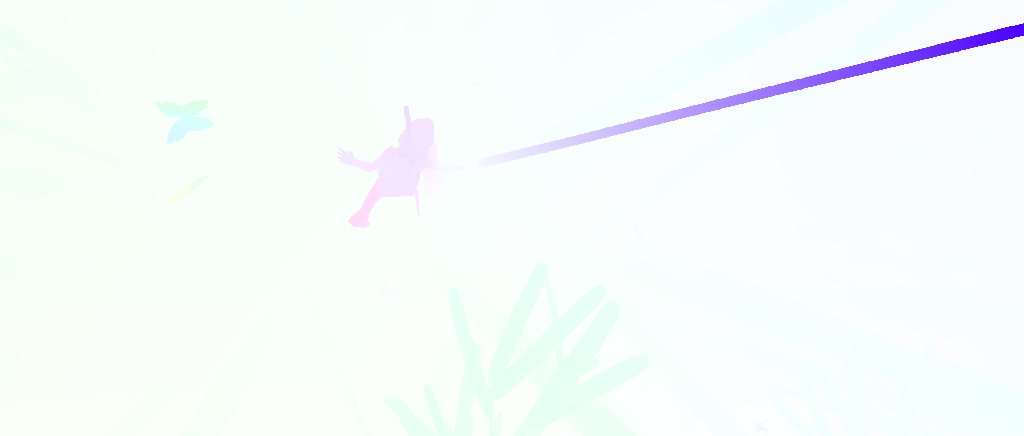}\end{overpic} \\
    \includegraphics[width=0.24\textwidth, frame]{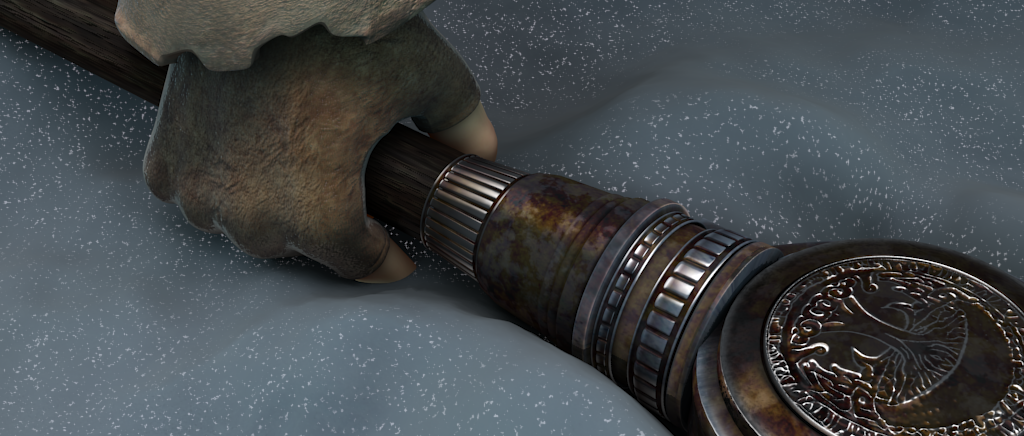} &
    \begin{overpic}[width=0.24\textwidth, frame]{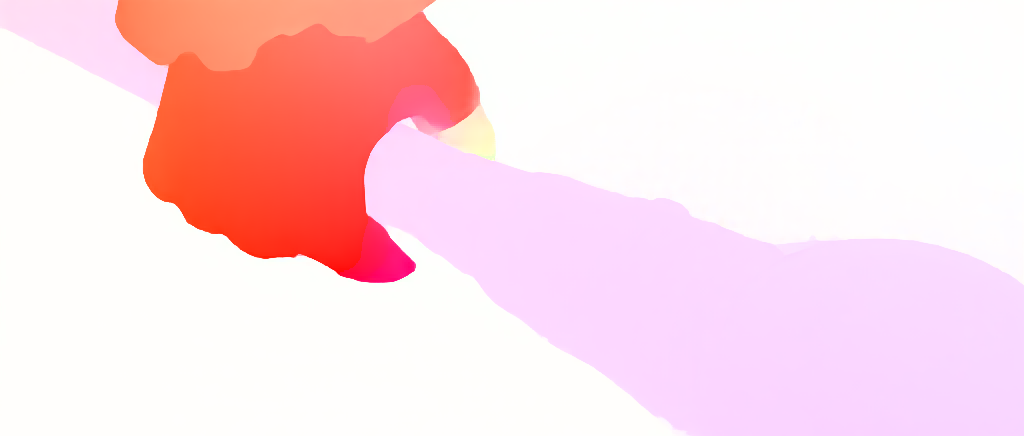}
    \put (2,35) {\textcolor{purple}{\textbf{\texttt{EPE: 0.871}}}}
    \end{overpic} &
    \begin{overpic}[width=0.24\textwidth, frame]{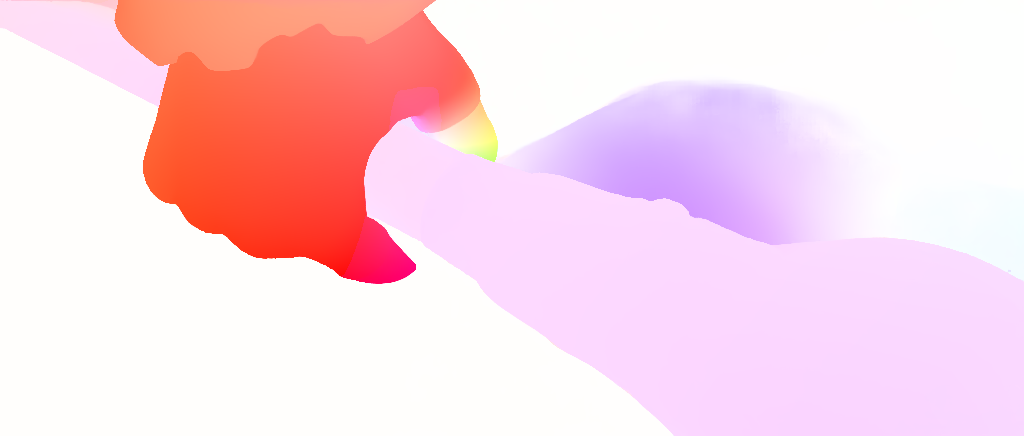}
    \put (2,35) {\textcolor{purple}{\textbf{\texttt{EPE: 0.257}}}}
    \end{overpic} &
    \begin{overpic}[width=0.24\textwidth, frame]{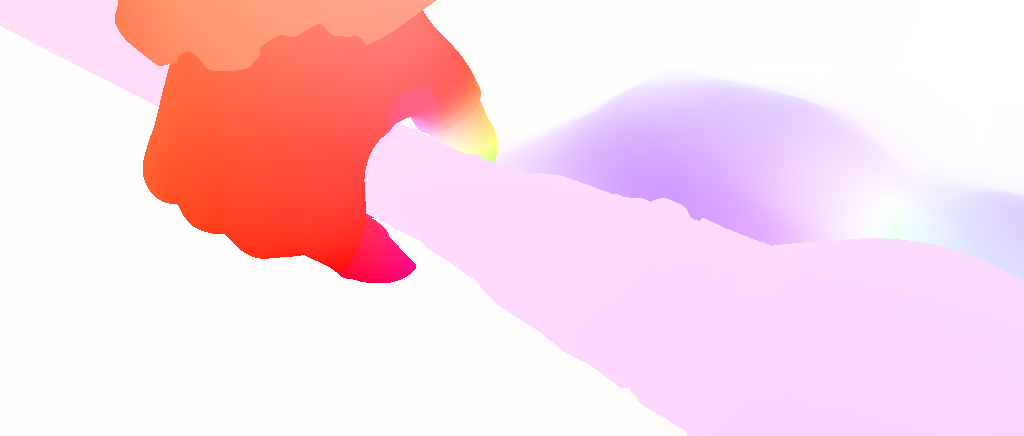}\end{overpic} \\
    \includegraphics[width=0.24\textwidth, frame]{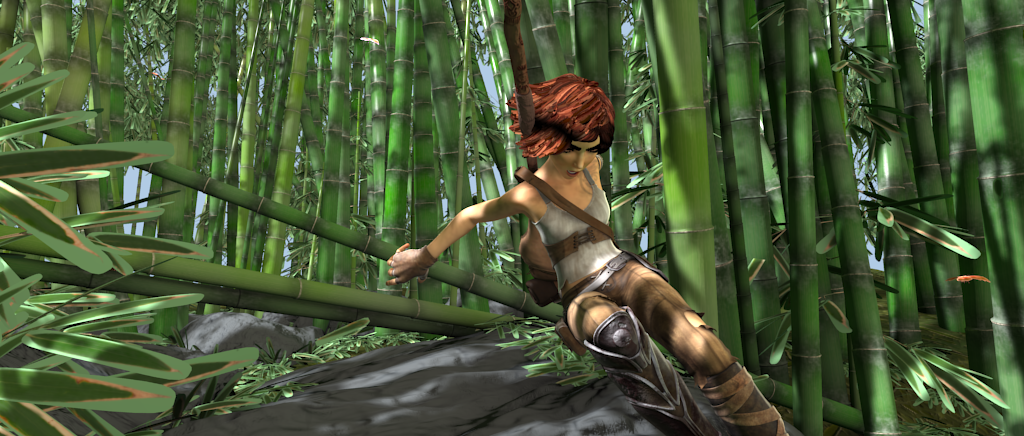} &
    \begin{overpic}[width=0.24\textwidth, frame]{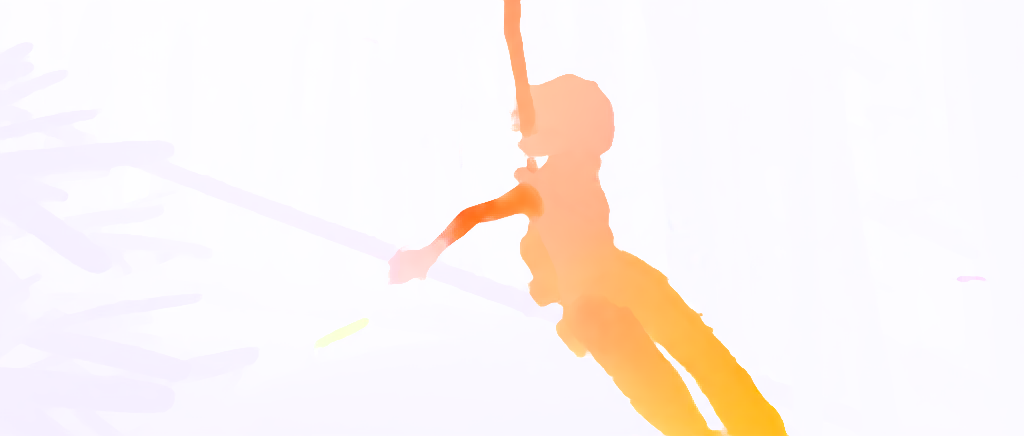}
    \put (2,35) {\textcolor{purple}{\textbf{\texttt{EPE: 1.237}}}}
    \end{overpic} &
    \begin{overpic}[width=0.24\textwidth, frame]{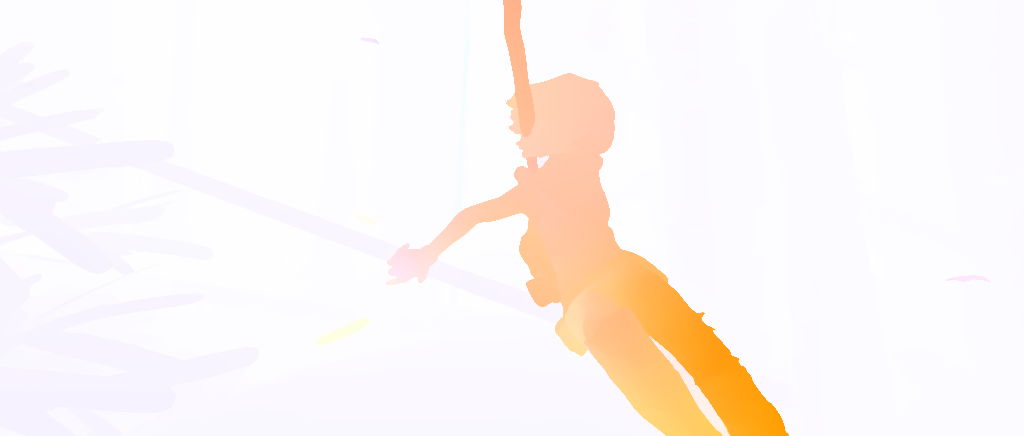}
    \put (2,35) {\textcolor{purple}{\textbf{\texttt{EPE: 0.450}}}}
    \end{overpic} &
    \begin{overpic}[width=0.24\textwidth, frame]{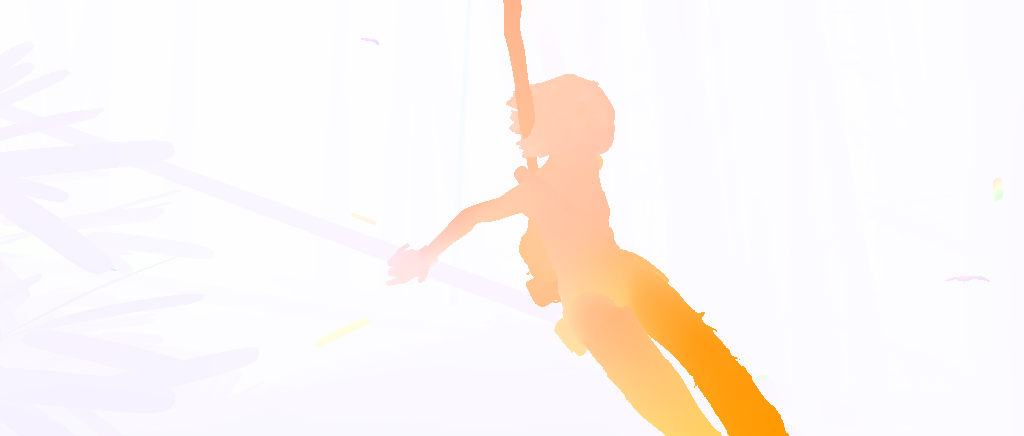}\end{overpic} \\
    \includegraphics[width=0.24\textwidth, frame]{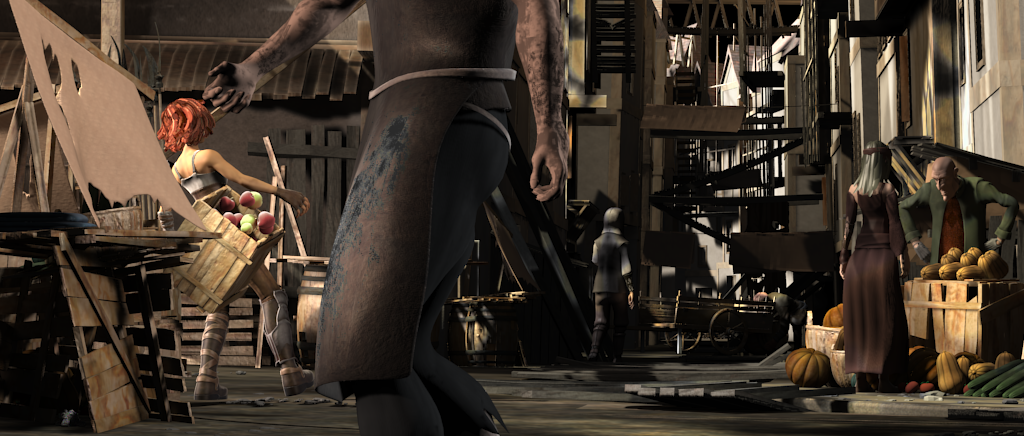} &
    \begin{overpic}[width=0.24\textwidth, frame]{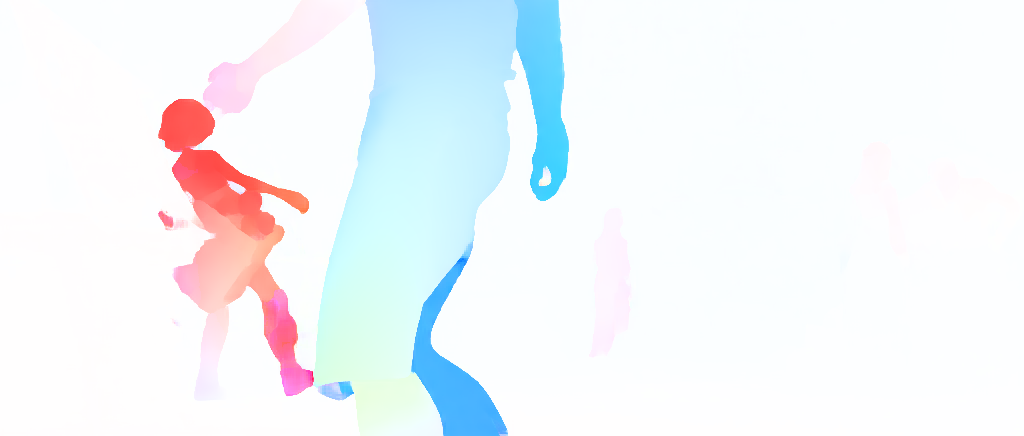}
    \put (2,35) {\textcolor{purple}{\textbf{\texttt{EPE: 0.521}}}}
    \end{overpic} &
    \begin{overpic}[width=0.24\textwidth, frame]{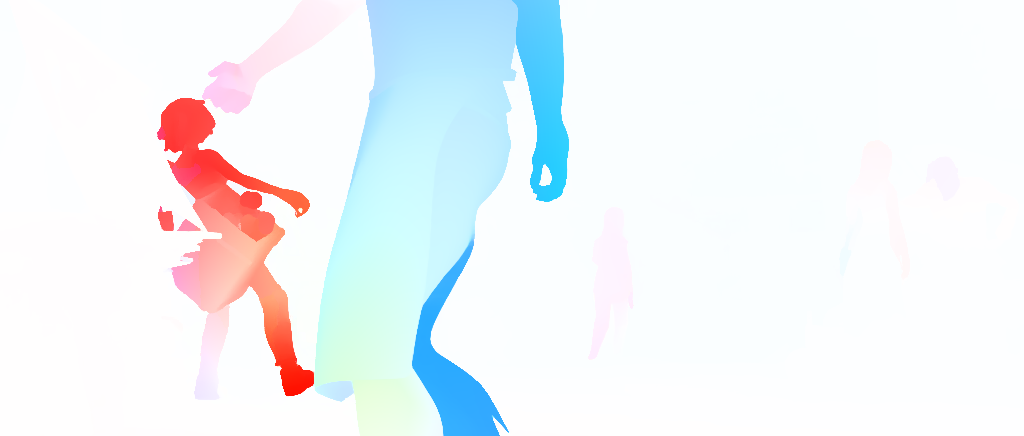}
    \put (2,35) {\textcolor{purple}{\textbf{\texttt{EPE: 0.268}}}}
    \end{overpic} &
    \begin{overpic}[width=0.24\textwidth, frame]{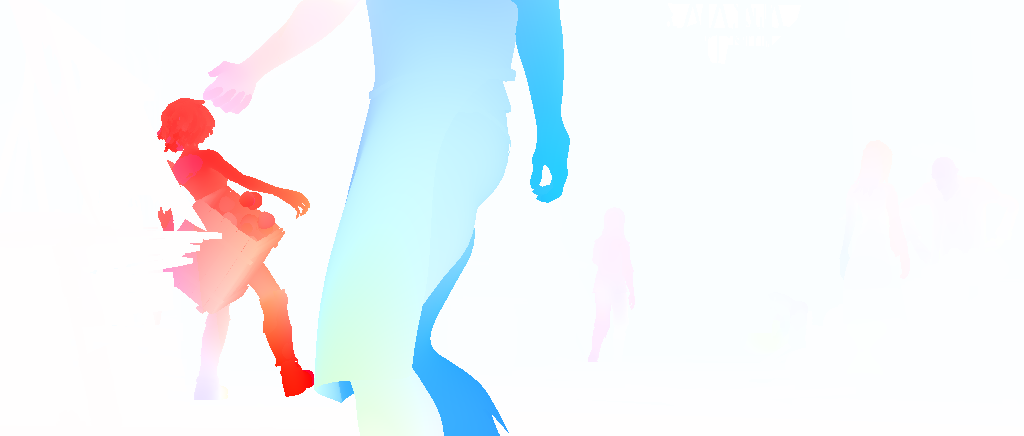}\end{overpic} \\

    \includegraphics[width=0.24\textwidth, frame]{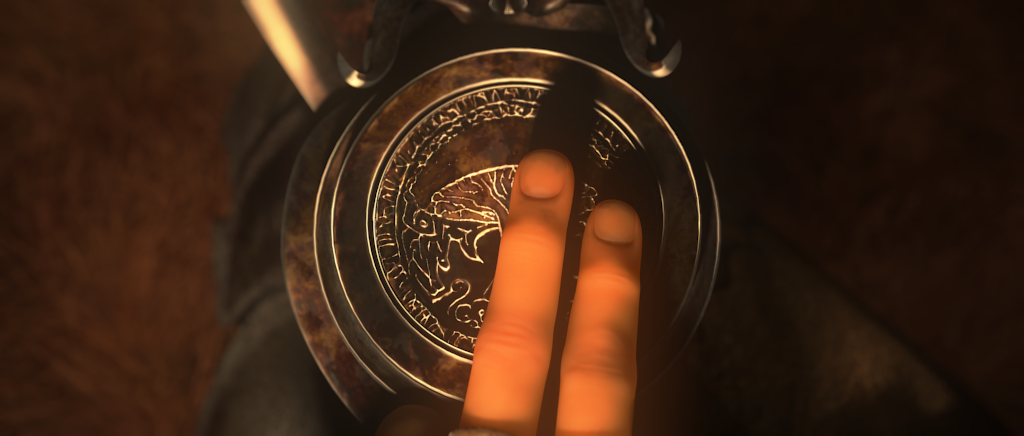} &
    \begin{overpic}[width=0.24\textwidth, frame]{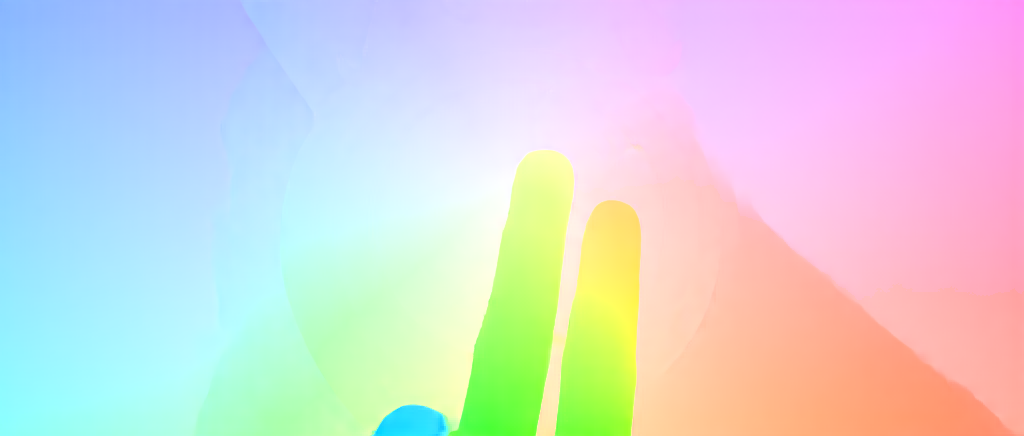}
    \put (3,35) {\textcolor{purple}{\textbf{\texttt{EPE: 0.263}}}}
    \end{overpic} &
    \begin{overpic}[width=0.24\textwidth, frame]{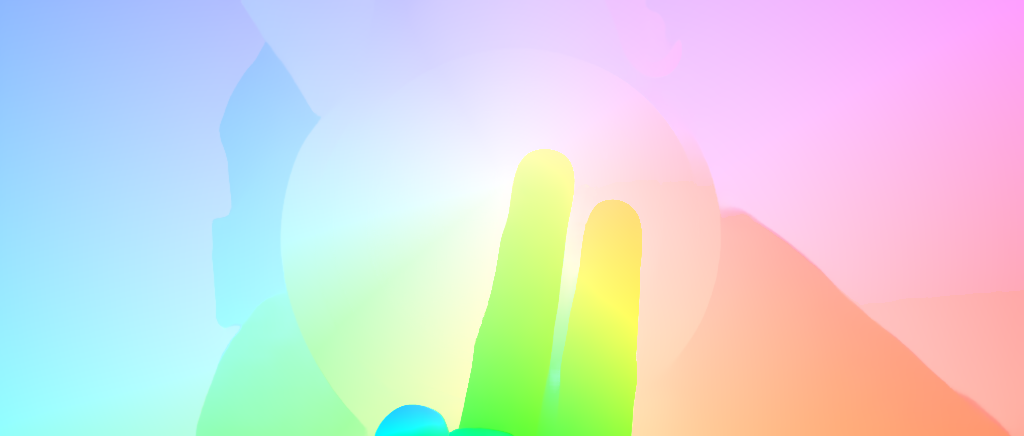} \put (3,35) {\textcolor{purple}{\textbf{\texttt{EPE: 0.228}}}} \end{overpic} &
    \begin{overpic}[width=0.24\textwidth, frame]{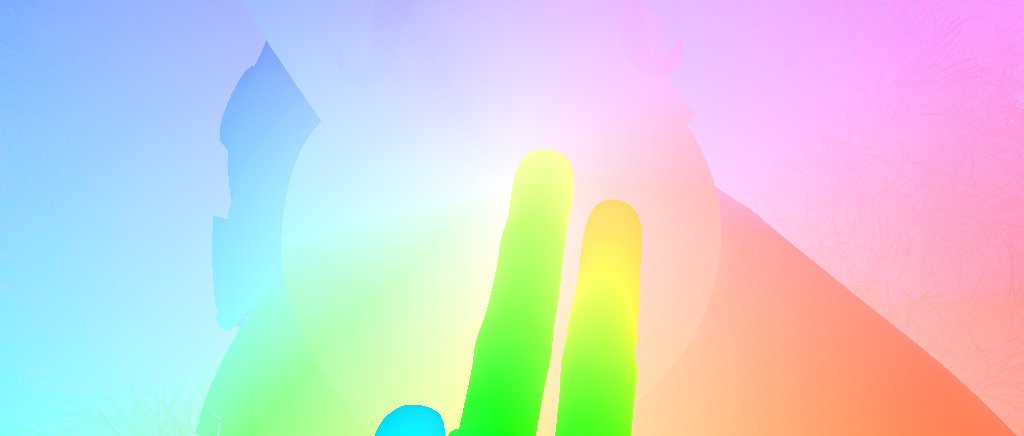}
    \end{overpic}
    \\

    \includegraphics[width=0.24\textwidth, frame]{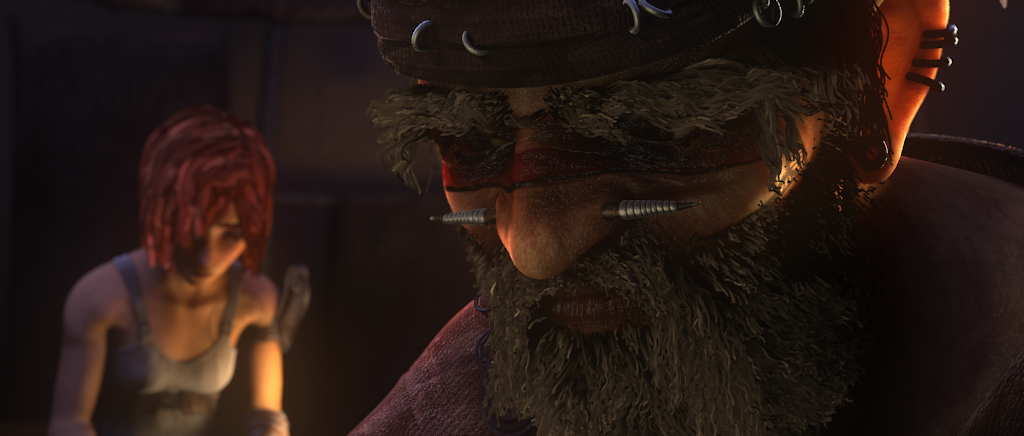} &
    \begin{overpic}[width=0.24\textwidth, frame]{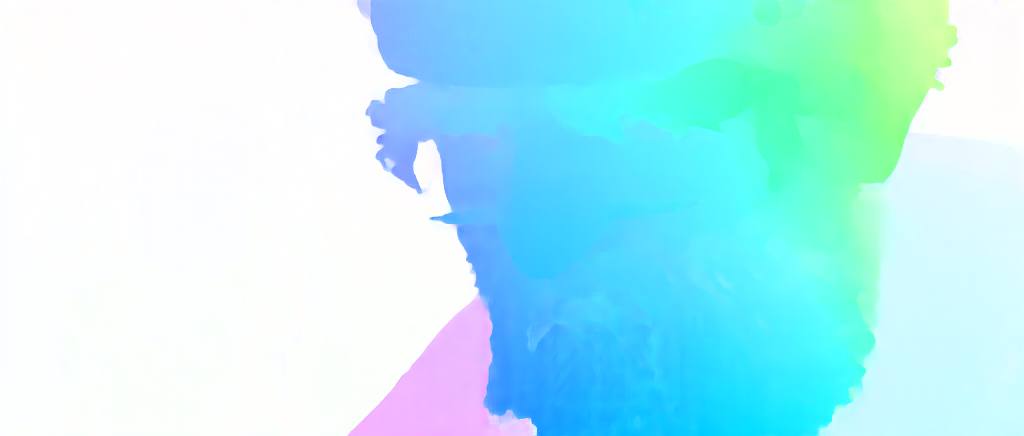}
    \put (3,35) {\textcolor{purple}{\textbf{\texttt{EPE: 0.402}}}}
    \end{overpic} &
    \begin{overpic}[width=0.24\textwidth, frame]{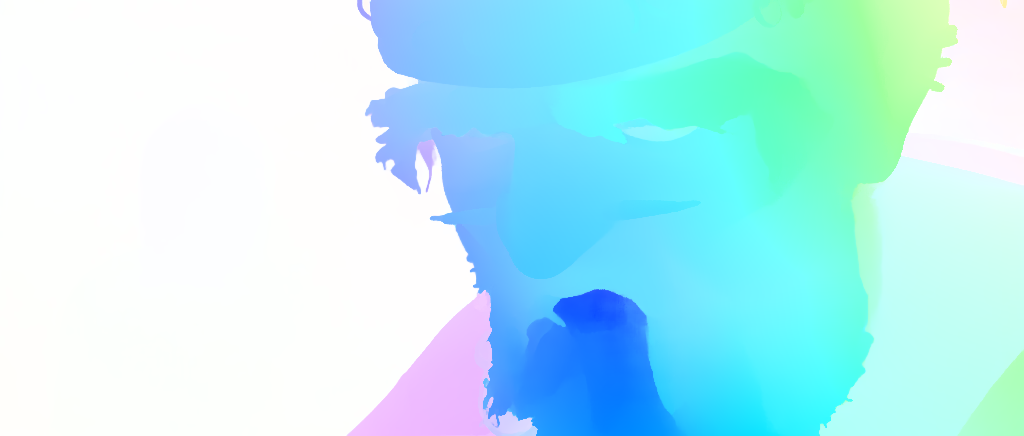} \put (3,35) {\textcolor{purple}{\textbf{\texttt{EPE: 0.221}}}} \end{overpic} &
    \begin{overpic}[width=0.24\textwidth, frame]{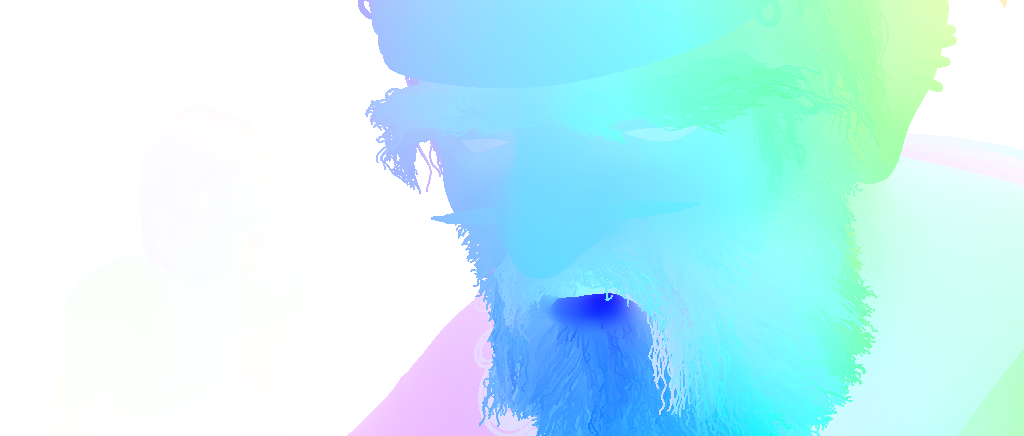}
    \end{overpic}
    \\

    \includegraphics[width=0.24\textwidth, frame]{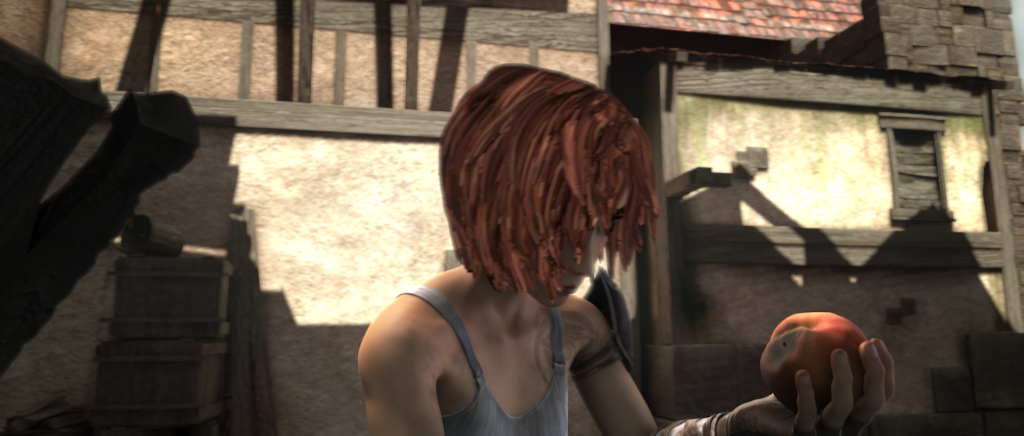} &
    \begin{overpic}[width=0.24\textwidth, frame]{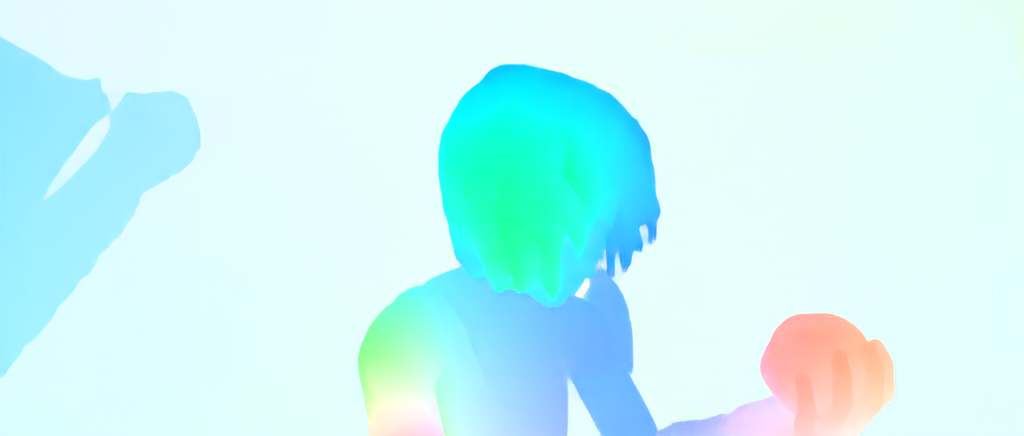} \put (3,35) {\textcolor{purple}{\textbf{\texttt{EPE: 0.163}}}}\end{overpic} &
    \begin{overpic}[width=0.24\textwidth, frame]{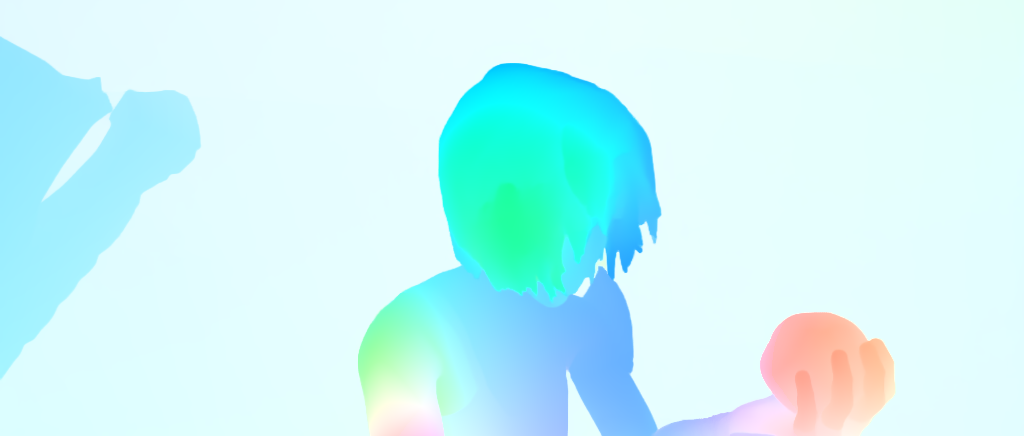} \put (3,35) {\textcolor{purple}{\textbf{\texttt{EPE: 0.128}}}} \end{overpic} &
    \begin{overpic}[width=0.24\textwidth, frame]{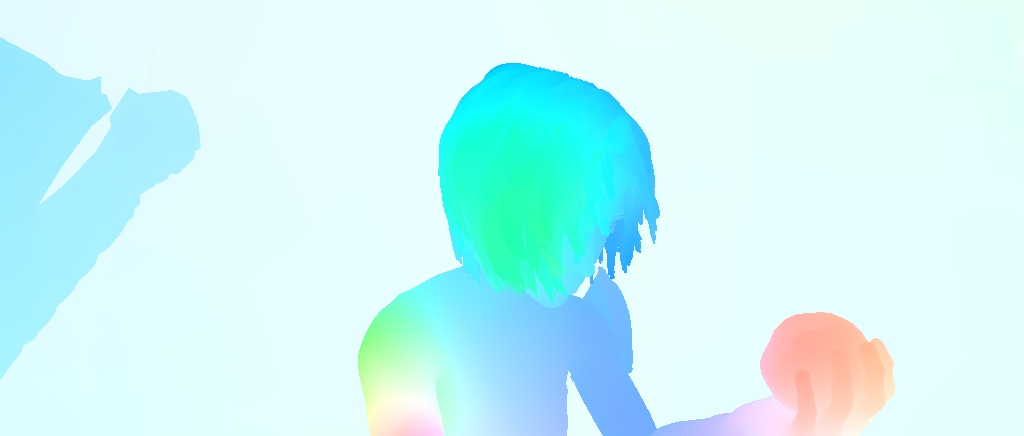}
    \end{overpic}
    \\

    \includegraphics[width=0.24\textwidth, frame]{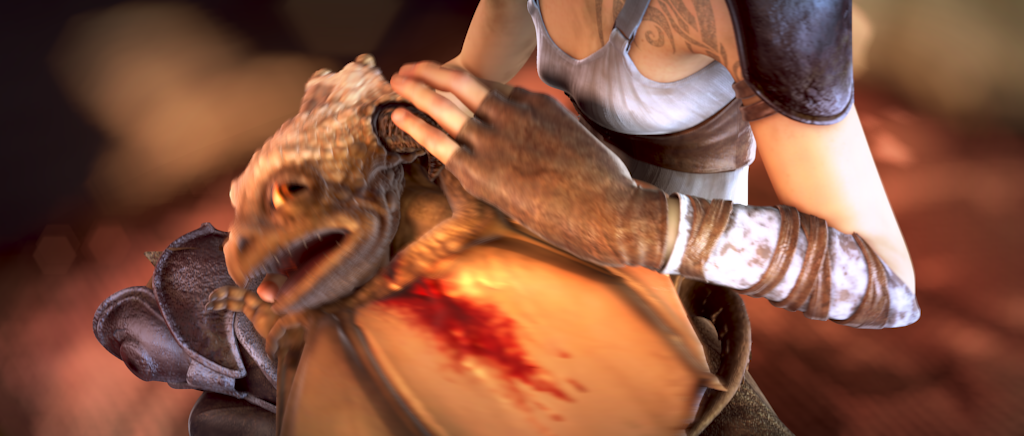} &
    \begin{overpic}[width=0.24\textwidth, frame]{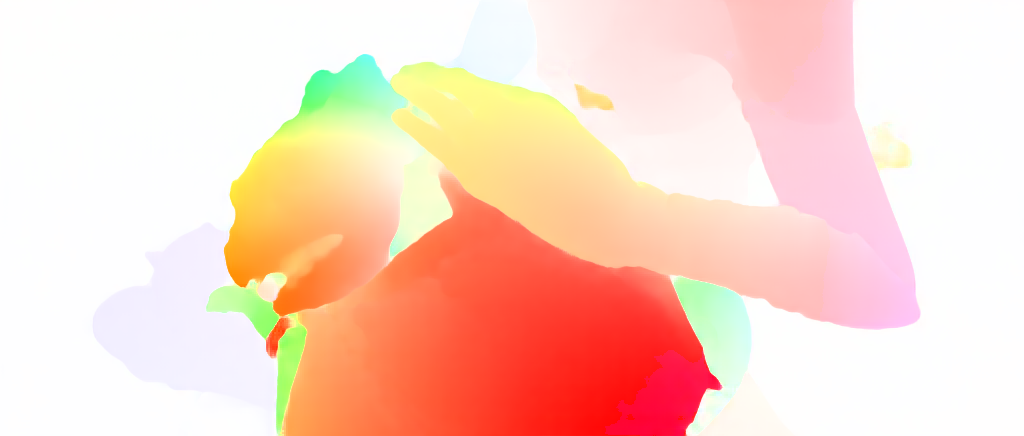} \put (3,35) {\textcolor{purple}{\textbf{\texttt{EPE: 0.478}}}}\end{overpic} &
    \begin{overpic}[width=0.24\textwidth, frame]{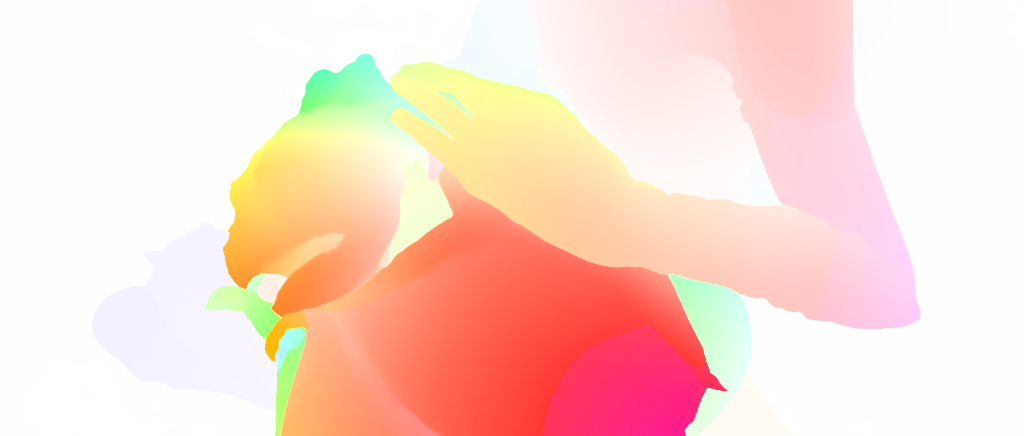} \put (3,35) {\textcolor{purple}{\textbf{\texttt{EPE: 0.305}}}} \end{overpic} &
    \begin{overpic}[width=0.24\textwidth, frame]{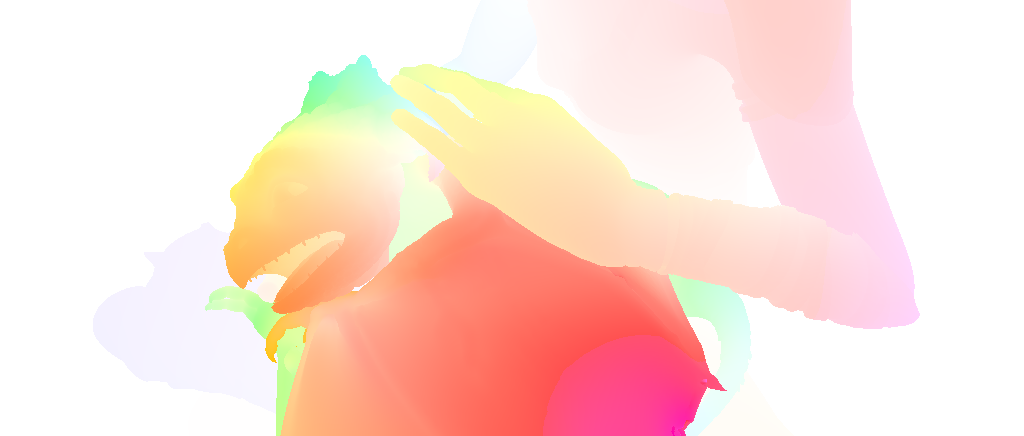}
    \end{overpic}
    \\

    \small Image & \small WAFT-DAv2-a1 \cite{wang2025waft} & \small \bf \net{} (XL) & \small Ground-truth \\

    \end{tabular}\vspace{0.3cm}
    \caption{\textbf{Qualitative Results on Sintel \cite{sintel} Training Set.} From left to right: first frame, flow by WAFT-DAv2-a1 and \net{} (XL), ground-truth flow. The top four rows are from the clean split, and the bottom four rows are from the final split.}
    \label{fig:sintel}
\end{figure*}

\begin{table}[t]
    \centering
    \resizebox{\linewidth}{!}{
    \renewcommand{\tabcolsep}{7pt}
    \begin{tabular}{cc}

    \begin{tabular}{lrrrrrr}
    \toprule
    \multirow{1}{*}{Method} &\multicolumn{2}{c}{Spring (train)} \\
    \cmidrule(l{0.5ex}r{0.5ex}){2-3} 
    & 
    1px$\downarrow$ & EPE$\downarrow$ 
    \\ 
        \midrule 
        MatchFlow(G)~\cite{dong2023rethinking}  & 4.504 & 0.407 \\ 
        Flowformer$++$\cite{shi2023flowformer++} & 4.482 & 0.447 \\ 
        MS-RAFT+~\cite{jahedi2024ms}  & \trd {3.577} & {0.397} \\ 
        RAFT~\cite{teed2020raft}  & 4.788 & 0.448 \\ 
        RPKNet~\cite{morimitsu2024recurrent}  & 4.472 & 0.416 \\ 
        DIP~\cite{zheng2022dip}  & 4.273 & 0.463 \\ 
        SKFlow~\cite{sun2022skflow}  & 4.521 & 0.408 \\ 
        GMFlow~\cite{xu2022gmflow}  & 29.49 & 0.930 \\ 
        GMFlow$+$~\cite{xu2023unifying}  & 4.292 & 0.433 \\ 
        Flowformer~\cite{huang2022flowformer}  & 4.508 & 0.470 & \\ 
        {SEA-RAFT (S)} \cite{wang2024sea} & 4.161 & 0.410 \\
        {SEA-RAFT (M)} \cite{wang2024sea} & {3.888} & {0.406} \\ 
        {SEA-RAFT (L)} \cite{wang2024sea} &  3.842 & 0.426 \\
        FlowSeek (T) \cite{Poggi_2025_ICCV} & 4.111 & 0.410 \\
        FlowSeek (S) \cite{Poggi_2025_ICCV} & 4.058 & 0.406 \\
        FlowSeek (M) \cite{Poggi_2025_ICCV} & 3.941 & 0.419 \\
        FlowSeek (L) \cite{Poggi_2025_ICCV} &  3.838 & 0.402 \\
        WAFT-DAv2-a1 \cite{wang2025waft} & 3.811 & 0.433 \\ 
        WAFT-Twins-a2 \cite{wang2025waft} & 5.611 & 0.557 \\
        WAFT-DAv2-a2 \cite{wang2025waft} & 5.340 & 0.487 \\
        WAFT-DINOv3-a2 \cite{wang2025waft} & 4.006 & 0.458 \\
        \hdashline
        \bf \net{} (S) & 3.937 & \trd 0.391 \\
        \bf \net{} (M) & 4.017 & 0.398 \\
        \bf \net{} (L) & \fst 3.450 & \snd 0.377 \\
        \bf \net{} (XL) & \snd 3.474 & \fst 0.371 \\
    \bottomrule
    \end{tabular} 
    &
    \begin{tabular}{l r rrrr r rrrr r rrrr r rrrr}
    \toprule
    \multirow{2}{*}{Method} && \multicolumn{4}{c}{LayeredFlow (All pixels)} 
    \\
    \cmidrule{2-21}
    && EPE$\downarrow$ & 1px$\downarrow$ & 3px$\downarrow$ & 5px$\downarrow$ 
    \\
    \midrule
    FlowNet-C \cite{chairs} && 9.71 & 89.07 & 61.51 & 43.93 \\
    FlowNet2 \cite{ilg2017flownet} && 10.07 & 77.56 & 54.22 & 42.13 \\
    PWC-Net \cite{sun2018pwc} && 9.49 & 74.93 & 50.47 & 39.05 \\
    GMA \cite{jiang2021learning} && 9.77 & 72.46 & 46.93 & 36.97 \\
    SKFlow \cite{sun2022skflow} && 9.86 & 72.02 & 47.44 & 36.88 \\
    CRAFT \cite{sui2022craft} && 10.36 & 72.34 & 47.54 & 37.00 \\
    GMFlow \cite{xu2022gmflow} &&  9.09 & 81.99 & 51.79 & 37.75 \\
    GMFlow+ \cite{xu2023unifying} && 9.46 & 82.71 & 53.14 & 39.70 \\
    FlowFormer \cite{huang2022flowformer} && 10.20 & 73.59 & 48.97 & 38.56\\ 
    RAFT \cite{teed2020raft} && 9.38 & 71.98 & 46.46 & 36.15 \\
    SEA-RAFT (S) \cite{wang2024sea} && 10.05 & 71.48 & 46.90 & 36.32 \\
    SEA-RAFT (M) \cite{wang2024sea} && 10.17 & 69.73 & 45.94 & 34.78 \\
    SEA-RAFT (L) \cite{wang2024sea} && 10.99 & 69.46 & 45.59 & 34.78 \\
    FlowSeek (T) \cite{Poggi_2025_ICCV} && 
     9.09  & 70.82  & 43.74 & 32.36 \\
    FlowSeek (S) \cite{Poggi_2025_ICCV} && 
    9.16 & 69.99 &  43.67 & \trd 31.90 \\
    FlowSeek (M) \cite{Poggi_2025_ICCV} && 
    \trd 8.30 & 68.85 & \trd 41.81 & 32.09 \\
    FlowSeek (L) \cite{Poggi_2025_ICCV} && 
    \trd 8.30 &  68.98 & \snd 41.49 & \snd 31.64 \\

    {WAFT-DAv2-a1} \cite{wang2025waft} && 10.37 & \trd 68.23 & 43.73 & 33.38 \\ 
    {WAFT-Twins-a2} \cite{wang2025waft} && 12.35 & 70.93 & 47.11 & 36.51 \\
    {WAFT-DAv2-a2} \cite{wang2025waft} && 10.75 & 73.06 & 51.40 & 40.85 \\
    {WAFT-DINOv3-a2} \cite{wang2025waft} && 9.30 & 73.05 & 48.25 & 38.22 \\ 
    \hdashline
    \bf \net{} (S) && 9.56 & 70.41 & 45.45 & 35.62 \\
    \bf \net{} (M) && 10.07 & 69.83 & 46.47 & 36.08 \\
    \bf \net{} (L) && \snd 7.93 & \snd 66.55 & 43.20 & 33.09 \\
    \bf \net{} (XL) && \fst 7.26 & \fst 65.62 & \fst 38.78 & \fst 28.23 \\
    \bottomrule
  \end{tabular}
    \end{tabular}
    }\vspace{0.3cm}
    \caption{\textbf{Generalization on Spring and LayeredFlow.}}
    \label{tab:spring_layered_zero_shot}
    
\end{table}

\begin{table}[t]
    \centering
    \renewcommand{\tabcolsep}{8pt}
    \resizebox{0.9\linewidth}{!}{
    \begin{tabular}{llcccc}
    \toprule
        {Experiment} & Configuration & EPE$\downarrow$ & 3px$\downarrow$ & \#GMACS & latency (ms)\\
    \midrule

    \multirow{4}{*}{\textcolor{blue}{Model Size}} 
        & (S)  & 0.617 & 2.295 & 344 & 99 \\
        & (M)  & 0.595 & 2.209 & 764 & 120 \\
        & (L)  & 0.499 & 1.635 & 1412 & 146 \\
        & (XL) & 0.447 & 1.328 & 3092 & 216 \\
    \midrule

    \multirow{2}{*}{\textcolor[RGB]{124, 173, 230}{Transformer Architecture}}
        & GMFlow & 0.652 & 2.457 & 397 & 155 \\
        & \underline{MRT} & 0.617 & 2.295 & 344 & 99 \\
    \midrule
    
    \multirow{2}{*}{\textcolor[RGB]{125,62,150}{Flow Initialization}}
        & Softmax-based & 0.631 & 2.384 & 344 & 47 \\
        & \underline{Optimal Transport} & 0.617 & 2.295 & 344 & 99 \\
    \midrule

    \multirow{3}{*}{\textcolor{orange}{Refinement Guidance}}
        & w/o Conf. \& w/o Occ. & 0.708 & 2.541 & 343 & 99 \\
        & w/o Conf. \& w Occ. & 0.638 & 2.370 & 344 & 99 \\
        & \underline{w/ Conf. \& w/ Occ.} & 0.617 & 2.295  & 344 & 99 \\
        \midrule

    \multirow{4}{*}{\textcolor{CadetBlue}{Refinement Steps}}
        & 0 steps & 1.622 & 6.388 & 148 & 78 \\
        & 1 step  & 0.687 & 2.732 & 213 & 85 \\
        & \underline{3 steps} & 0.617 & 2.295 & 344 & 99 \\
        & 5 steps & 0.620 & 2.287 & 474 & 113 \\
    \midrule

    \multirow{2}{*}{\textcolor{teal}{Axis-wise Refinement}}
        & Coupled & 0.897 & 3.625 & 227 & 87 \\
        & \underline{Axis-wise}  & 0.617 & 2.295 & 344 & 99 \\
    \midrule

    \multirow{2}{*}{\textcolor{purple}{Loss Function}}
        & RAFT-style & 0.639 & 2.411 & 344 & 99 \\
        & \underline{Flow \& Conf. \& Occ.} & 0.617 & 2.295 & 344 & 99 \\

    \bottomrule
    \end{tabular}
    }
    \vspace{0.3cm}
    \caption{\textbf{Ablation study on FlyingChairs validation set.} We ablate \textcolor{blue}{model size}, \textcolor[RGB]{124, 173, 230}{tranformer architecture}, flow initialization using \textcolor[RGB]{125,62,150}{ optimal transport},  conditioning refinement on \textcolor{orange}{ confidence and occlusion maps}, number of \textcolor{CadetBlue}{refinement steps}, \textcolor{teal}{axis-wise refinement}, and \textcolor{purple}{loss function}. Configurations used in our final model are underlined. Latencies are measured on an NVIDIA L40S GPU.}
    \label{tab:ablations}
\end{table}

\begin{figure*}[t]
    \centering
    \renewcommand{\tabcolsep}{0.5pt}
    \begin{tabular}{cccc}
    \rotatebox[origin=l]{90}{\scriptsize{\quad \quad Image 1}} &
    \hspace{0.1cm}\includegraphics[width=0.315\textwidth, frame]{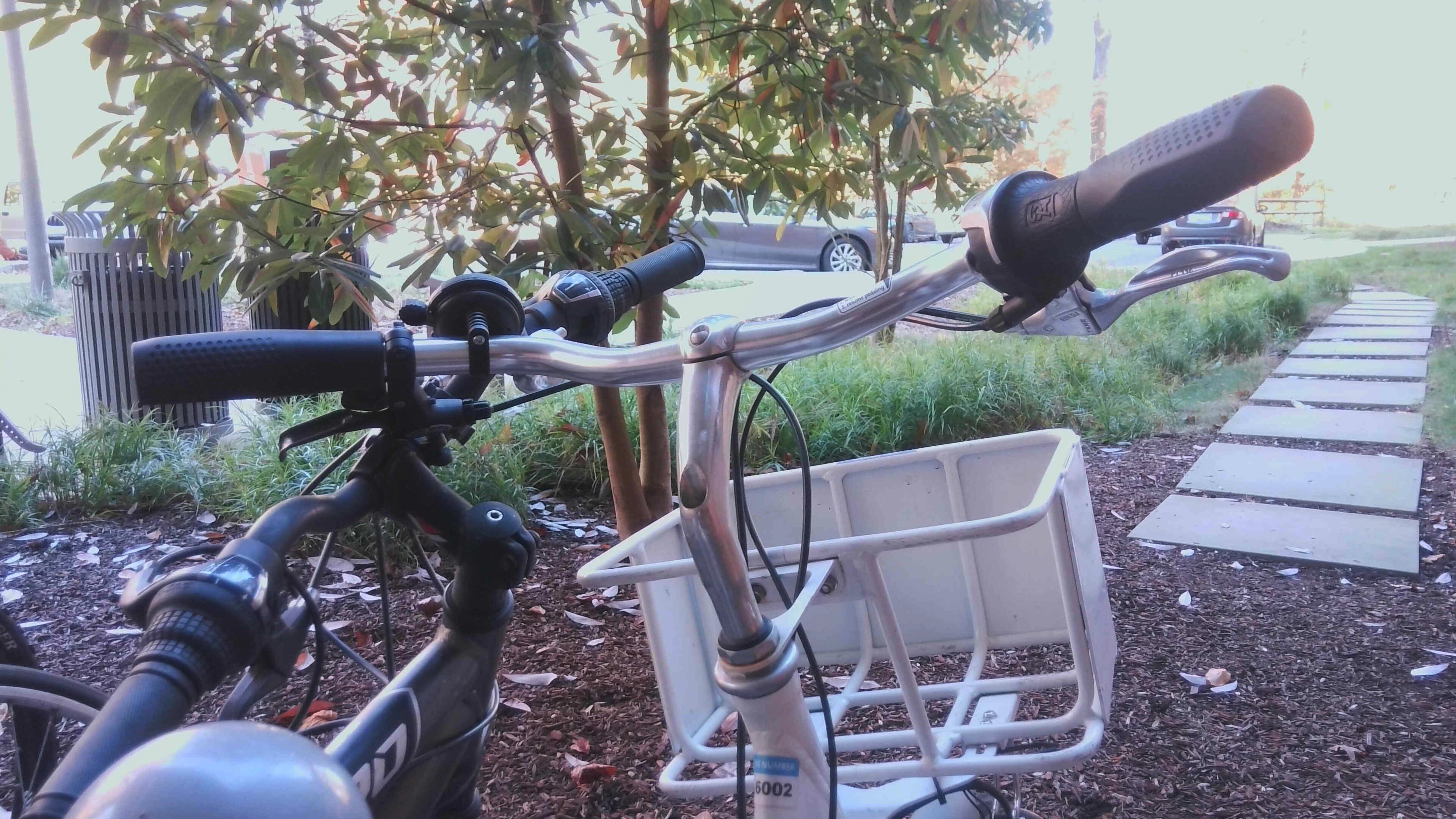} &
     \hspace{0.1cm}\includegraphics[width=0.315\textwidth, frame]{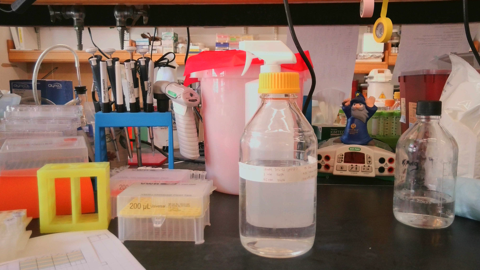}
     &
     \hspace{0.1cm}\includegraphics[width=0.315\textwidth, frame]{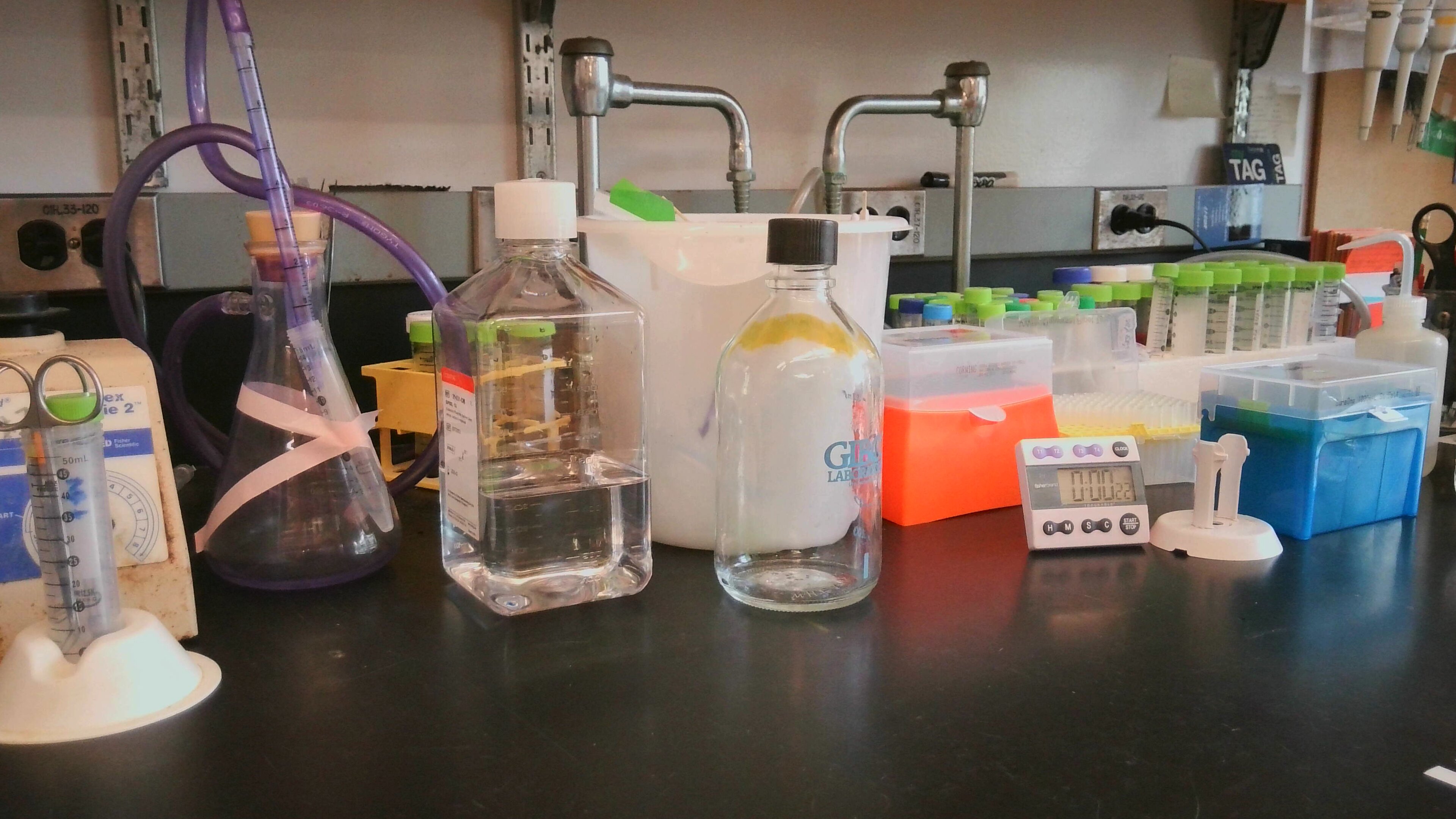}
    \\
    \rotatebox[origin=l]{90}{\scriptsize{\quad SEA-RAFT~\cite{wang2024sea}}} &
    \begin{overpic}[width=0.315\textwidth, frame]{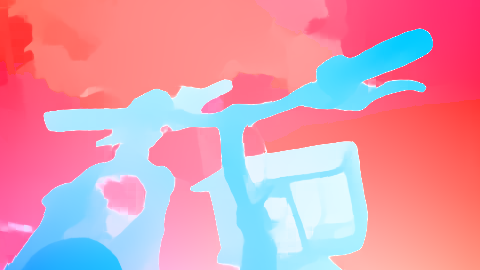} \put (3,48) {\textcolor{purple}{\textbf{\texttt{}}}}
    \end{overpic} &
     \begin{overpic}[width=0.315\textwidth, frame]{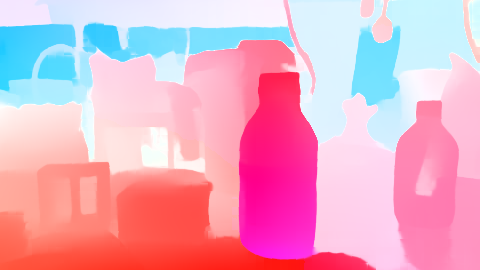} \put (3,48) {\textcolor{purple}{\textbf{\texttt{}}}}\end{overpic}
     &
     \begin{overpic}[width=0.315\textwidth, frame]{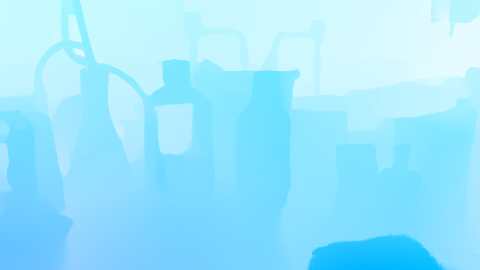} \put (3,48) {\textcolor{purple}{\textbf{\texttt{}}}}
     \end{overpic}
     \\
     \rotatebox[origin=l]{90}{\quad \scriptsize{WAFT~\cite{wang2025waft}}} &
    \begin{overpic}[width=0.315\textwidth, frame]{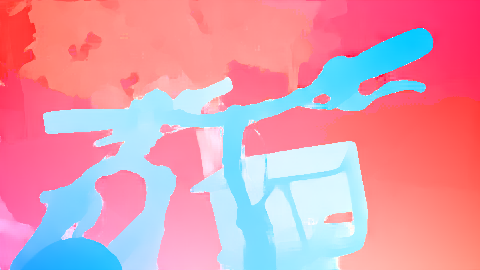} 
    \put (3,48) {\textcolor{purple}{\textbf{\texttt{}}}}
    \end{overpic} &
     \begin{overpic}[width=0.315\textwidth, frame]{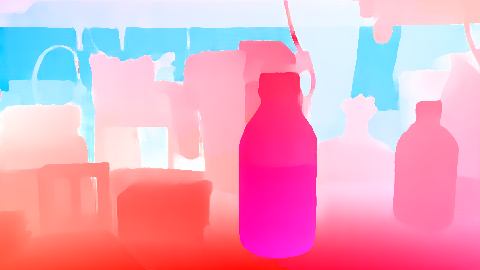} \put (3,48) {\textcolor{purple}{\textbf{\texttt{}}}}\end{overpic}
     &
     \begin{overpic}[width=0.315\textwidth, frame]{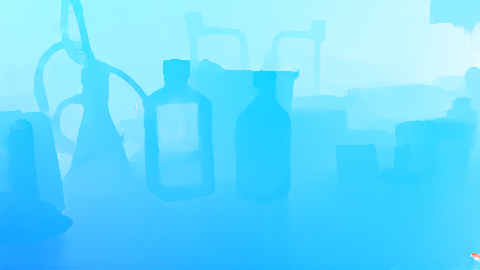} \put (3,48) {\textcolor{purple}{\textbf{\texttt{}}}}\end{overpic}

     \\
     \rotatebox[origin=l]{90}{\quad \scriptsize{FlowSeek~\cite{Poggi_2025_ICCV}}} &
    \begin{overpic}[width=0.315\textwidth, frame]{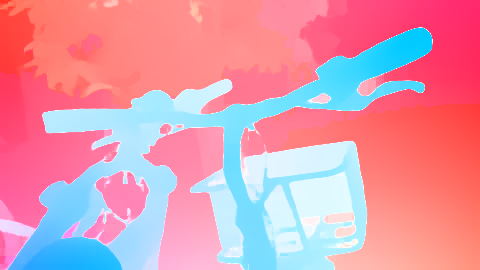} 
    \put (3,48) {\textcolor{purple}{\textbf{\texttt{}}}}
    \end{overpic} &
     \begin{overpic}[width=0.315\textwidth, frame]{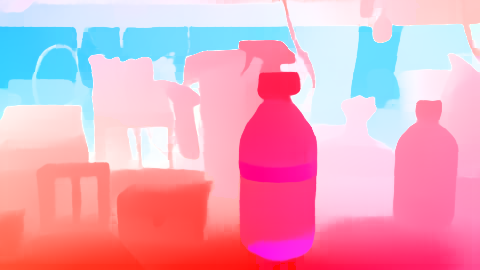} \put (3,48) {\textcolor{purple}{\textbf{\texttt{}}}}\end{overpic}
     &
     \begin{overpic}[width=0.315\textwidth, frame]{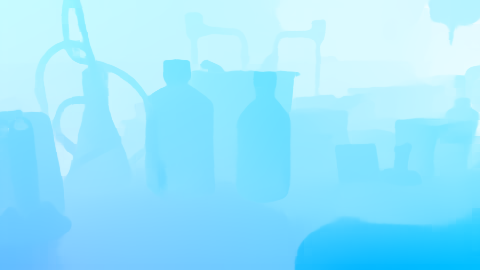} \put (3,48) {\textcolor{purple}{\textbf{\texttt{}}}}\end{overpic}
    \\
    \rotatebox[origin=l]{90}{\quad \scriptsize{\net{} (XL)}} &
    \begin{overpic}[width=0.315\textwidth, frame]{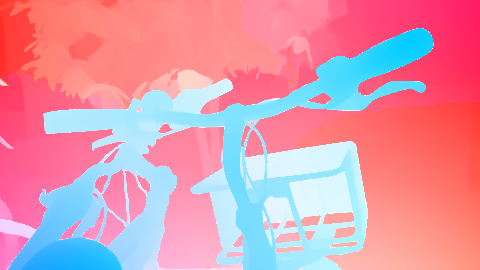} \put (3,48) {\textcolor{purple}{\textbf{\texttt{}}}} \end{overpic} &
     \begin{overpic}[width=0.315\textwidth, frame]{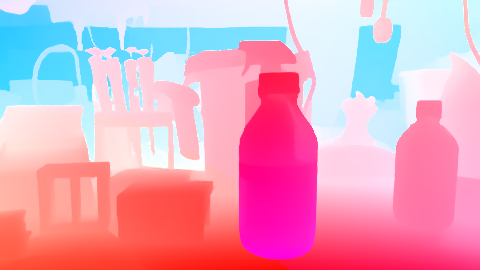}  \put (3,48) {\textcolor{purple}{\textbf{\texttt{}}}} \end{overpic}
     &
     \begin{overpic}[width=0.315\textwidth, frame]{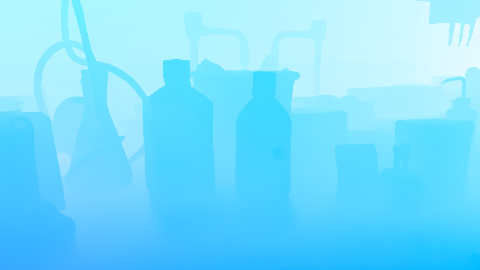}  \put (3,48) {\textcolor{purple}{\textbf{\texttt{}}}} \end{overpic}

    \end{tabular}
    \vspace{0.1cm}
    \caption{\textbf{Qualitative Results of Zero-Shot Generalization on LayeredFlow~\cite{wen2024layeredflow}.} From top to bottom: first frame, flow by  SEA-RAFT (L)~\cite{wang2024sea}, WAFT-DAv2-a1~\cite{wang2025waft}, FlowSeek (L)~\cite{Poggi_2025_ICCV}, and \net{} (XL).}
    \label{fig:layered}
\end{figure*}

\subsection{Ablation Study}
\label{sec:ablation}

We analyze the impact of our key design choices for \net{}. All variants are trained on FlyingChairs \cite{chairs} and evaluated on its validation split. Results are summarized in \tabref{tab:ablations}. Unless otherwise stated, we use the (S) variant as our default configuration.

\boldparagraph{\textcolor{blue}{Model Size}}
By varying the feature channel dimensions and the number of transformer blocks (see \secref{sec:arch}), we define four model variants. 
As expected, increasing model capacity consistently improves performance, demonstrating the scalability of our architecture.

\boldparagraph{\textcolor[RGB]{124, 173, 230}{Multi-Resolution Transformer}} We evaluate the effect of our Multi-Resolution Transformer (MRT) by replacing it with the transformer architecture from GMFlow~\cite{xu2022gmflow}. MRT delivers superior results while using fewer parameters and offering substantially lower runtime, highlighting the benefit of our hierarchical design.

\boldparagraph{\textcolor[RGB]{125,62,150}{Flow Initialization}} We study the impact of optimal transport-based initialization by replacing it with the softmax-based scheme used in GMFlow~\cite{xu2022gmflow}. Optimal transport initialization yields improved performance, indicating the importance of a more structured matching prior.

\boldparagraph{\textcolor{orange}{Refinement Guidance}} We analyze the role of conditioning the refinement module on confidence and occlusion maps. Removing these inputs leads to consistently degraded performance, underscoring the importance of confidence and occlusion-guided refinement. %

\boldparagraph{\textcolor{CadetBlue}{Number of Refinement Steps}} %
We study the effect of varying the number of refinement iterations. Omitting refinement leads to substantially lower performance, while increasing the number of iterations improves accuracy up to three steps. Beyond this point, additional iterations yield little benefit and can even slightly degrade performance. Balancing accuracy and computational cost, we adopt three refinement steps as the optimal configuration.%

\boldparagraph{\textcolor{teal}{Axis-wise Flow Refinement}}
We investigate two strategies for refining the optical flow components. In our default axis-wise refinement, the horizontal (\textit{u}) and vertical (\textit{v}) components are refined independently, each with its own refinement head and update step as mentioned in Sec. \ref{sec:refine}. In contrast, coupled refinement predicts a two-channel update and jointly refines both components within a single head.
We observe that the coupled design leads to degraded performance. Axis-wise refinement, on the other hand, enables better accuracy, indicating that decoupling the two motion directions is crucial for a more effective refinement.

\boldparagraph{\textcolor{purple}{Loss Function}}
We evaluate the contribution of our loss formulation by comparing it to a baseline that uses exponentially increasing weights, similar to RAFT~\cite{teed2020raft}. In that baseline, confidence and occlusion maps are not directly supervised, preventing the flow predictions from effectively leveraging these signals. Our proposed loss yields clear performance improvements, highlighting the benefit of explicit supervision on confidence and occlusion maps in addition to flow.

\section{Conclusion}

We presented \net{}, a global matching network for optical flow that leverages optimal transport for robust initialization and iterative refinement to propagate reliable motion cues into ambiguous regions. %
Results across Sintel, KITTI, Spring, and LayeredFlow validate fine-tuned and zero-shot performance, while multiple model variants offer flexibility across computational budgets.

Future work will focus on improving FlowIt capabilities to efficiently process high-resolution images, as well as extending it to process multiple consecutive frames and thus perform multi-frame optical flow estimation.

\maketitlesupp

\setcounter{table}{0}
\renewcommand{\thetable}{S\arabic{table}}
\setcounter{figure}{0}
\renewcommand{\thefigure}{S\arabic{figure}}
\setcounter{equation}{0}
This document provides additional qualitative and quantitative results. %
Figure~\ref{fig:supp_kitti_train} provides qualitative results of zero-shot cross-dataset generalization results on the KITTI training set~\cite{kitti}, while
Figures~\ref{fig:supp_sintel_test_clean} and~\ref{fig:supp_sintel_test_final} present qualitative results on the clean and final splits of the Sintel test set~\cite{sintel}, respectively. Additionally, Figure~\ref{fig:supp_spring_new} shows zero-shot cross-dataset generalization results on the Spring dataset~\cite{mehl2023spring}. Finally, Table~\ref{tab:supp_big_sintel_kitti_benchmark} extends upon Table~1 of the main paper
by providing additional details regarding the training data used by each method. Additionally, it provides the results of \net{} without TartanAir~\cite{wang2020tartanair} pretraining. For clarity, the top three 2-frame methods trained with and without external data are highlighted separately.

\begin{figure*}[!h]
    \centering
    \renewcommand{\tabcolsep}{0.5pt}
    \begin{tabular}{cccc}
    \rotatebox[origin=l]{90}{\tiny{\quad Image 1}} &
    \hspace{0.1cm}\includegraphics[width=0.32\textwidth]{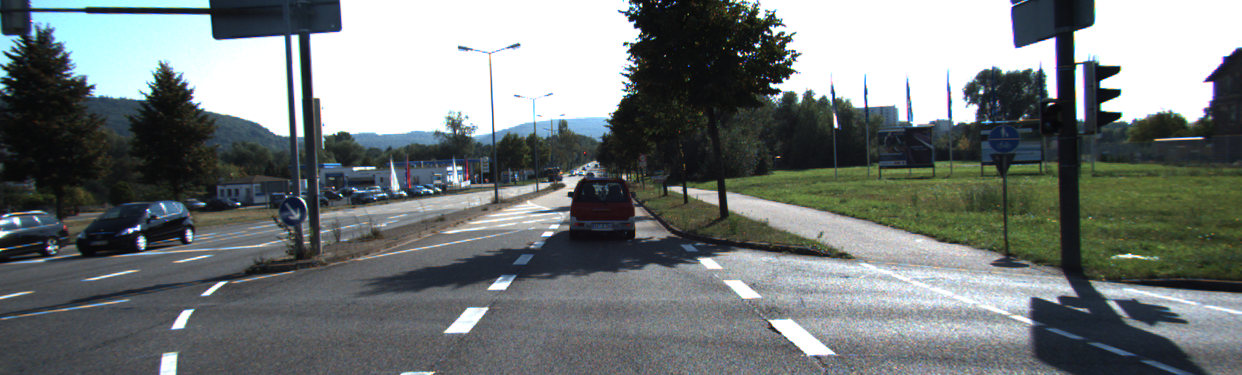} &
     \hspace{0.1cm}\includegraphics[width=0.32\textwidth]{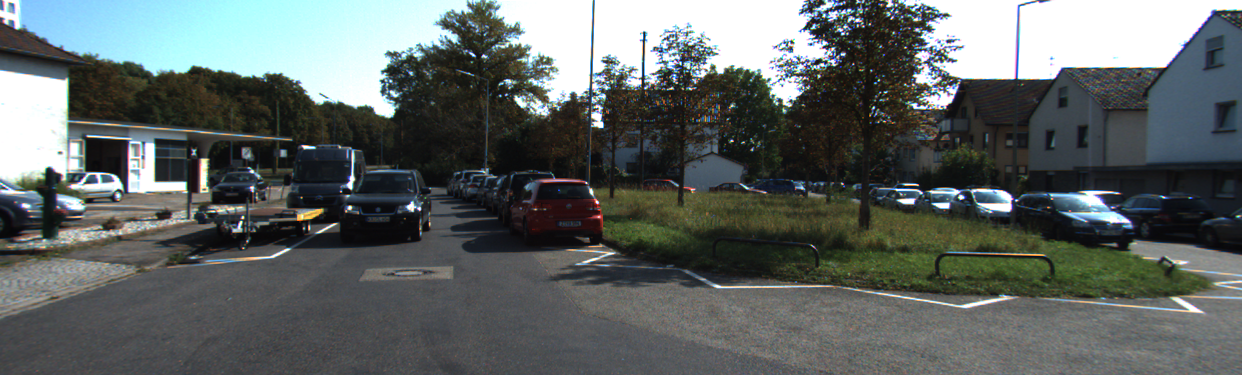}
     &
     \hspace{0.1cm}\includegraphics[width=0.32\textwidth]{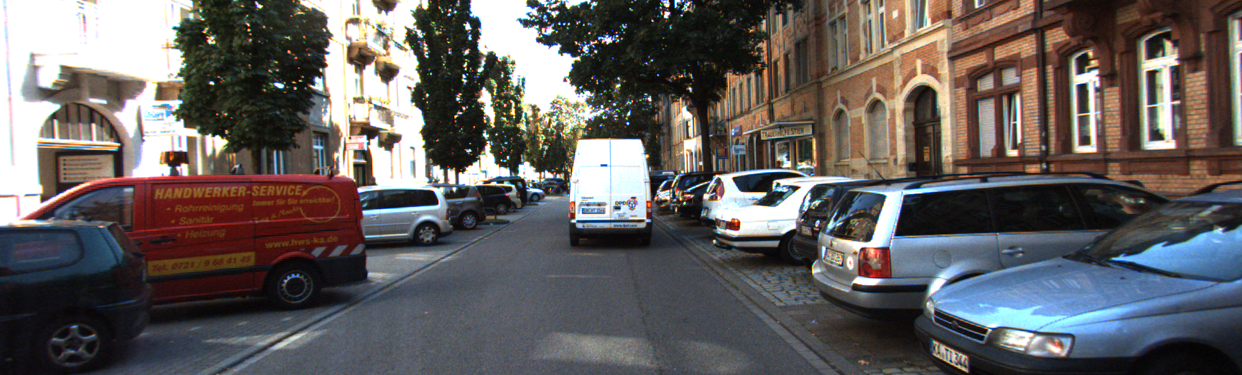}
    \\ 
    \rotatebox[origin=l]{90}{\tiny{SEA-RAFT \cite{wang2024sea}}} &
    \begin{overpic}[width=0.32\textwidth]{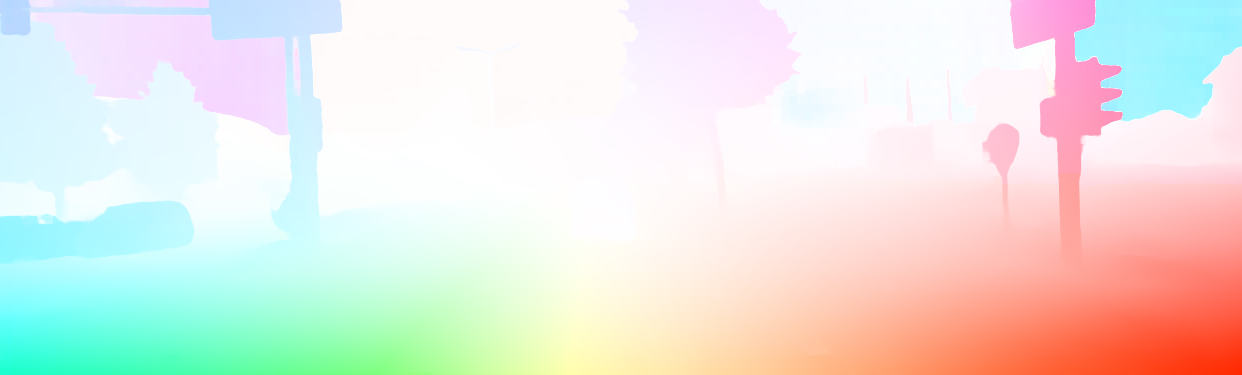} \put (3,24) {\textcolor{purple}{\textbf{\texttt{EPE: 0.898}}}} \end{overpic} &
     \begin{overpic}[width=0.32\textwidth]{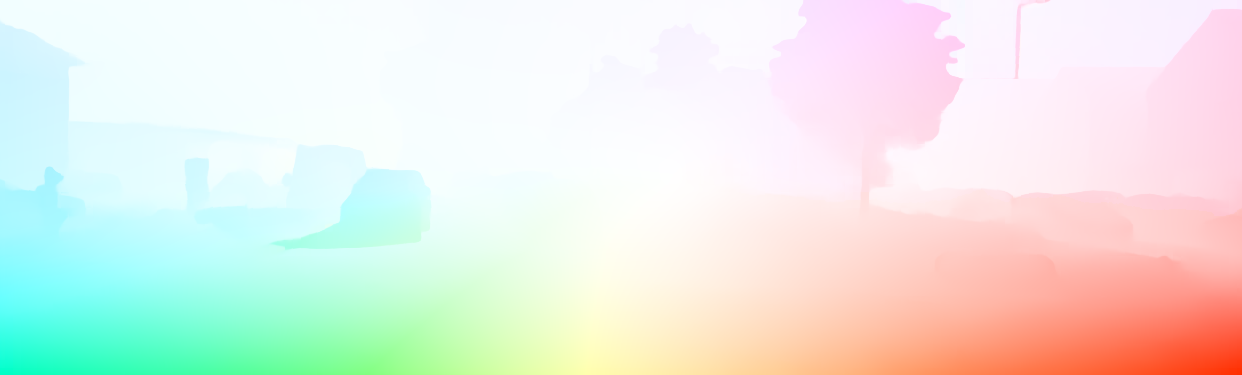}  \put (3,24) {\textcolor{purple}{\textbf{\texttt{EPE: 0.550}}}} \end{overpic}
     &
     \begin{overpic}[width=0.32\textwidth]{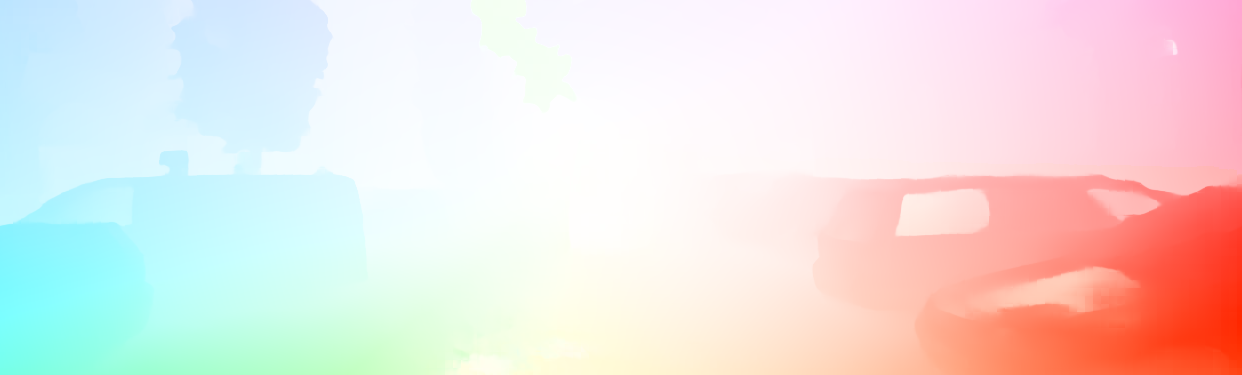}  \put (3,24) {\textcolor{purple}{\textbf{\texttt{EPE: 1.675}}}} \end{overpic}
    \\
    \rotatebox[origin=l]{90}{\tiny{ FlowSeek~\cite{Poggi_2025_ICCV}}} &
    \begin{overpic}[width=0.32\textwidth]{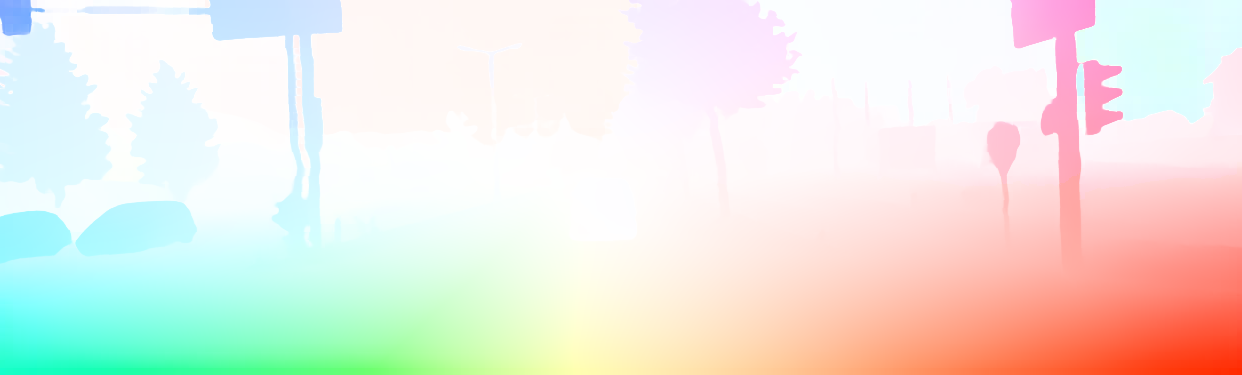} \put (3,24) {\textcolor{purple}{\textbf{\texttt{EPE: 1.013}}}}
    \end{overpic} &
     \begin{overpic}[width=0.32\textwidth]{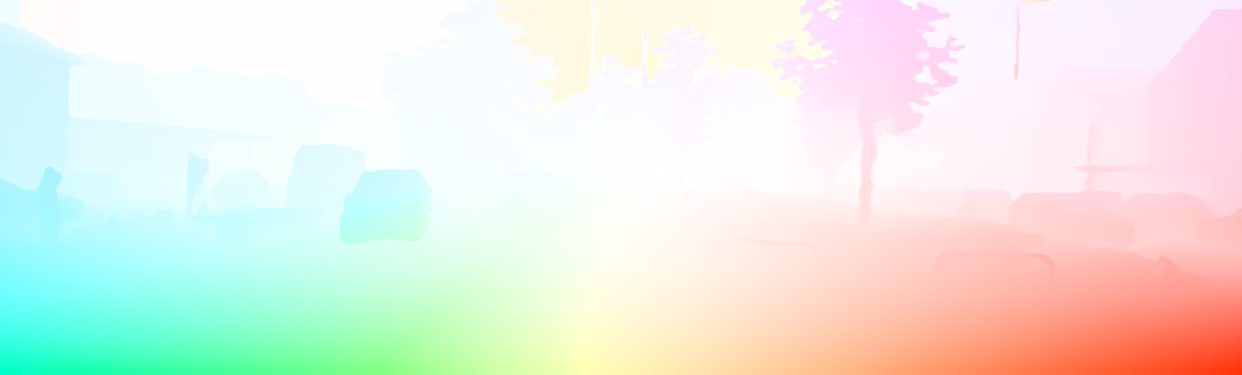} \put (3,24) {\textcolor{purple}{\textbf{\texttt{EPE: 0.480}}}}
     \end{overpic}
     &
     \begin{overpic}[width=0.32\textwidth]{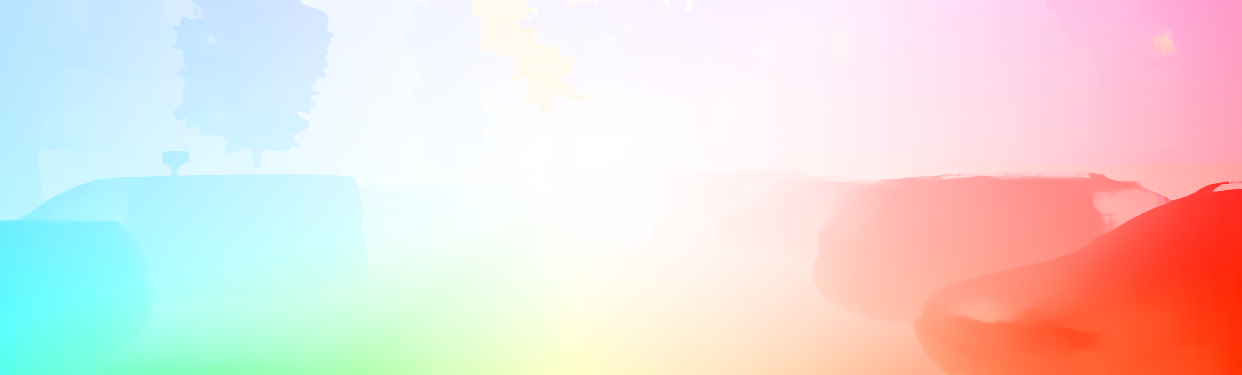} \put (3,24) {\textcolor{purple}{\textbf{\texttt{EPE: 1.395}}}}
     \end{overpic}
    \\
    \rotatebox[origin=l]{90}{\tiny{WAFT~\cite{wang2025waft}}} &
    \begin{overpic}[width=0.32\textwidth]{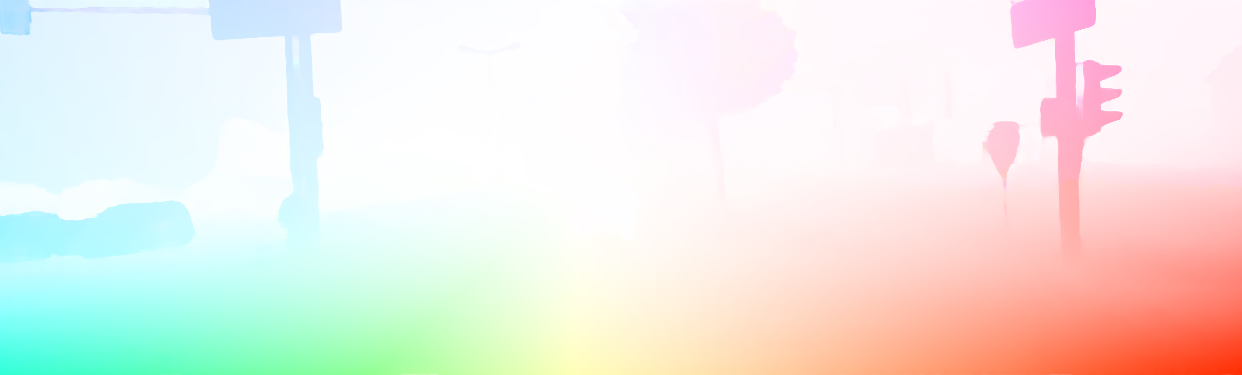} 
    \put (3,24) {\textcolor{purple}{\textbf{\texttt{EPE: 0.671}}}}
    \end{overpic} &
     \begin{overpic}[width=0.32\textwidth]{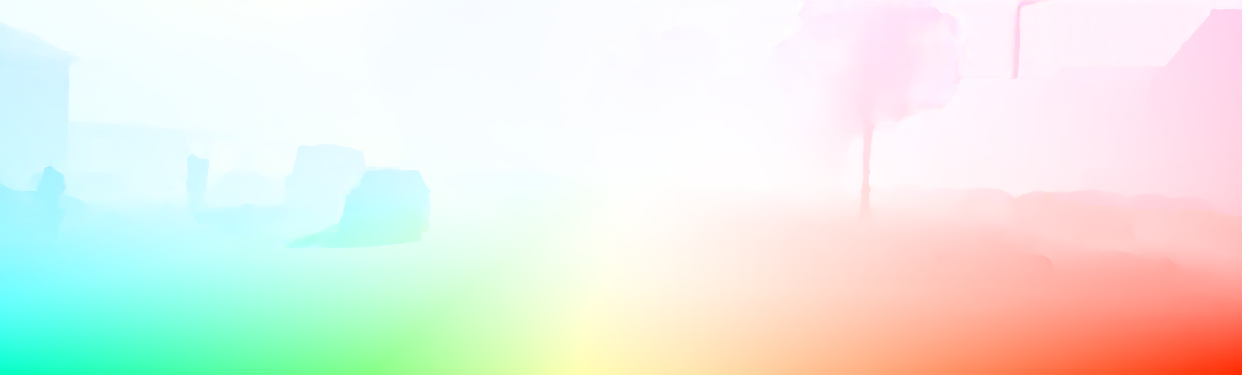} \put (3,24) {\textcolor{purple}{\textbf{\texttt{EPE: 0.514}}}}\end{overpic}
     &
     \begin{overpic}[width=0.32\textwidth]{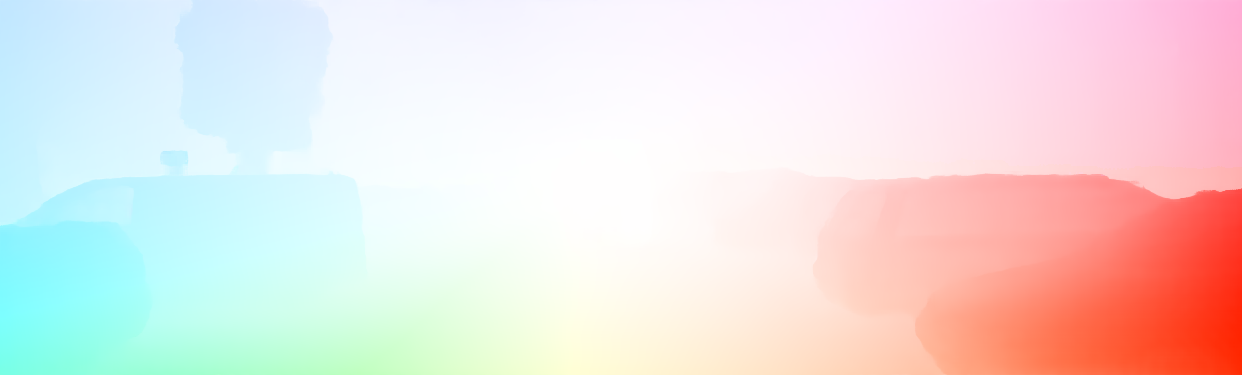} \put (3,24) {\textcolor{purple}{\textbf{\texttt{EPE: 0.764}}}}\end{overpic}
    \\
    \rotatebox[origin=l]{90}{\tiny{\net{} (XL)}} &
    \begin{overpic}[width=0.32\textwidth]{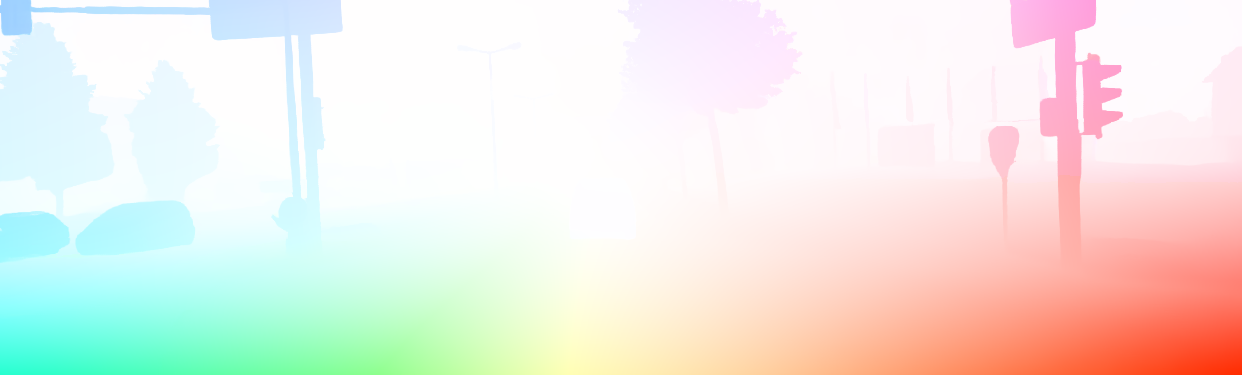} \put (3,24) {\textcolor{purple}{\textbf{\texttt{EPE: 0.596}}}} \end{overpic} &
     \begin{overpic}[width=0.32\textwidth]{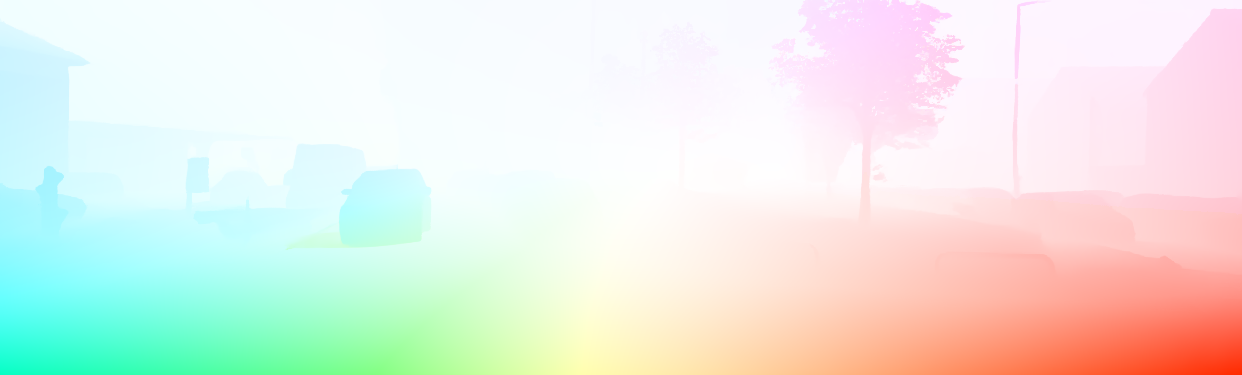}  \put (3,24) {\textcolor{purple}{\textbf{\texttt{EPE: 0.399}}}} \end{overpic}
     &
     \begin{overpic}[width=0.32\textwidth]{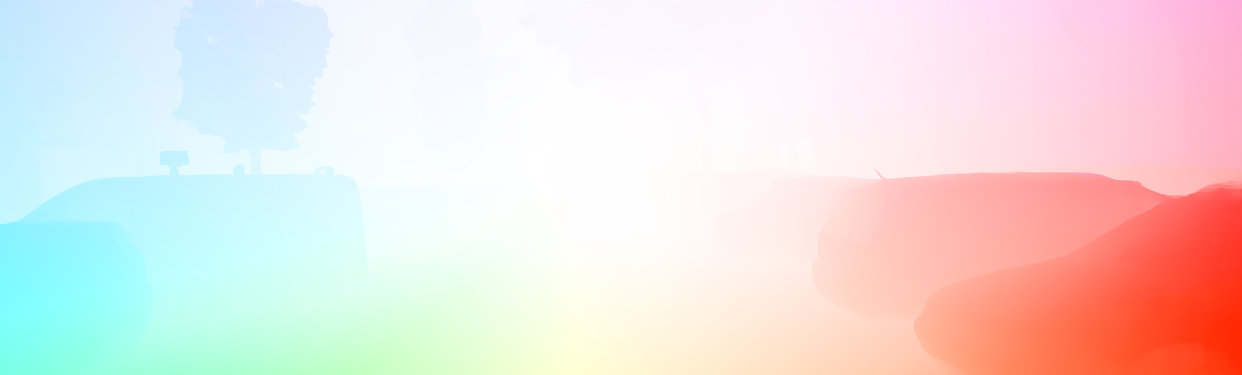}  \put (3,24) {\textcolor{purple}{\textbf{\texttt{EPE: 0.710}}}} \end{overpic}
    \\
    \end{tabular}
    \vspace{0.1cm}
    \caption{\textbf{Qualitative Results of Zero-Shot Generalization on KITTI~\cite{kitti} Training Set.} From top to bottom: first frame, flow by  SEA-RAFT (L)~\cite{wang2024sea}, FlowSeek (L)~\cite{Poggi_2025_ICCV}, WAFT-Twins-a2~\cite{wang2025waft}, and \net{} (XL).}
    \label{fig:supp_kitti_train}
\end{figure*}
\begin{figure*}[h]
    \centering
    \renewcommand{\tabcolsep}{0.1pt}
    \begin{tabular}{cccc}
    \rotatebox[origin=l]{90}{\scriptsize{\quad Image 1}} &
    \hspace{0.1cm}\includegraphics[width=0.315\textwidth, frame]{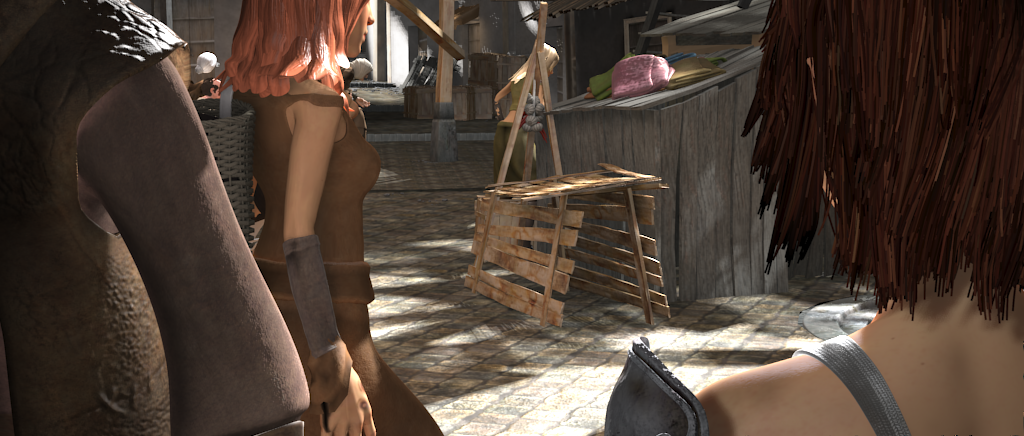} &
     \hspace{0.1cm}\includegraphics[width=0.315\textwidth, frame]{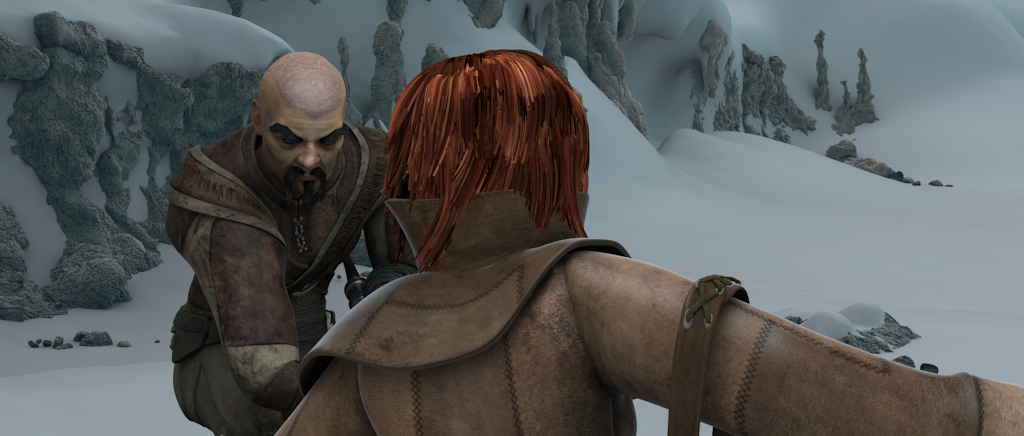}
     &
     \hspace{0.1cm}\includegraphics[width=0.315\textwidth, frame]{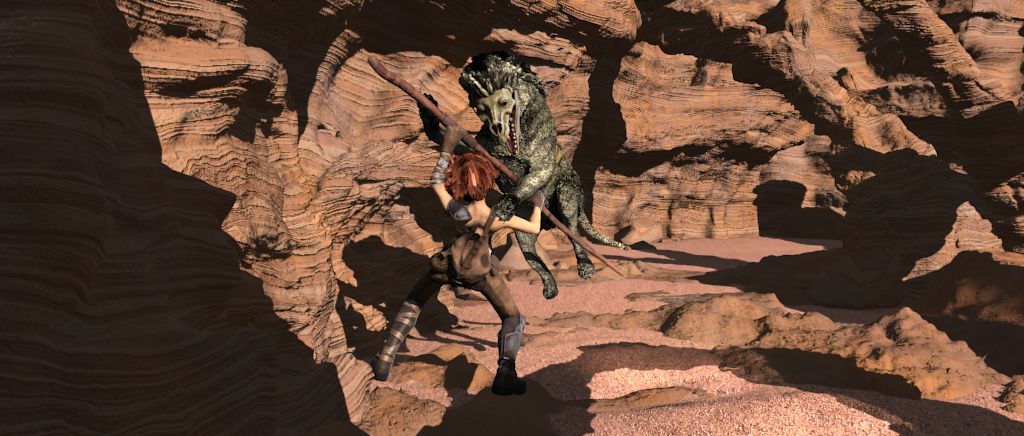}
    \\
    \rotatebox[origin=l]{90}{\scriptsize{SEA-RAFT}} &
    \begin{overpic}[width=0.315\textwidth, frame]{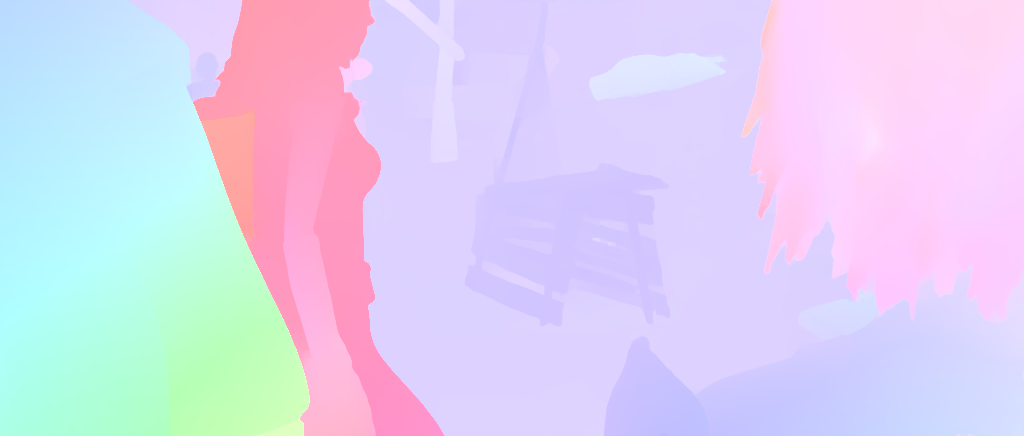} \put (53,36) {\textcolor{purple}{\textbf{\texttt{\small{EPE: 0.478}}}}}
    \end{overpic} &
     \begin{overpic}[width=0.315\textwidth, frame]{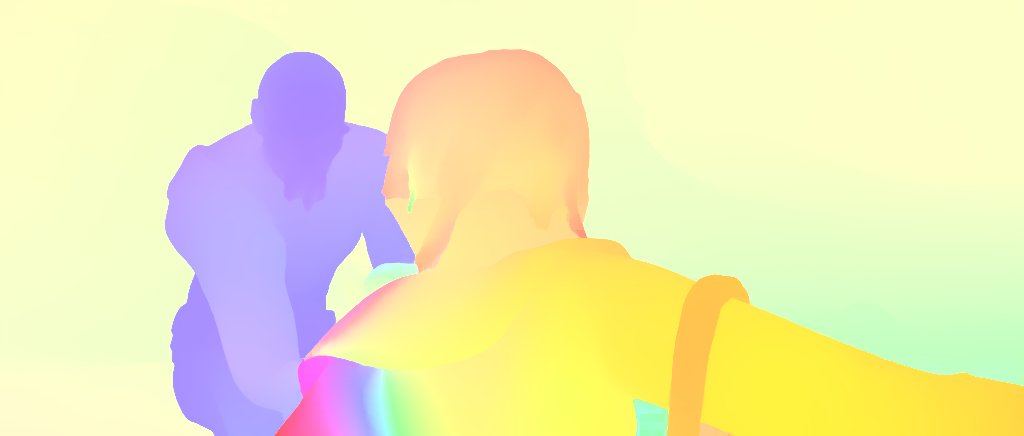} \put (53,36) {\textcolor{purple}{\textbf{\texttt{\small{EPE: 3.040}}}}}
     \end{overpic}
     &
     \begin{overpic}[width=0.315\textwidth, frame]{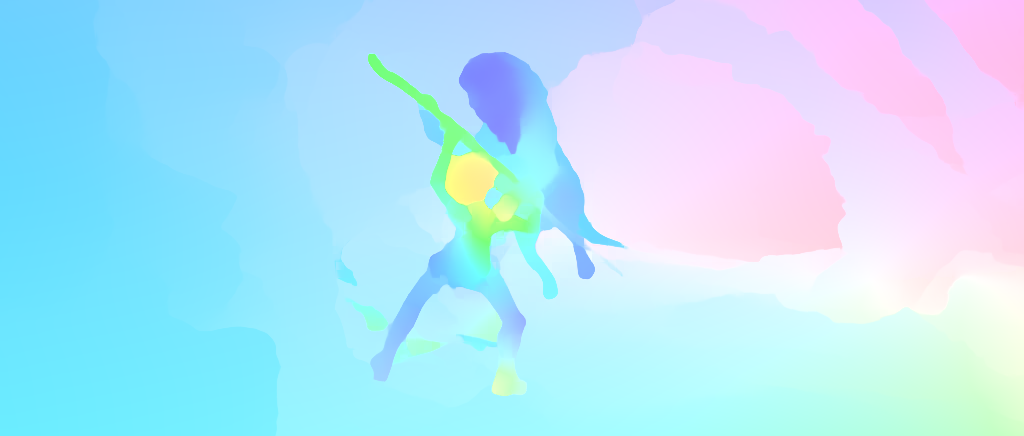} \put (53,36) {\textcolor{purple}{\textbf{\texttt{\small{EPE: 0.356}}}}}
     \end{overpic}
     \\
     \rotatebox[origin=l]{90}{\quad \scriptsize{WAFT}} &
    \begin{overpic}[width=0.315\textwidth, frame]{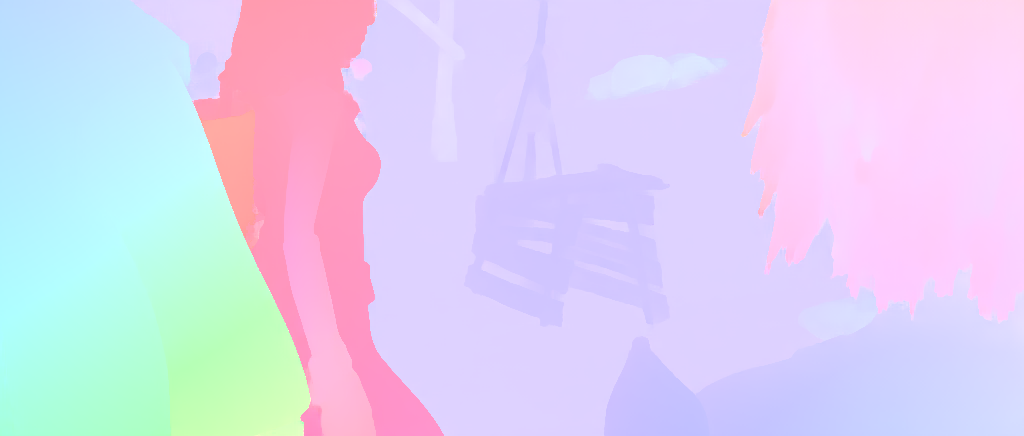} 
    \put (53,36) {\textcolor{purple}{\textbf{\texttt{\small{EPE: 0.398}}}}}
    \end{overpic} &
     \begin{overpic}[width=0.315\textwidth, frame]{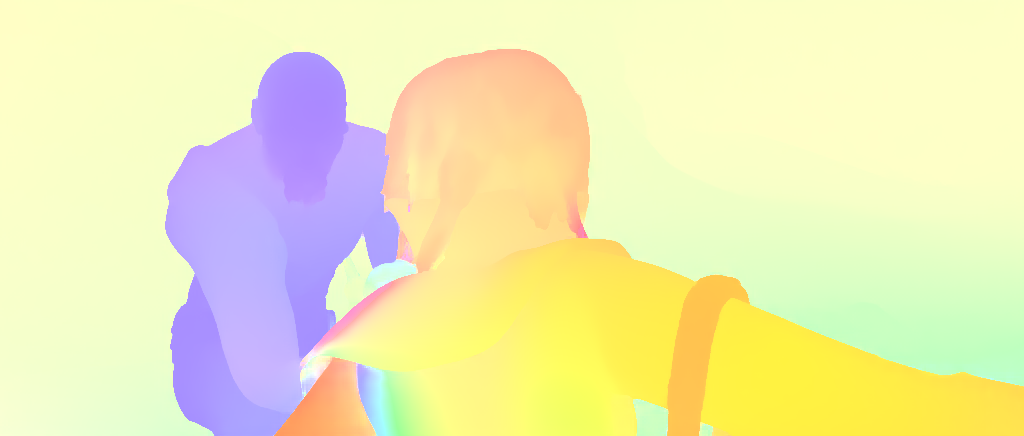} \put (53,36) {\textcolor{purple}{\textbf{\texttt{\small{EPE: 2.974}}}}}\end{overpic}
     &
     \begin{overpic}[width=0.315\textwidth, frame]{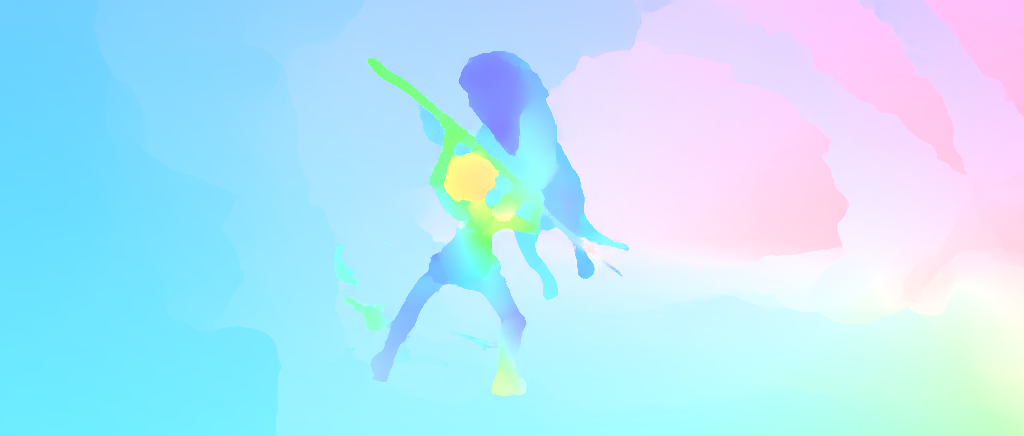} \put (53,36) {\textcolor{purple}{\textbf{\texttt{\small{EPE: 0.331}}}}}\end{overpic}
    \\
    \rotatebox[origin=l]{90}{\scriptsize{\quad \net{} (XL)}} &
    \begin{overpic}[width=0.315\textwidth, frame]{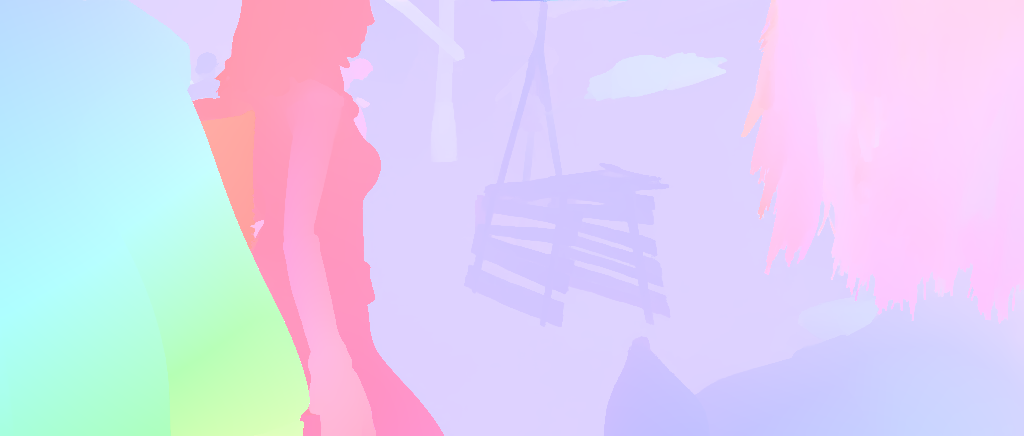} \put (53,36) {\textcolor{purple}{\textbf{\texttt{\small{EPE: 0.328}}}}} \end{overpic} &
     \begin{overpic}[width=0.315\textwidth, frame]{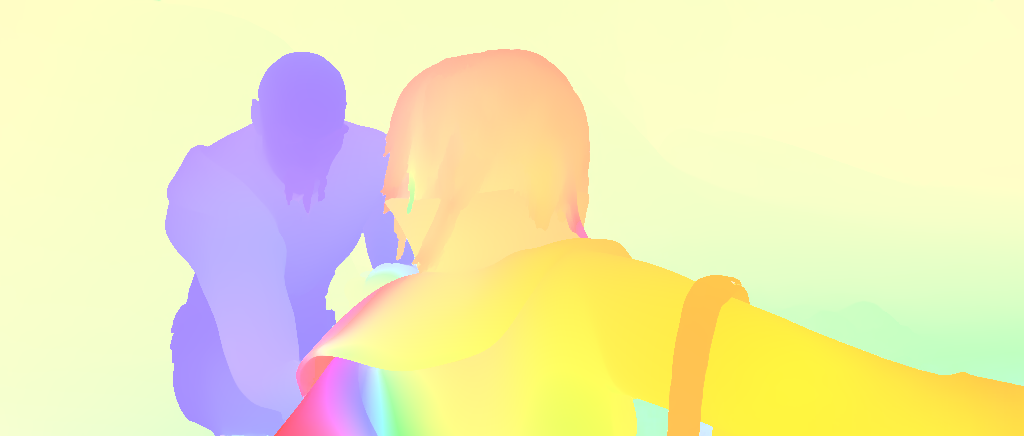}  \put (53,36) {\textcolor{purple}{\textbf{\texttt{\small{EPE: 1.952}}}}} \end{overpic}
     &
     \begin{overpic}[width=0.315\textwidth, frame]{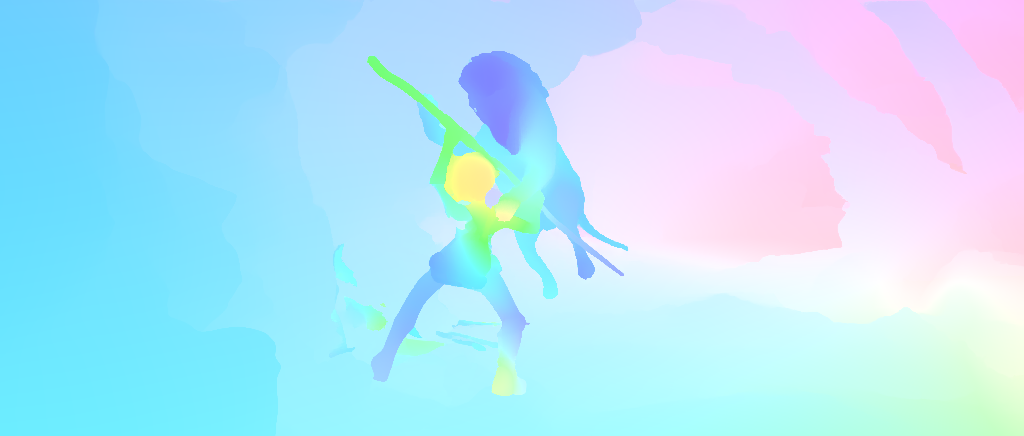}  \put (53,36) {\textcolor{purple}{\textbf{\texttt{\small{EPE: 0.277}}}}} \end{overpic}
     \\
    \rotatebox[origin=l]{90}{\scriptsize{SEA-RAFT}} &
    \begin{overpic}[width=0.315\textwidth, frame]{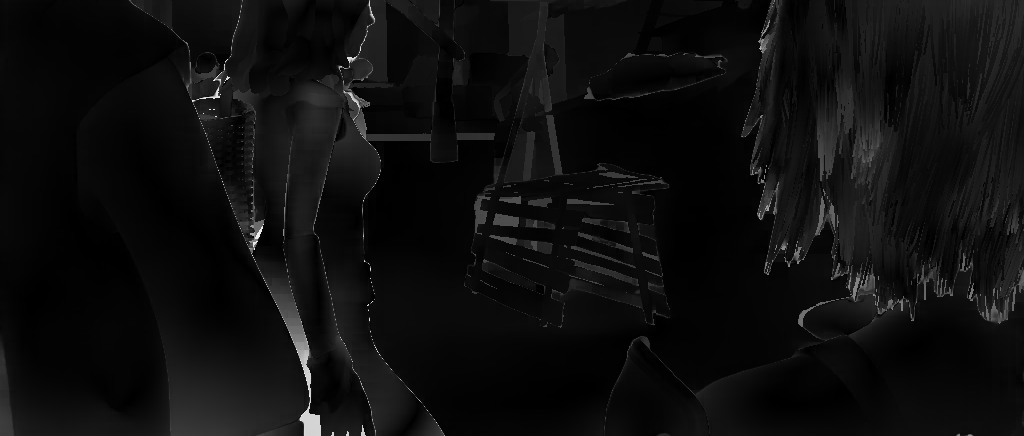} \put (3,36) {\textcolor{purple}{\textbf{\texttt{\small{}}}}}
    \end{overpic} &
     \begin{overpic}[width=0.315\textwidth, frame]{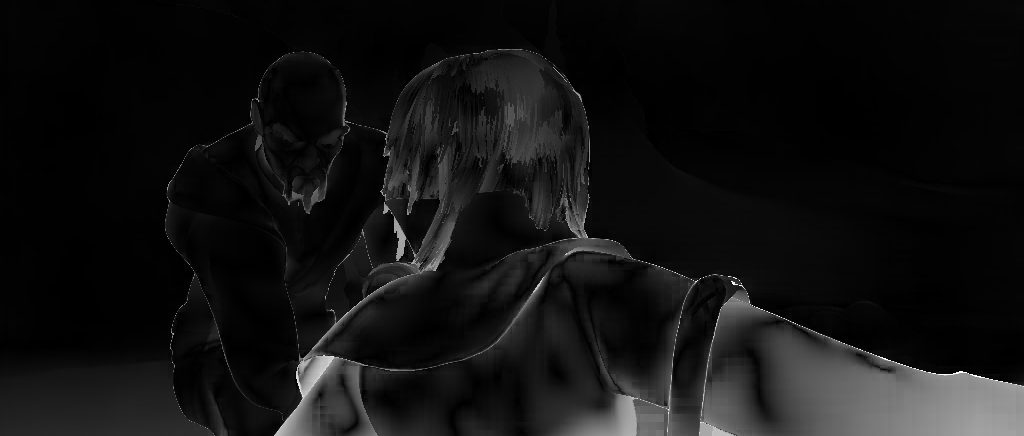} \put (53,36) {\textcolor{purple}{\textbf{\texttt{\small{}}}}}
     \end{overpic}
     &
     \begin{overpic}[width=0.315\textwidth, frame]{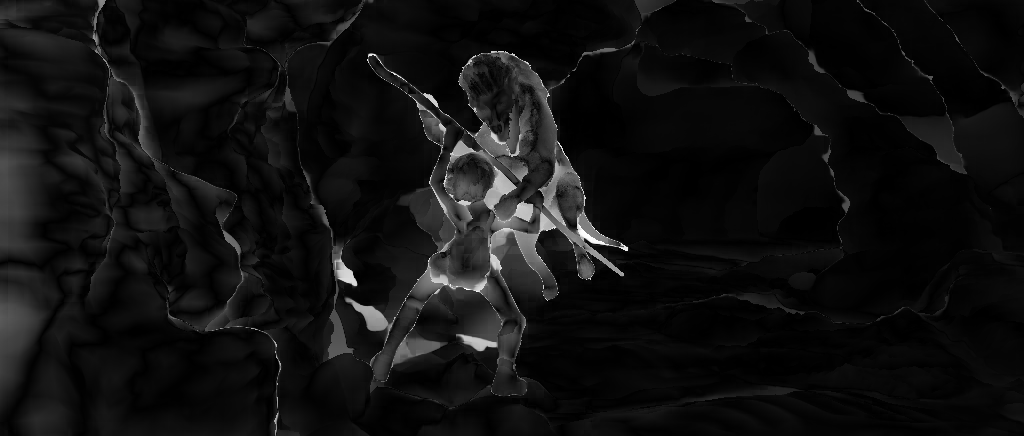} \put (3,36) {\textcolor{purple}{\textbf{\texttt{\small{}}}}}
     \end{overpic}
     \\
     \rotatebox[origin=l]{90}{\quad \scriptsize{WAFT}} &
    \begin{overpic}[width=0.315\textwidth, frame]{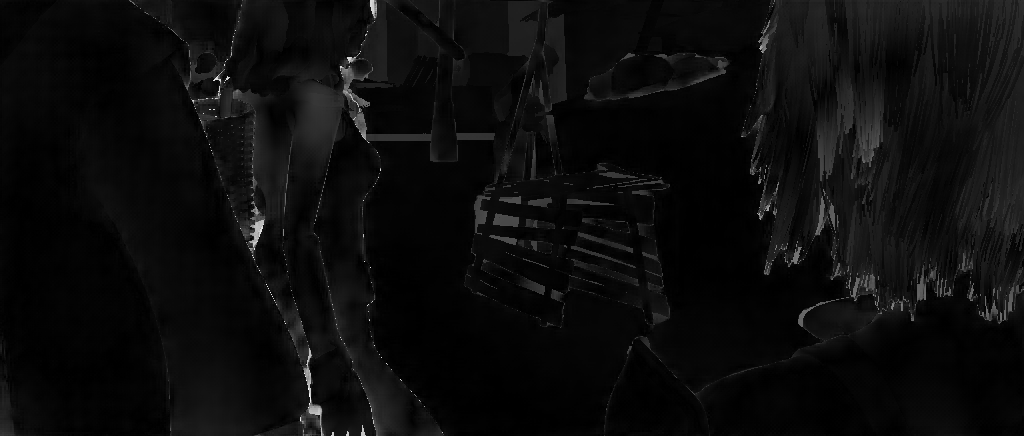} 
    \put (3,36) {\textcolor{purple}{\textbf{\texttt{\small{}}}}}
    \end{overpic} &
     \begin{overpic}[width=0.315\textwidth, frame]{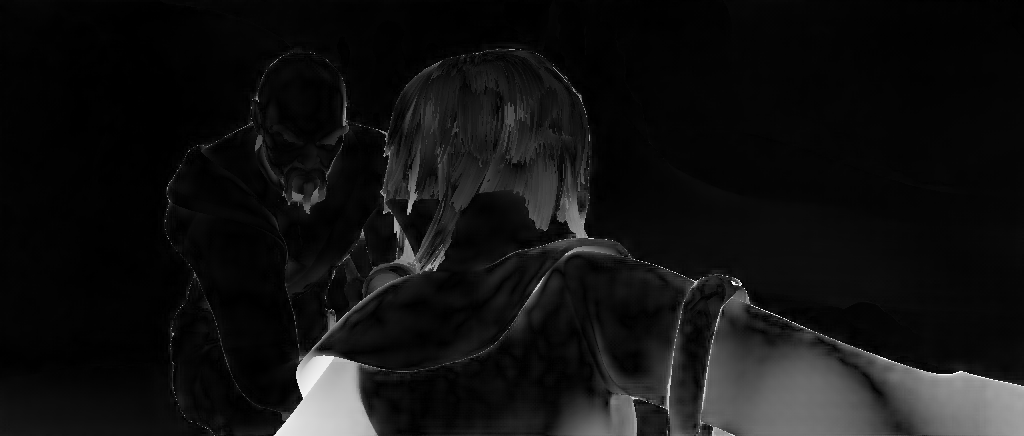} \put (53,36) {\textcolor{purple}{\textbf{\texttt{\small{}}}}}\end{overpic}
     &
     \begin{overpic}[width=0.315\textwidth, frame]{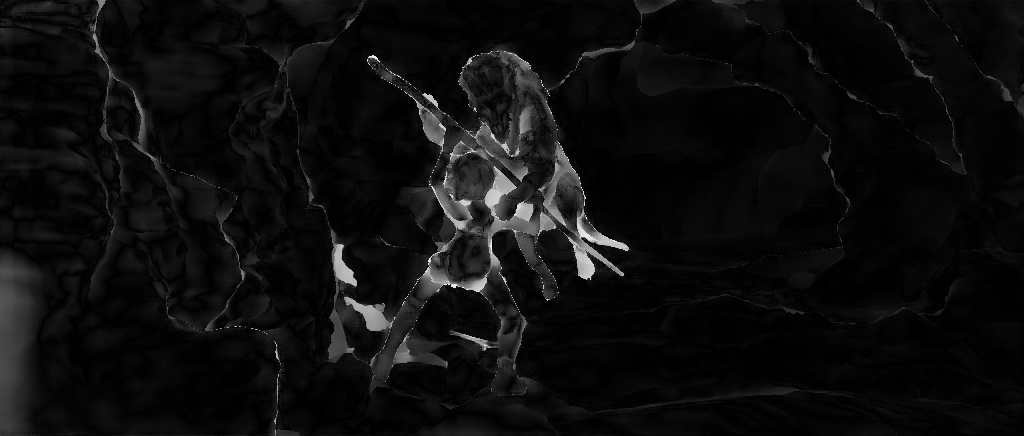} \put (3,36) {\textcolor{purple}{\textbf{\texttt{\small{}}}}}\end{overpic}
    \\
    \rotatebox[origin=l]{90}{\scriptsize{\quad \net{} (XL)}} &
    \begin{overpic}[width=0.315\textwidth, frame]{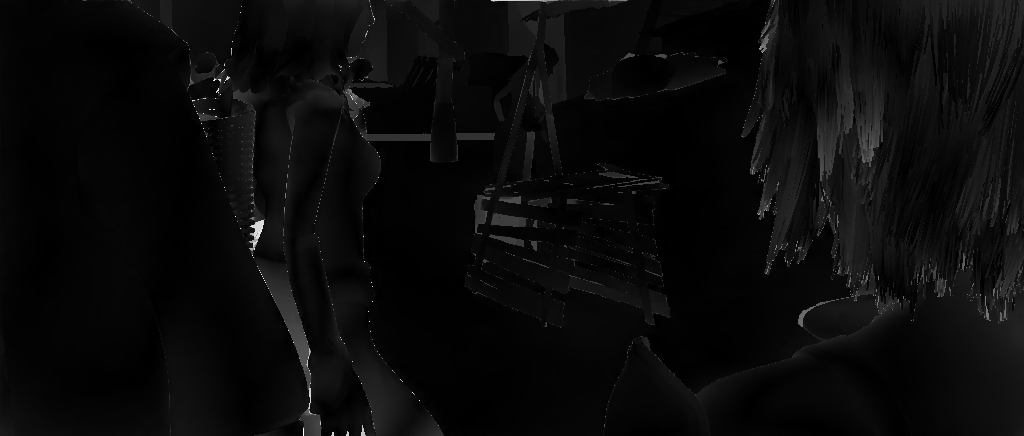} \put (3,36) {\textcolor{purple}{\textbf{\texttt{\small{}}}}} \end{overpic} &
     \begin{overpic}[width=0.315\textwidth, frame]{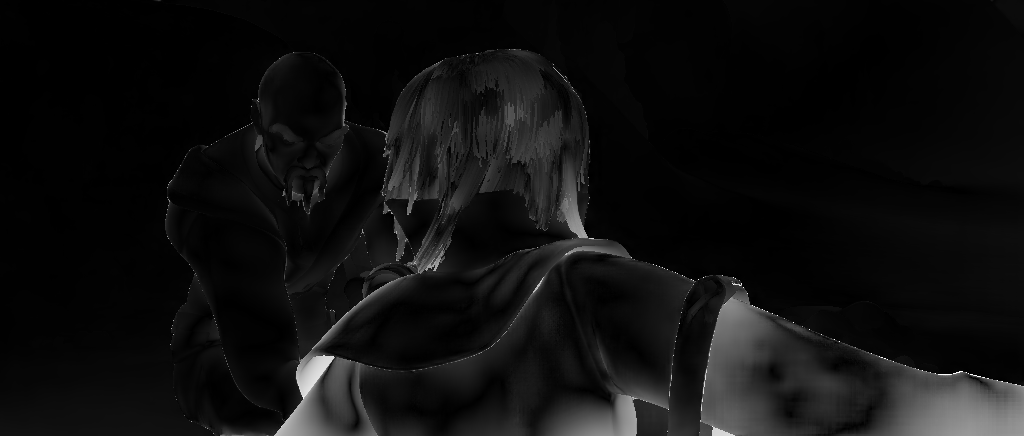}  \put (53,36) {\textcolor{purple}{\textbf{\texttt{\small{}}}}} \end{overpic}
     &
     \begin{overpic}[width=0.315\textwidth, frame]{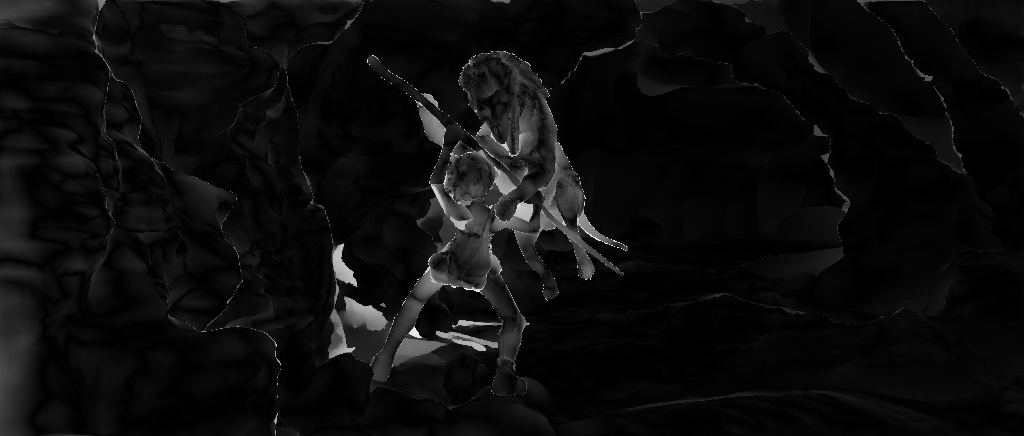}  \put (3,36) {\textcolor{purple}{\textbf{\texttt{\small{}}}}} \end{overpic}
    \end{tabular}
    \vspace{0.1cm}
    \caption{\textbf{Qualitative Results on Sintel (clean)~\cite{sintel} Test Set.} From top to bottom: first frame, flow predicted by SEA-RAFT (L)~\cite{wang2024sea}, WAFT-DINOv3-a2~\cite{wang2025waft}, and \net{} (XL), followed by the corresponding error maps. Results are obtained directly from the official Sintel evaluation server.}
    \label{fig:supp_sintel_test_clean}
\end{figure*}
\begin{figure*}[h]
    \centering
    \renewcommand{\tabcolsep}{0.1pt}
    \begin{tabular}{cccc}
    \rotatebox[origin=l]{90}{\scriptsize{\quad Image 1}} &
    \hspace{0.1cm}\includegraphics[width=0.315\textwidth, frame]{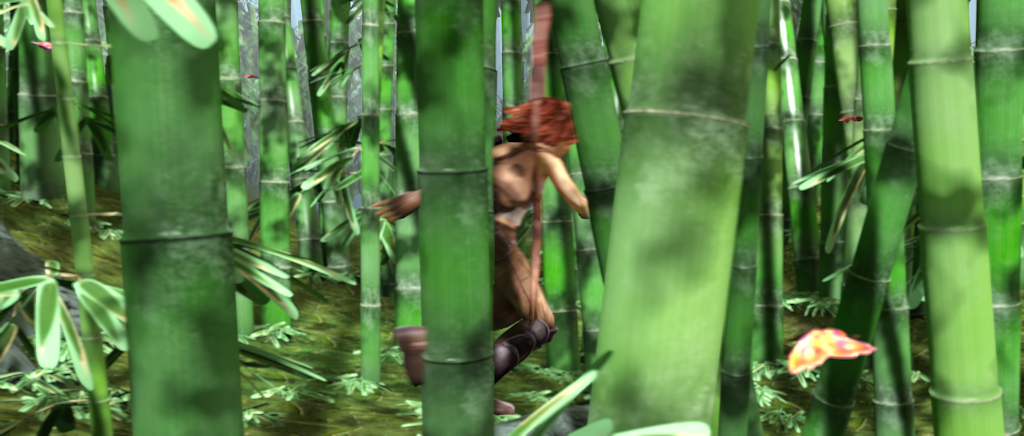} &
     \hspace{0.1cm}\includegraphics[width=0.315\textwidth, frame]{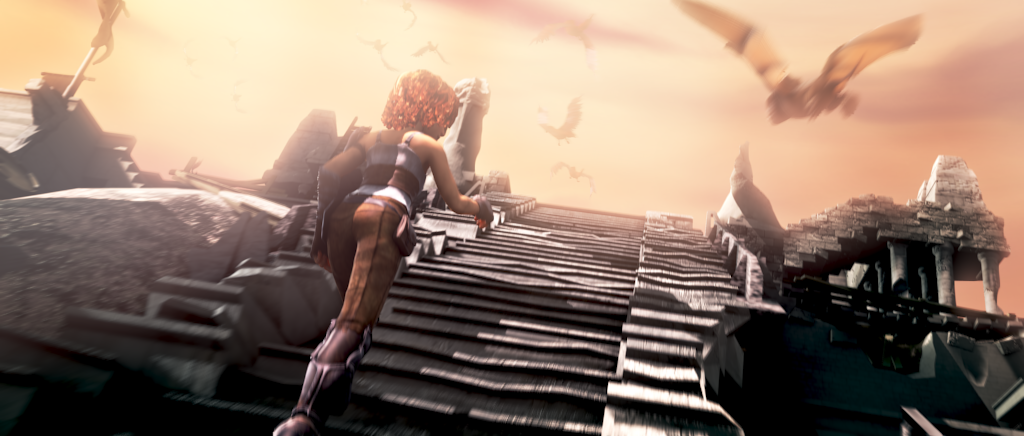}
     &
     \hspace{0.1cm}\includegraphics[width=0.315\textwidth, frame]{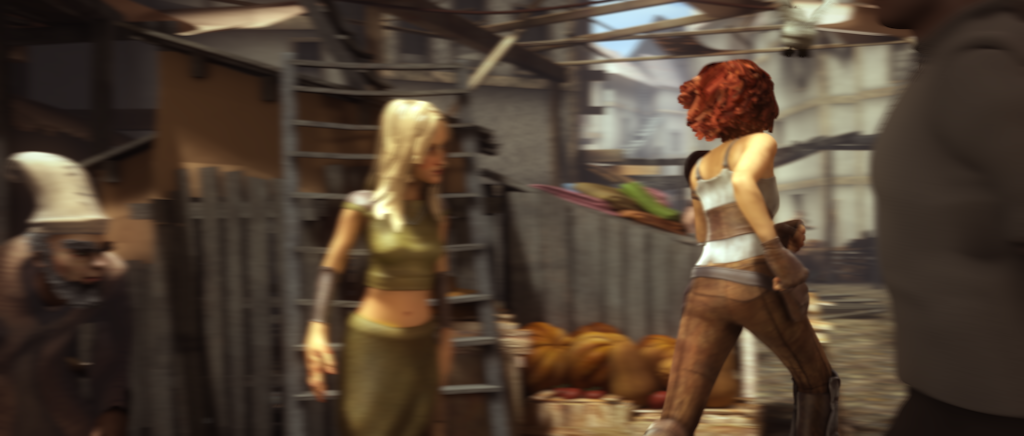}
    \\
    \rotatebox[origin=l]{90}{\scriptsize{SEA-RAFT}} &
    \begin{overpic}[width=0.315\textwidth, frame]{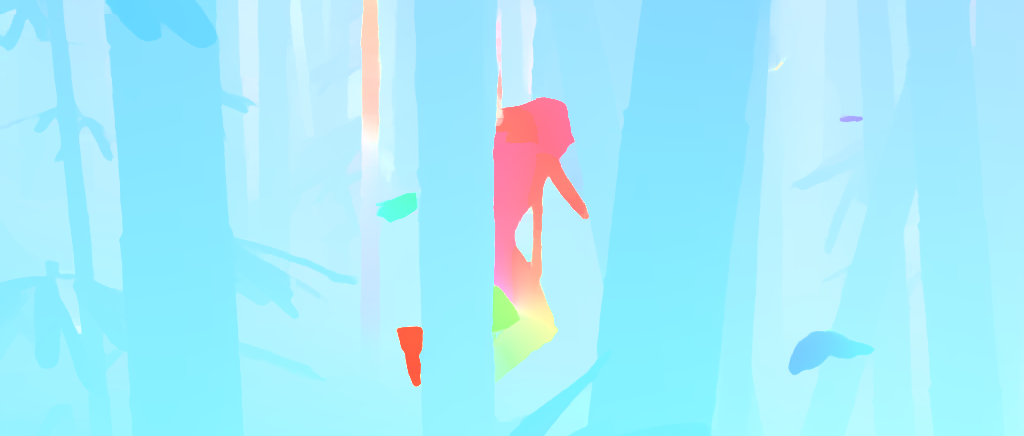} \put (3,36) {\textcolor{purple}{\textbf{\texttt{\small{EPE: 0.532}}}}}
    \end{overpic} &
     \begin{overpic}[width=0.315\textwidth, frame]{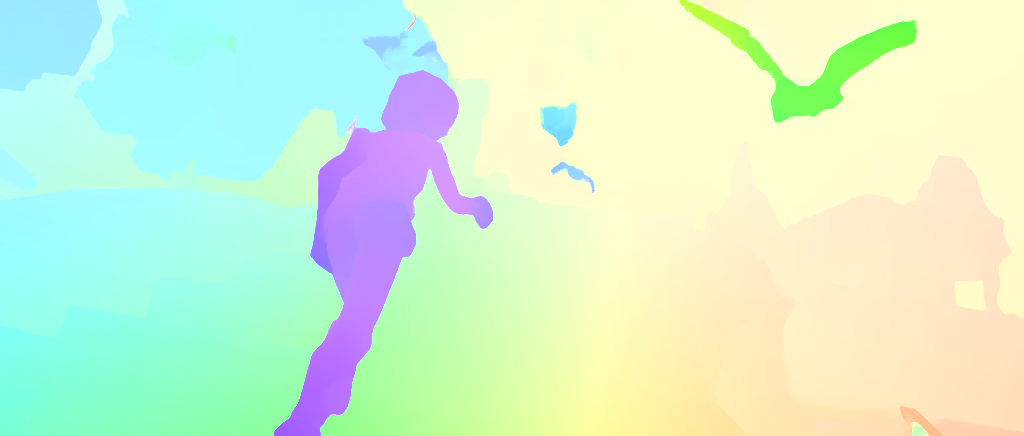} \put (3,36) {\textcolor{purple}{\textbf{\texttt{\small{EPE: 0.654}}}}}
     \end{overpic}
     &
     \begin{overpic}[width=0.315\textwidth, frame]{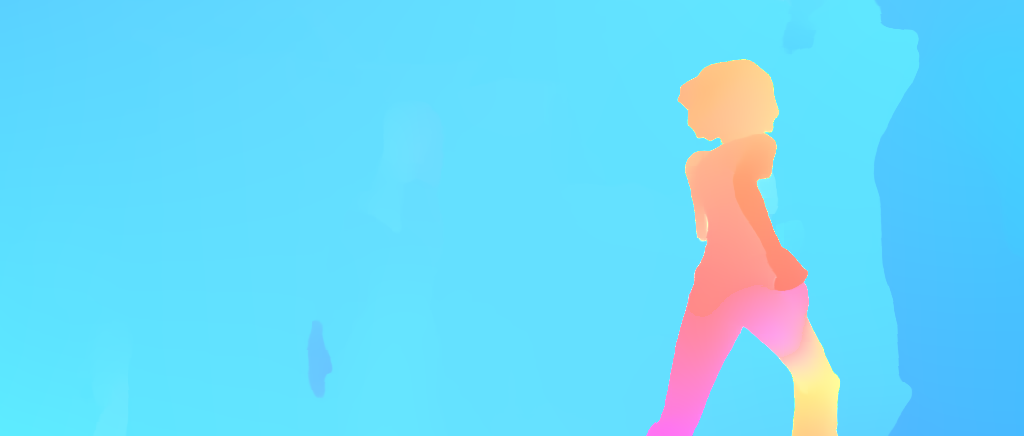} \put (3,36) {\textcolor{purple}{\textbf{\texttt{\small{EPE: 1.077}}}}}
     \end{overpic}
     \\
     \rotatebox[origin=l]{90}{\quad \scriptsize{WAFT}} &
    \begin{overpic}[width=0.315\textwidth, frame]{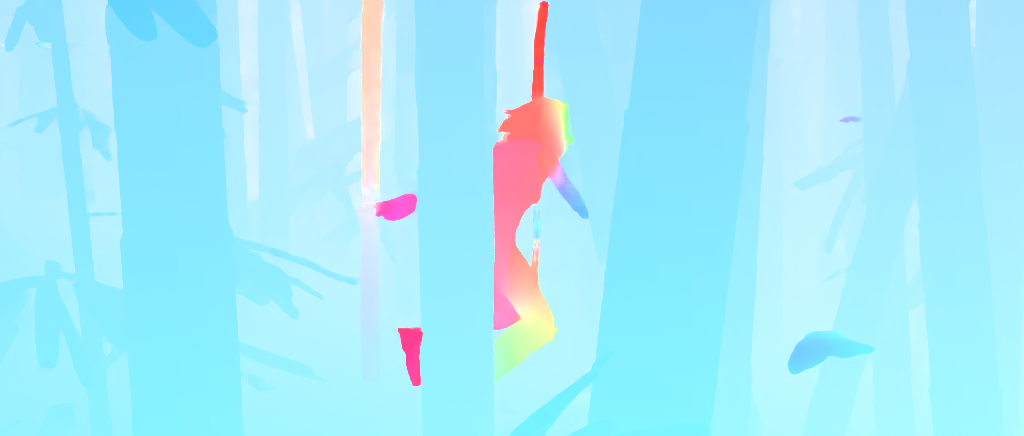} 
    \put (3,36) {\textcolor{purple}{\textbf{\texttt{\small{EPE: 0.467}}}}}
    \end{overpic} &
     \begin{overpic}[width=0.315\textwidth, frame]{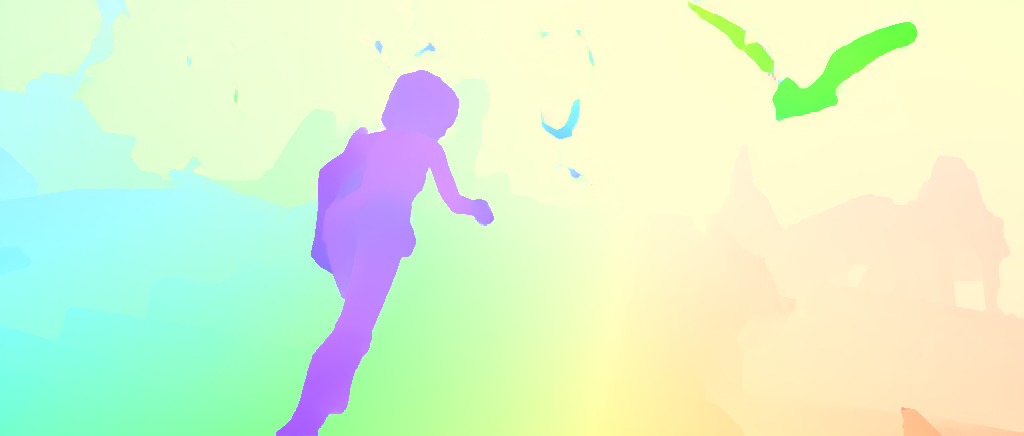} \put (3,36) {\textcolor{purple}{\textbf{\texttt{\small{EPE: 0.520}}}}}\end{overpic}
     &
     \begin{overpic}[width=0.315\textwidth, frame]{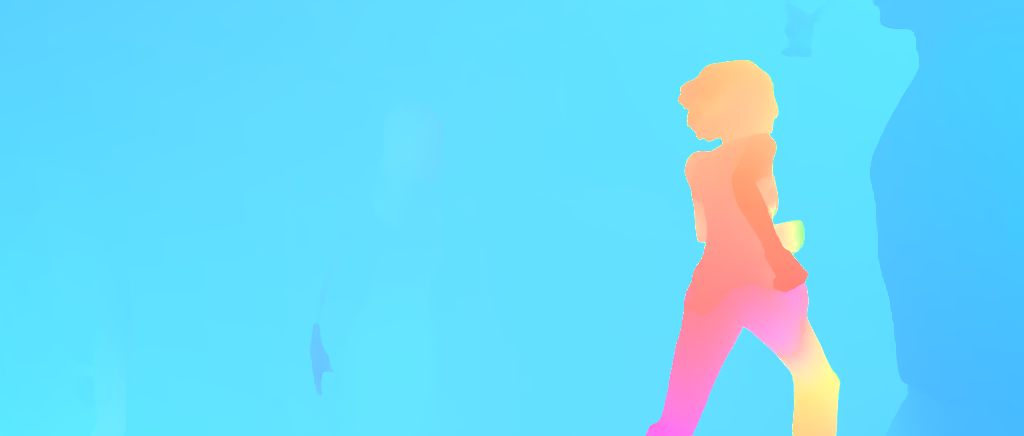} \put (3,36) {\textcolor{purple}{\textbf{\texttt{\small{EPE: 0.836}}}}}\end{overpic}
    \\
    \rotatebox[origin=l]{90}{\scriptsize{\quad \net{} (XL)}} &
    \begin{overpic}[width=0.315\textwidth, frame]{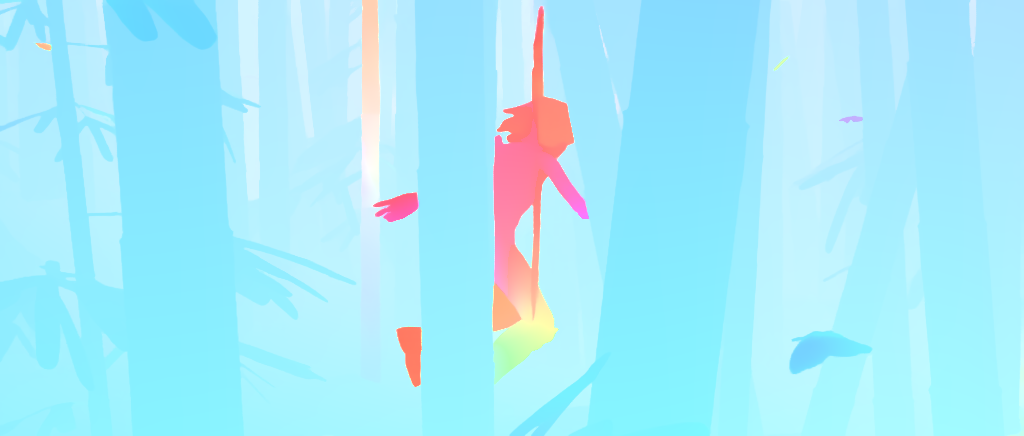} \put (3,36) {\textcolor{purple}{\textbf{\texttt{\small{EPE: 0.346}}}}} \end{overpic} &
     \begin{overpic}[width=0.315\textwidth, frame]{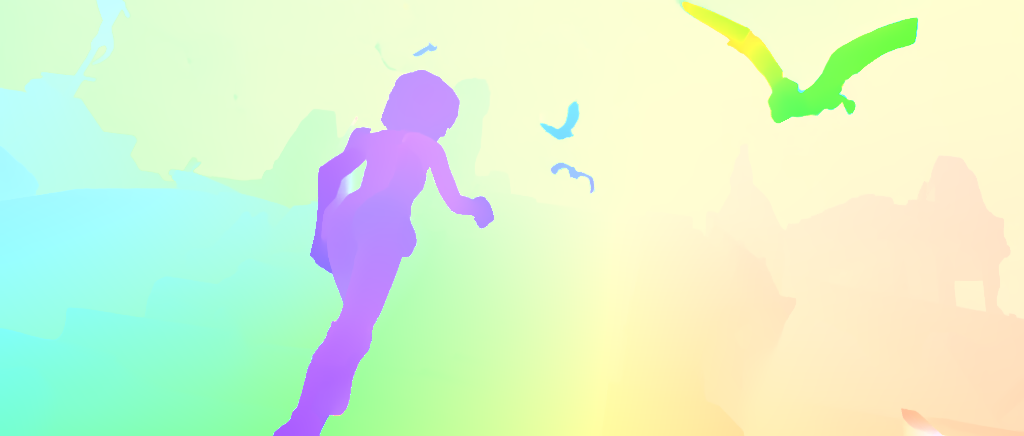}  \put (3,36) {\textcolor{purple}{\textbf{\texttt{\small{EPE: 0.515}}}}} \end{overpic}
     &
     \begin{overpic}[width=0.315\textwidth, frame]{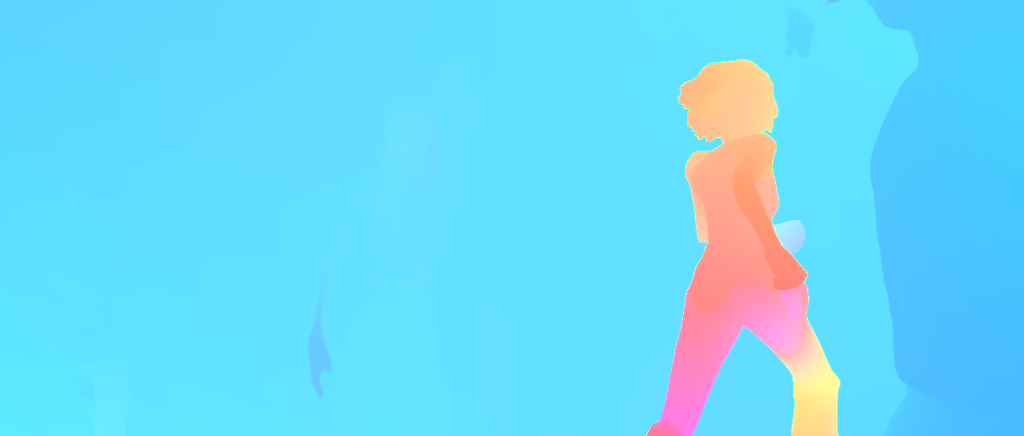}  \put (3,36) {\textcolor{purple}{\textbf{\texttt{\small{EPE: 0.788}}}}} \end{overpic}
     \\
    \rotatebox[origin=l]{90}{\scriptsize{SEA-RAFT}} &
    \begin{overpic}[width=0.315\textwidth, frame]{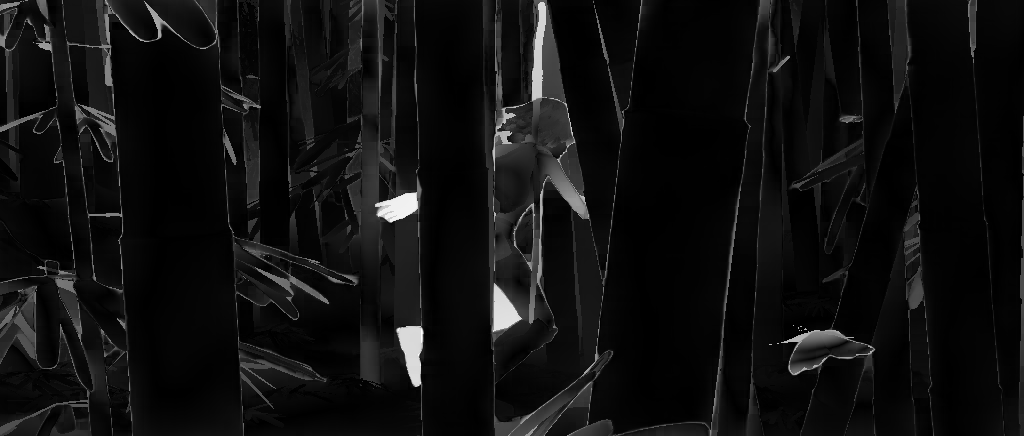} \put (3,36) {\textcolor{purple}{\textbf{\texttt{\small{}}}}}
    \end{overpic} &
     \begin{overpic}[width=0.315\textwidth, frame]{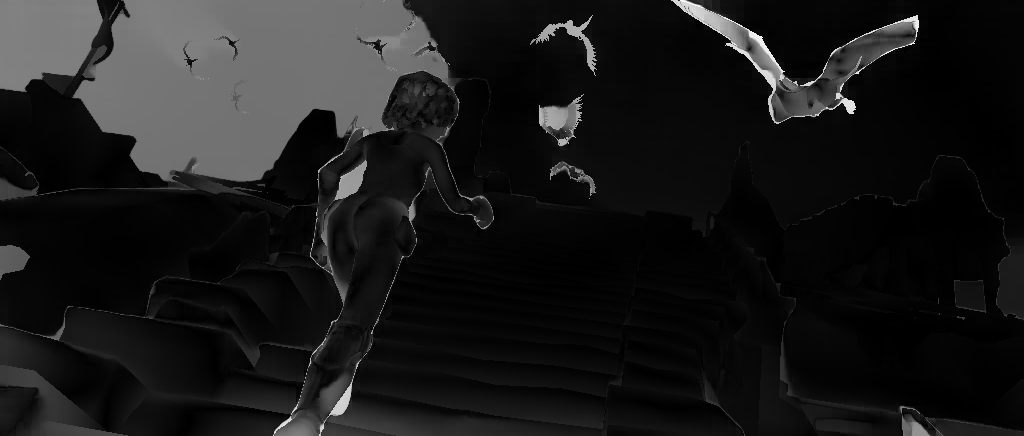} \put (3,36) {\textcolor{purple}{\textbf{\texttt{\small{}}}}}
     \end{overpic}
     &
     \begin{overpic}[width=0.315\textwidth, frame]{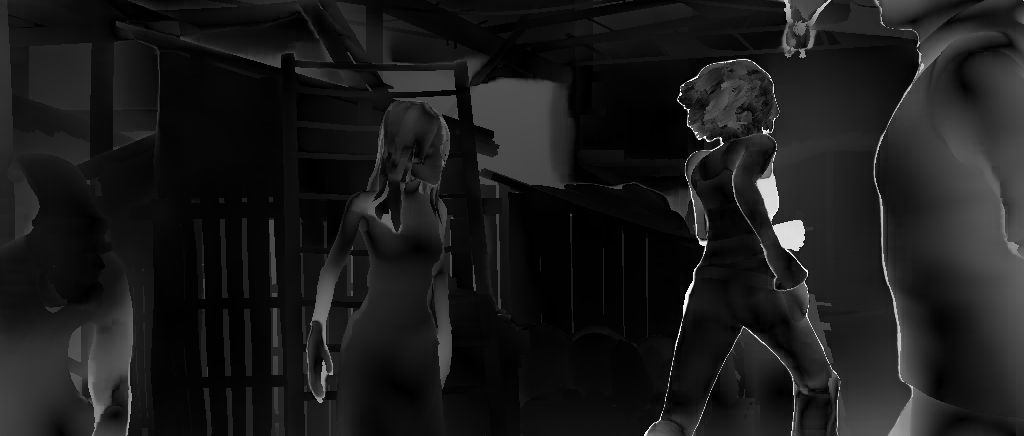} \put (3,36) {\textcolor{purple}{\textbf{\texttt{\small{}}}}}
     \end{overpic}
     \\
     \rotatebox[origin=l]{90}{\quad \scriptsize{WAFT}} &
    \begin{overpic}[width=0.315\textwidth, frame]{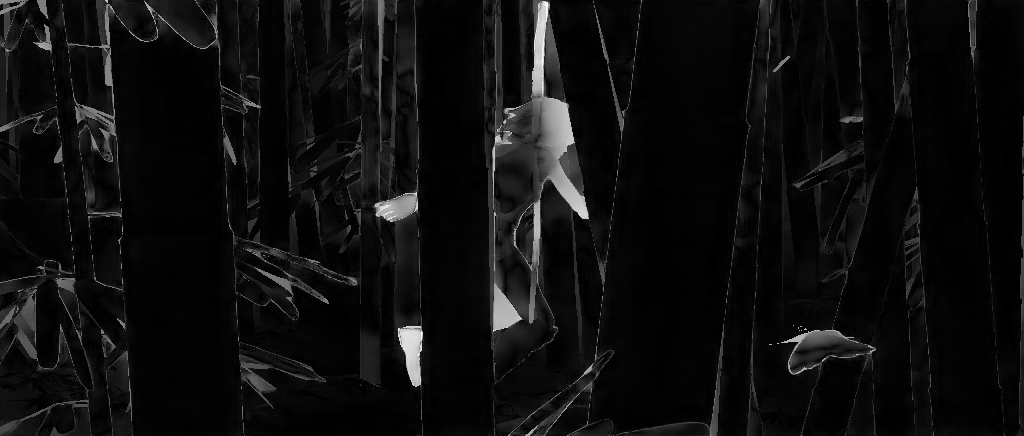} 
    \put (3,36) {\textcolor{purple}{\textbf{\texttt{\small{}}}}}
    \end{overpic} &
     \begin{overpic}[width=0.315\textwidth, frame]{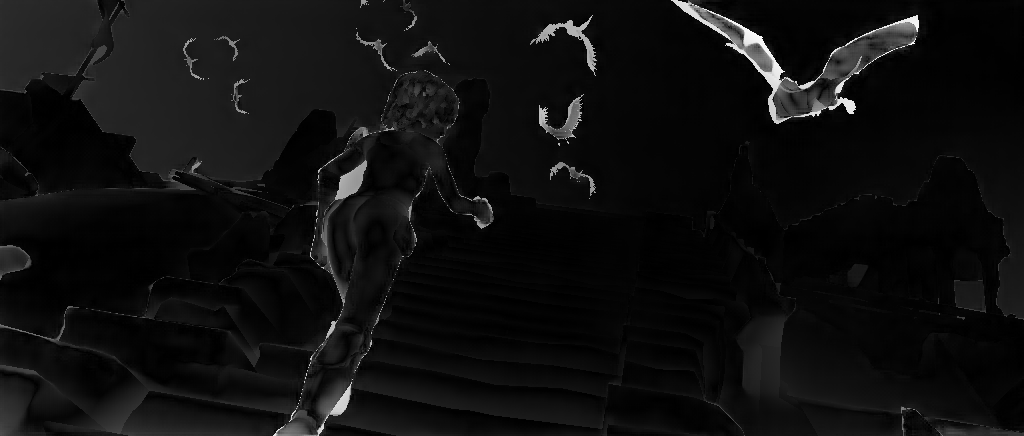} \put (3,36) {\textcolor{purple}{\textbf{\texttt{\small{}}}}}\end{overpic}
     &
     \begin{overpic}[width=0.315\textwidth, frame]{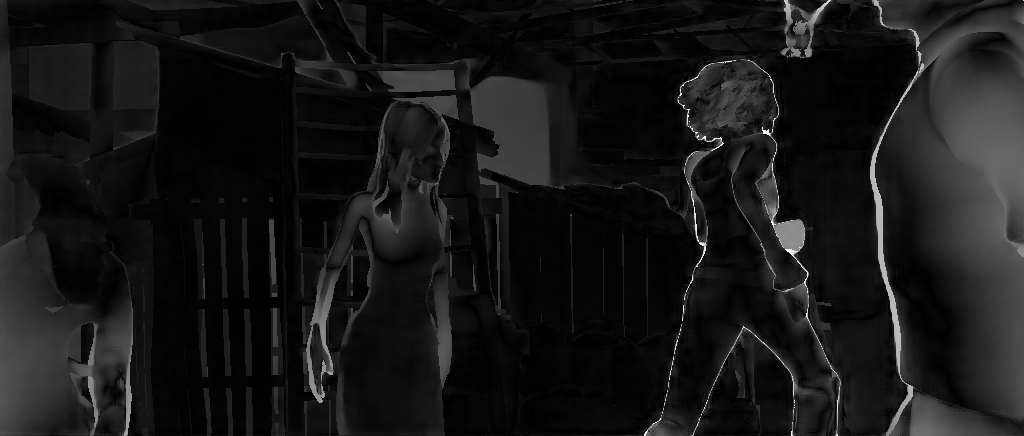} \put (3,36) {\textcolor{purple}{\textbf{\texttt{\small{}}}}}\end{overpic}
    \\
    \rotatebox[origin=l]{90}{\scriptsize{\quad \net{} (XL)}} &
    \begin{overpic}[width=0.315\textwidth, frame]{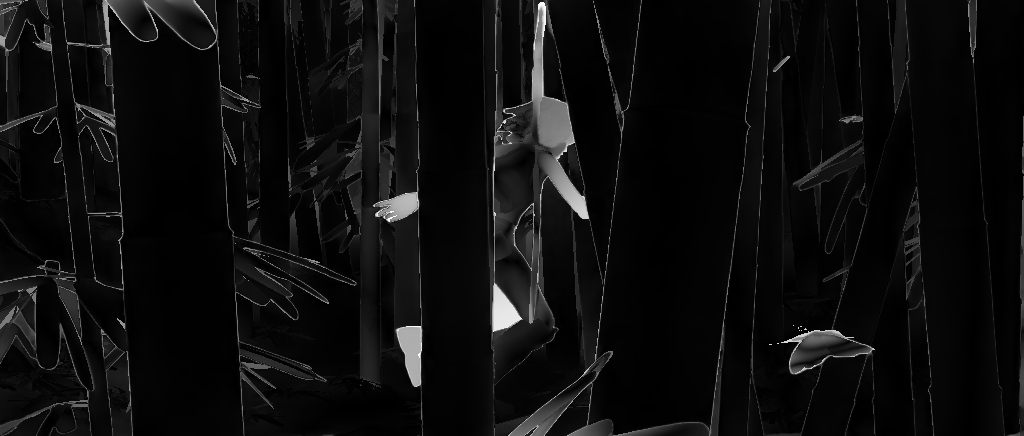} \put (3,36) {\textcolor{purple}{\textbf{\texttt{\small{}}}}} \end{overpic} &
     \begin{overpic}[width=0.315\textwidth, frame]{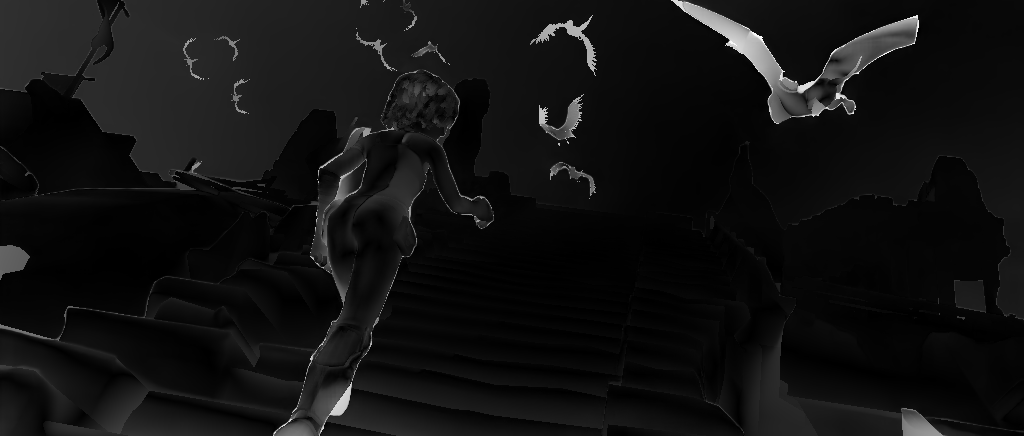}  \put (3,36) {\textcolor{purple}{\textbf{\texttt{\small{}}}}} \end{overpic}
     &
     \begin{overpic}[width=0.315\textwidth, frame]{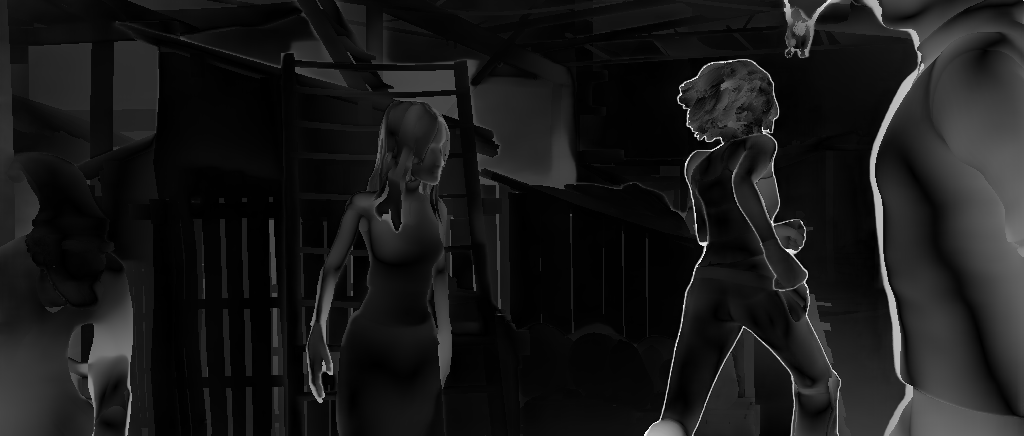}  \put (3,36) {\textcolor{purple}{\textbf{\texttt{\small{}}}}} \end{overpic}
    \end{tabular}
    \vspace{0.1cm}
    \caption{\textbf{Qualitative Results on Sintel (final)~\cite{sintel} Test Set.} From top to bottom: first frame, flow predicted by SEA-RAFT (L)~\cite{wang2024sea}, WAFT-DINOv3-a2~\cite{wang2025waft}, and \net{} (XL), followed by the corresponding error maps. Results are obtained directly from the official Sintel evaluation server.}
    \label{fig:supp_sintel_test_final}
\end{figure*}
\begin{figure*}[t]
    \centering
    \renewcommand{\tabcolsep}{0.5pt}
    \begin{tabular}{cccc}
    \rotatebox[origin=l]{90}{\scriptsize{\quad \quad Image 1}} &
    \hspace{0.1cm}\includegraphics[width=0.315\textwidth, frame]{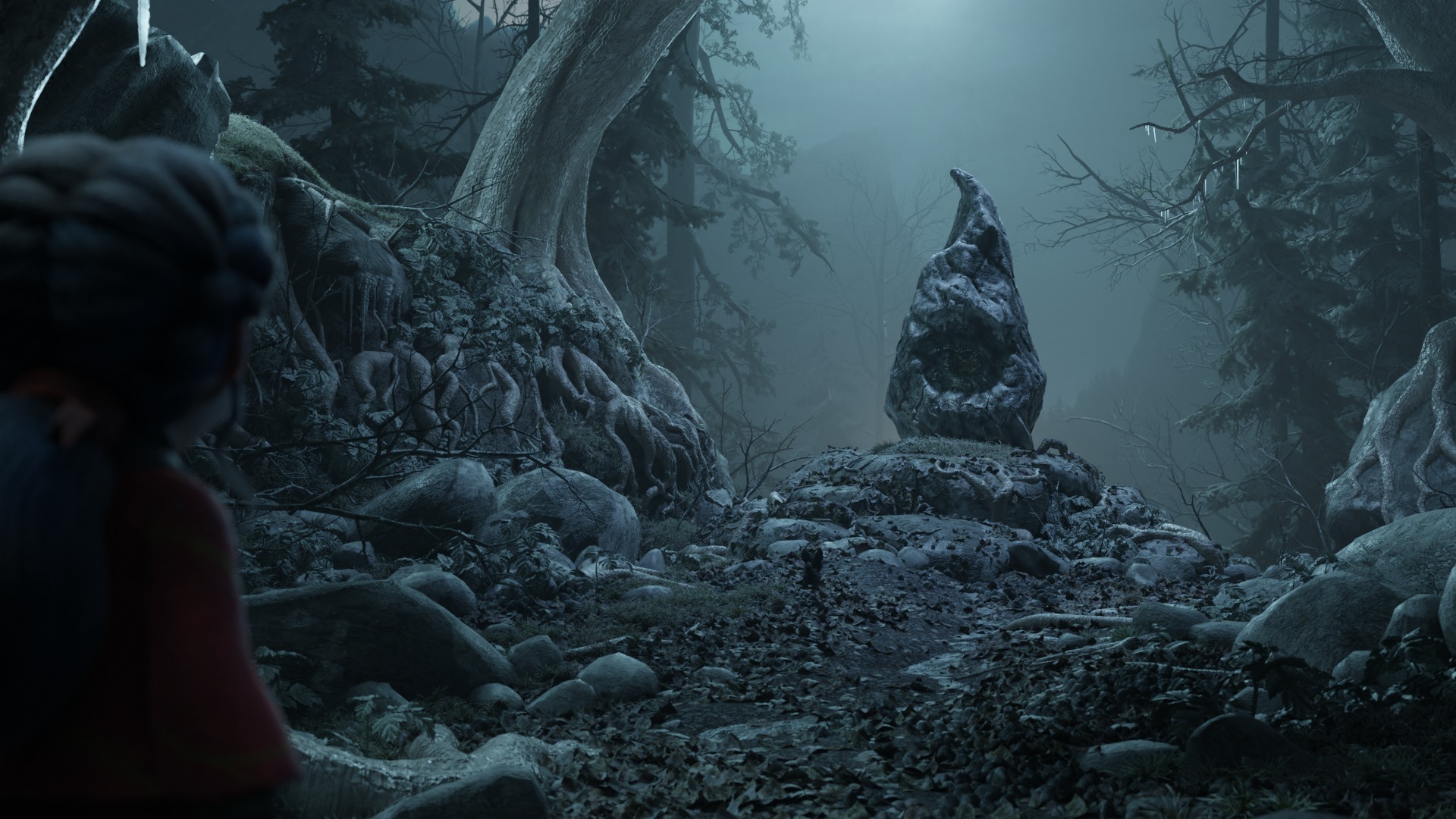} &
     \hspace{0.1cm}\includegraphics[width=0.315\textwidth, frame]{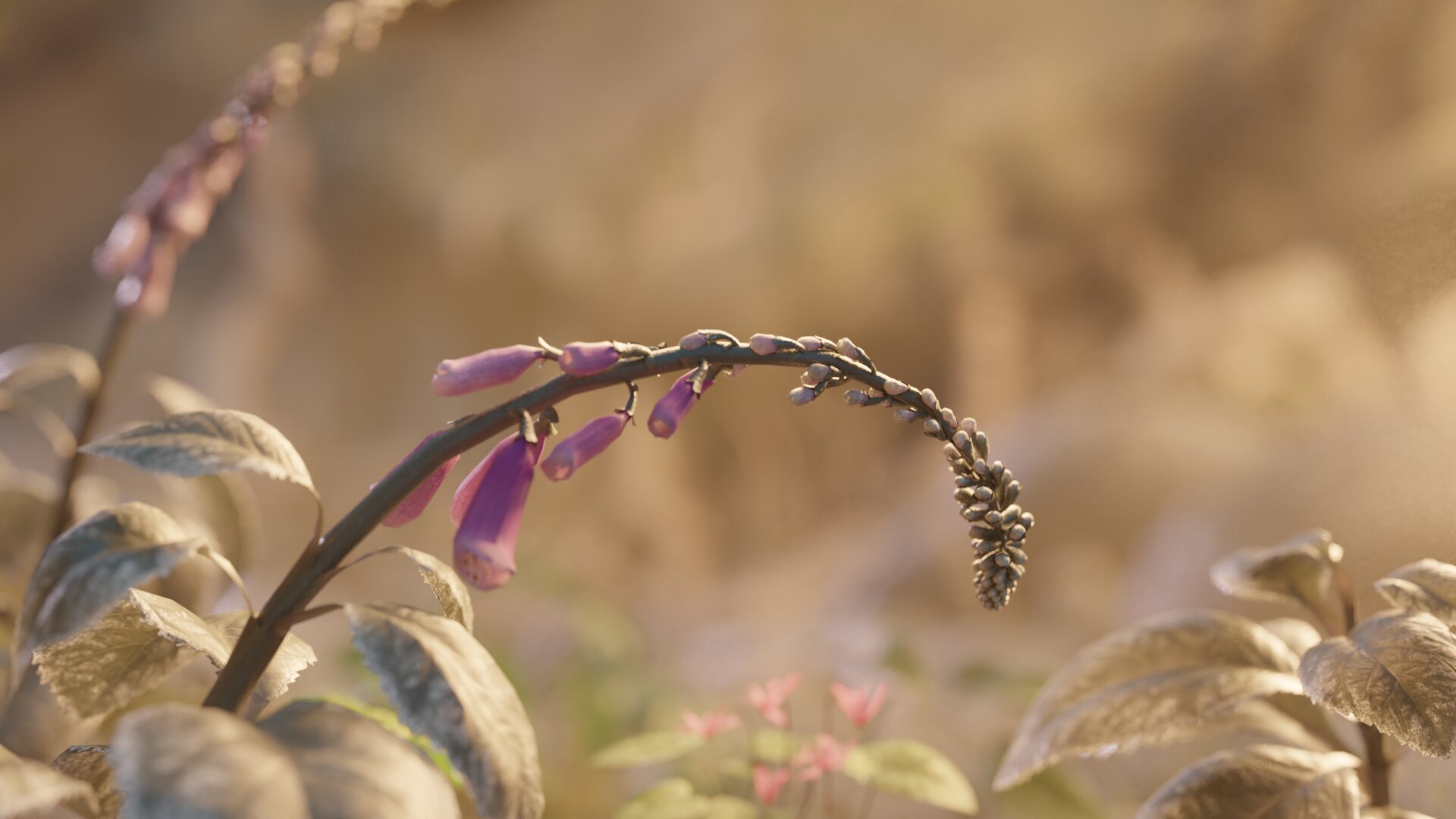}
     &
     \hspace{0.1cm}\includegraphics[width=0.315\textwidth, frame]{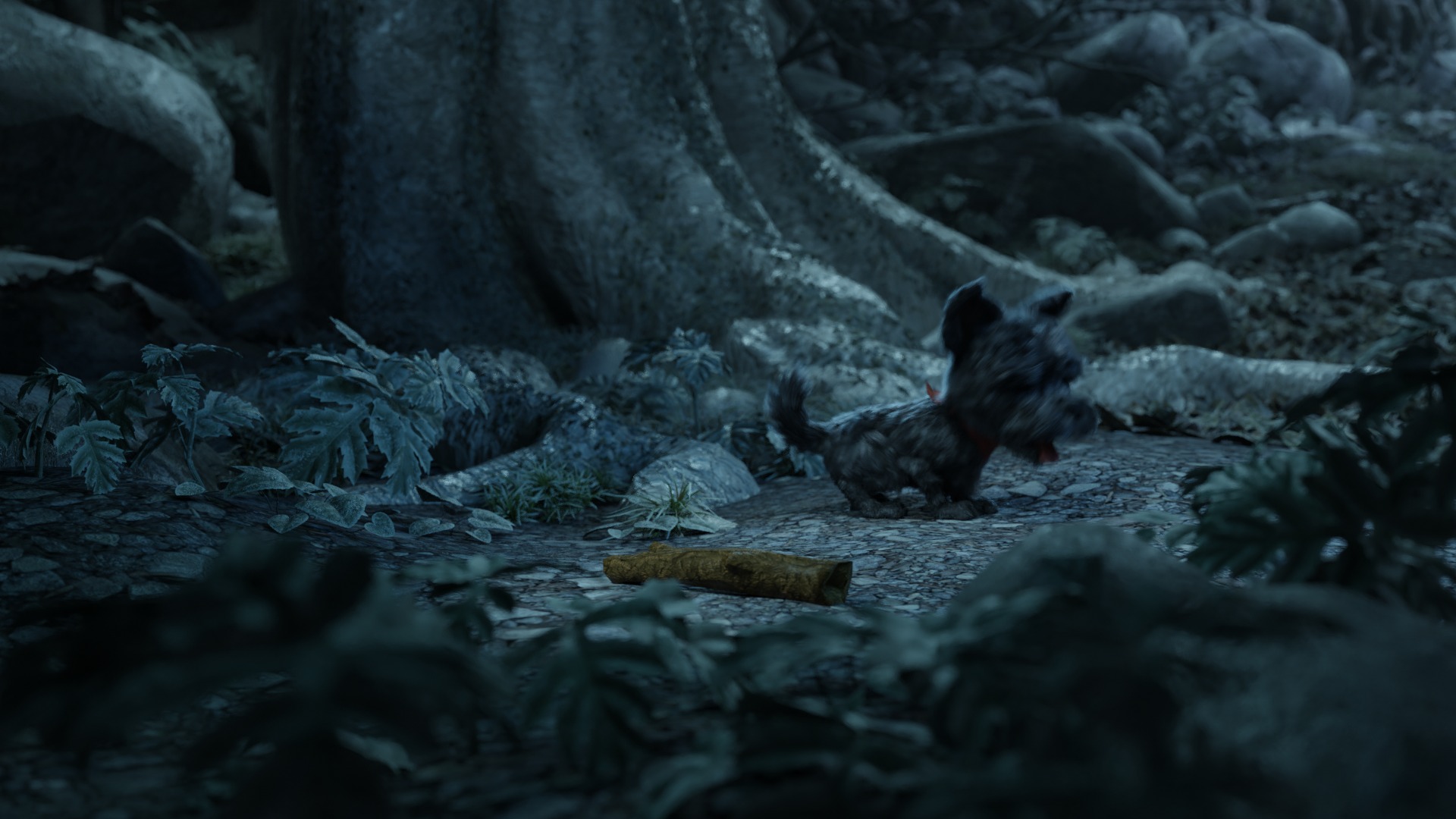}
     \\
    \rotatebox[origin=l]{90}{\scriptsize{\quad SEA-RAFT~\cite{wang2024sea}}} &
    \begin{overpic}[width=0.315\textwidth, frame]{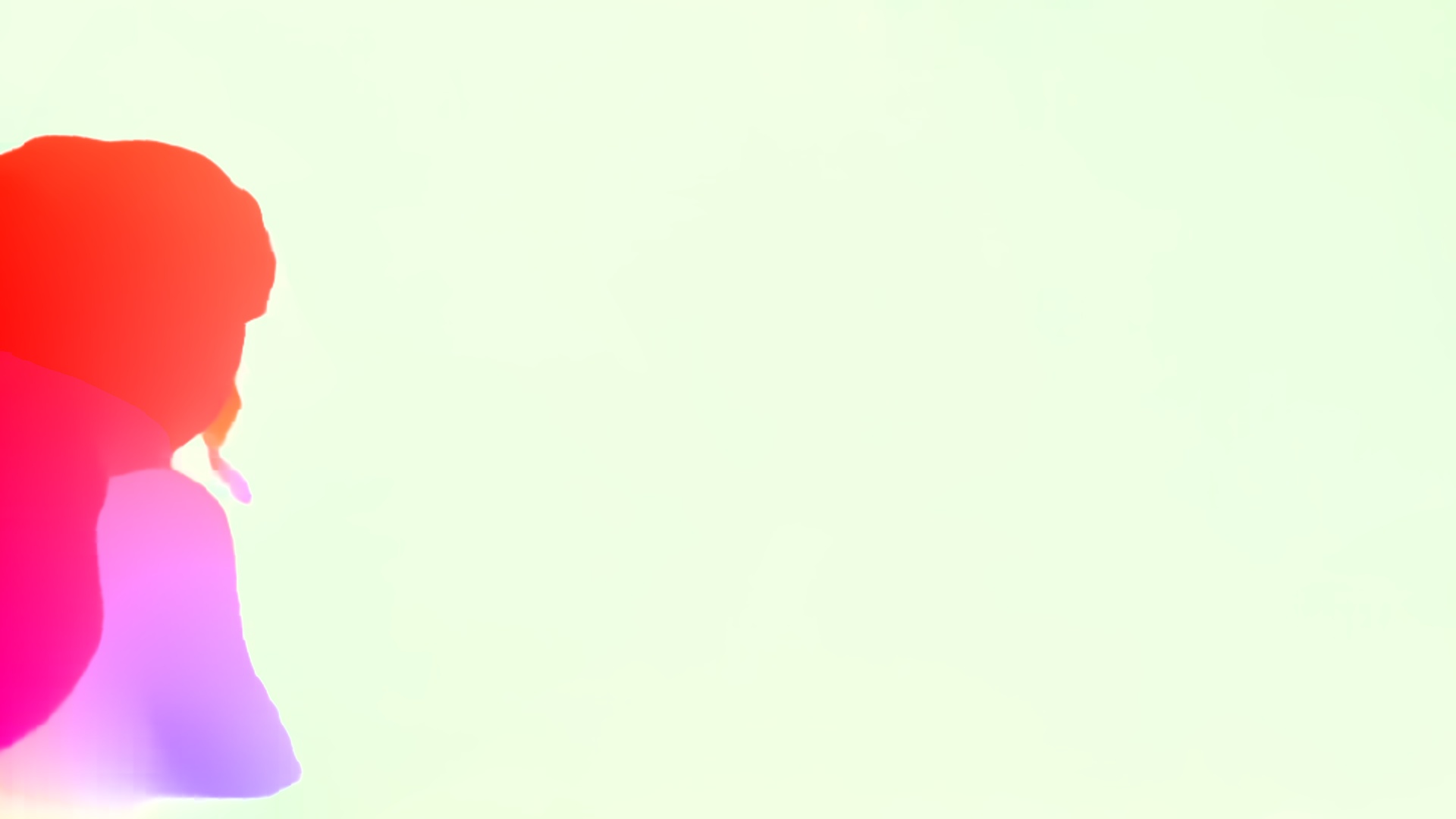} \put (3,48) {\textcolor{purple}{\textbf{\texttt{\small{EPE: 0.264}}}}}
    \end{overpic} &
     \begin{overpic}[width=0.315\textwidth, frame]{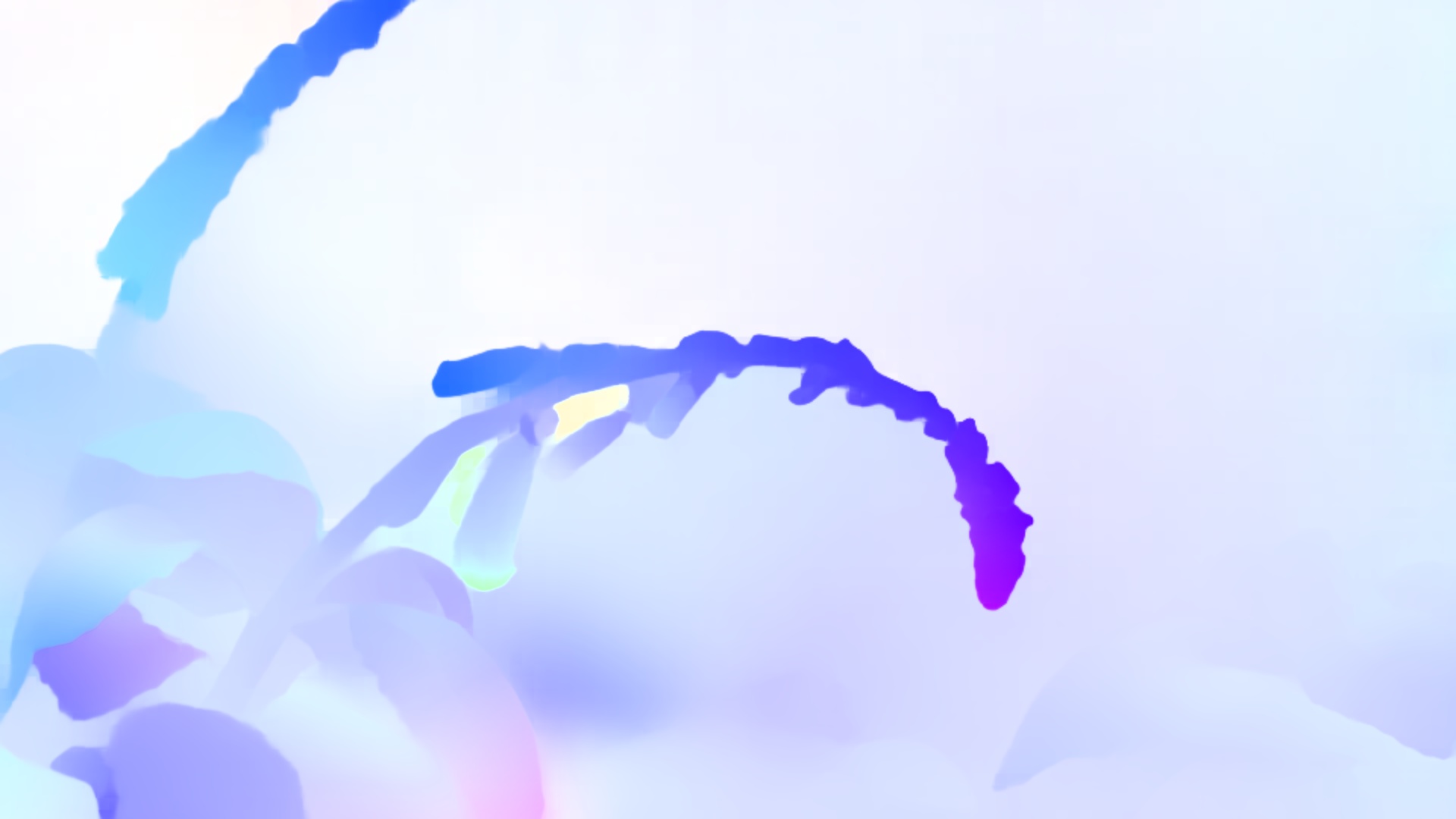} \put (50,48) {\textcolor{purple}{\textbf{\texttt{\small{EPE: 0.457}}}}}
     \end{overpic}
     &
     \begin{overpic}[width=0.315\textwidth, frame]{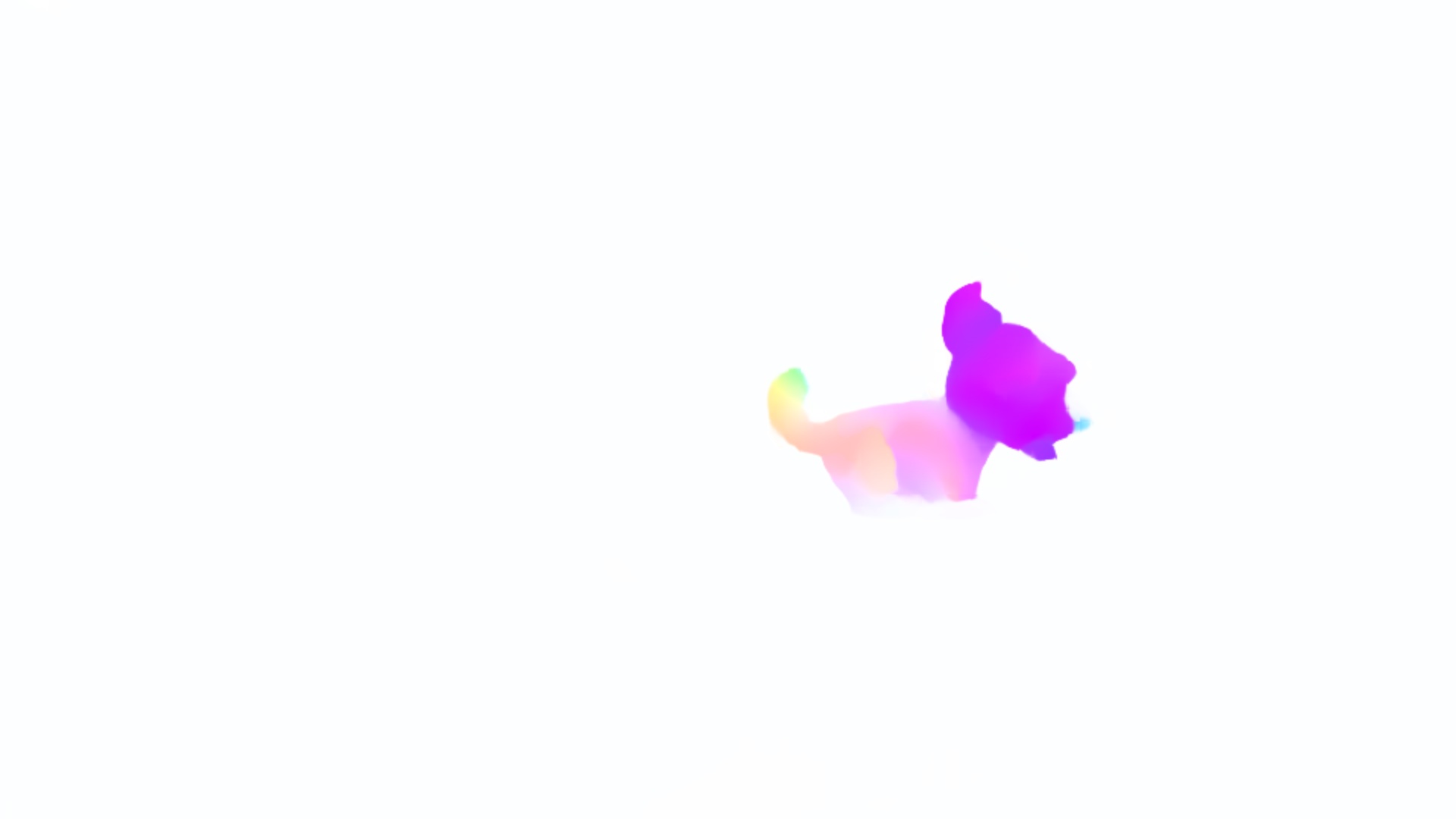} \put (3,48) {\textcolor{purple}{\textbf{\texttt{\small{EPE: 0.353}}}}}
     \end{overpic}
     \\
    \rotatebox[origin=l]{90}{\quad \scriptsize{WAFT~\cite{wang2025waft}}} &
    \begin{overpic}[width=0.315\textwidth, frame]{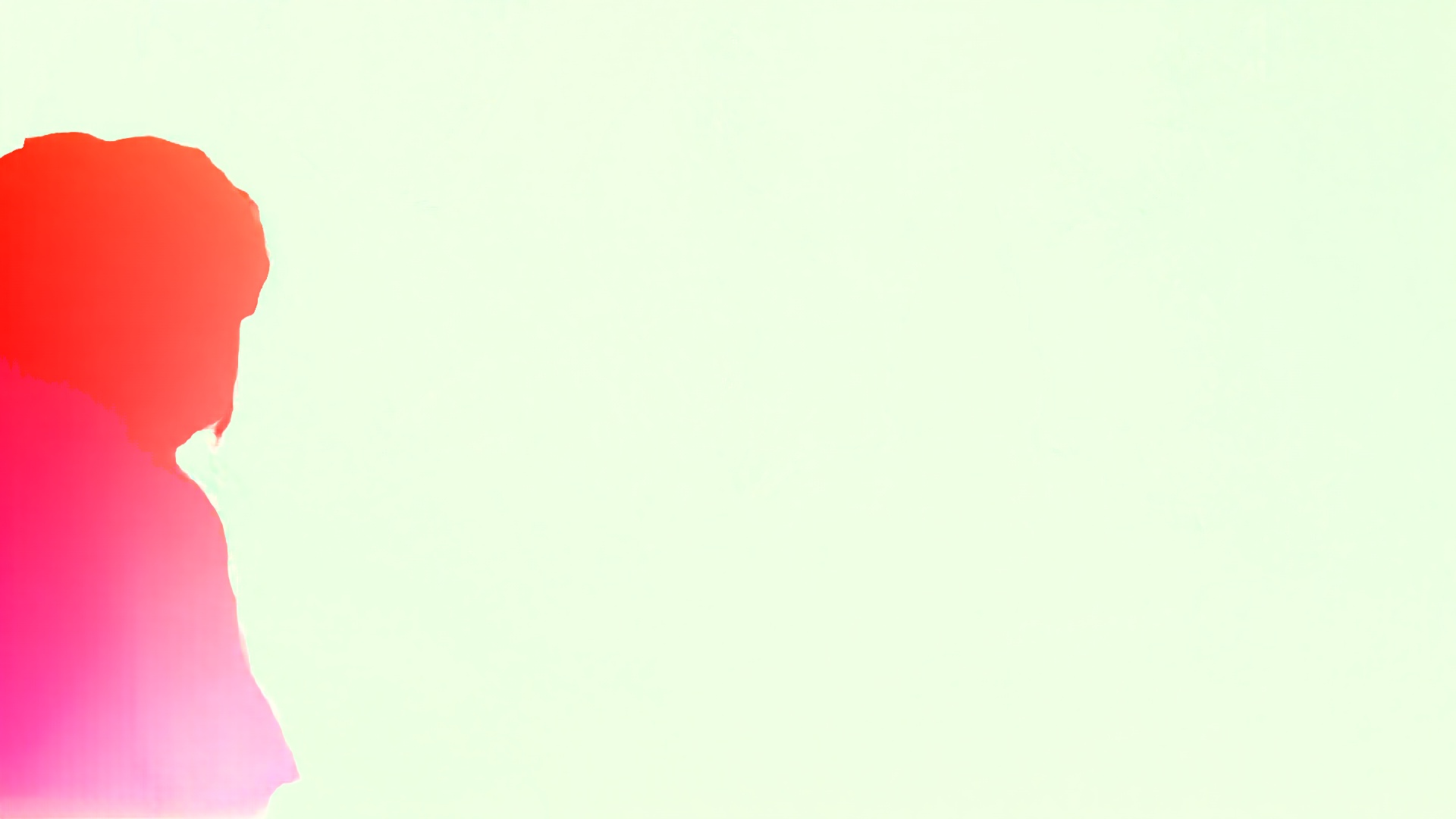} 
    \put (3,48) {\textcolor{purple}{\textbf{\texttt{\small{EPE: 0.344}}}}}
    \end{overpic} &
     \begin{overpic}[width=0.315\textwidth, frame]{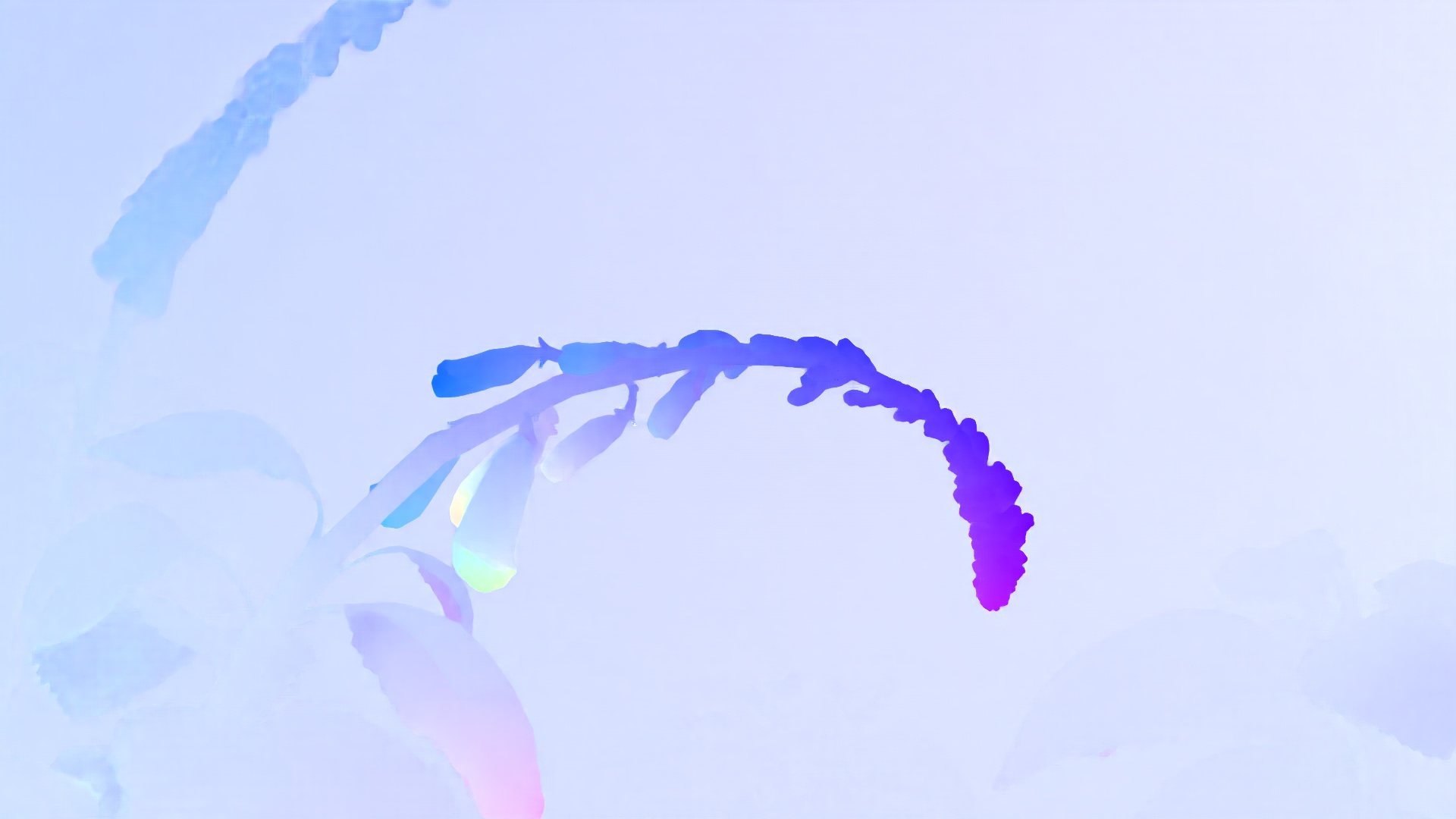} \put (50,48) {\textcolor{purple}{\textbf{\texttt{\small{EPE: 0.334}}}}}\end{overpic}
     &
     \begin{overpic}[width=0.315\textwidth, frame]{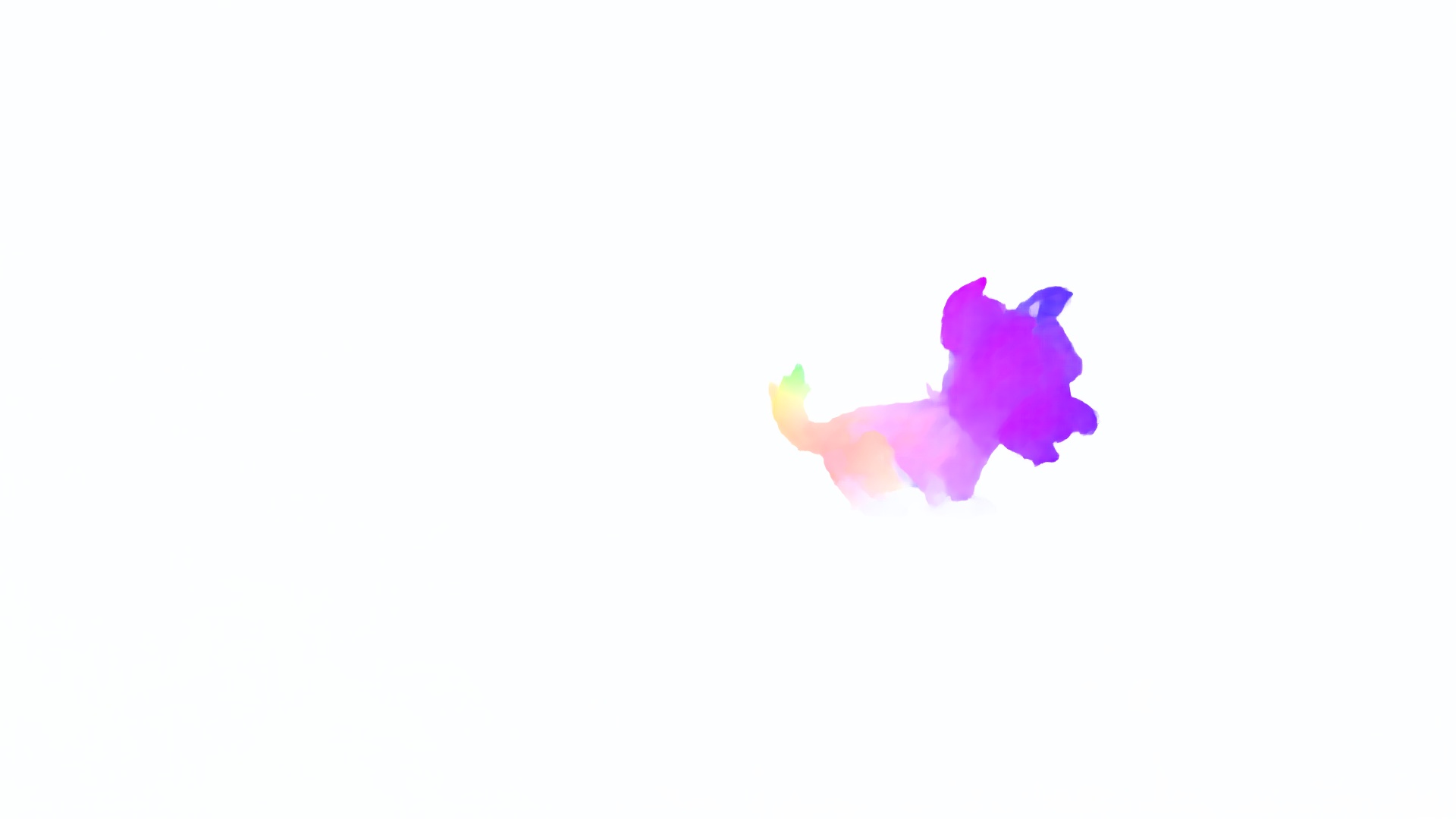} \put (3,48) {\textcolor{purple}{\textbf{\texttt{\small{EPE: 0.312}}}}}\end{overpic}
     \\
    \rotatebox[origin=l]{90}{\scriptsize{\quad FlowSeek~\cite{Poggi_2025_ICCV}}} &
    \begin{overpic}[width=0.315\textwidth, frame]{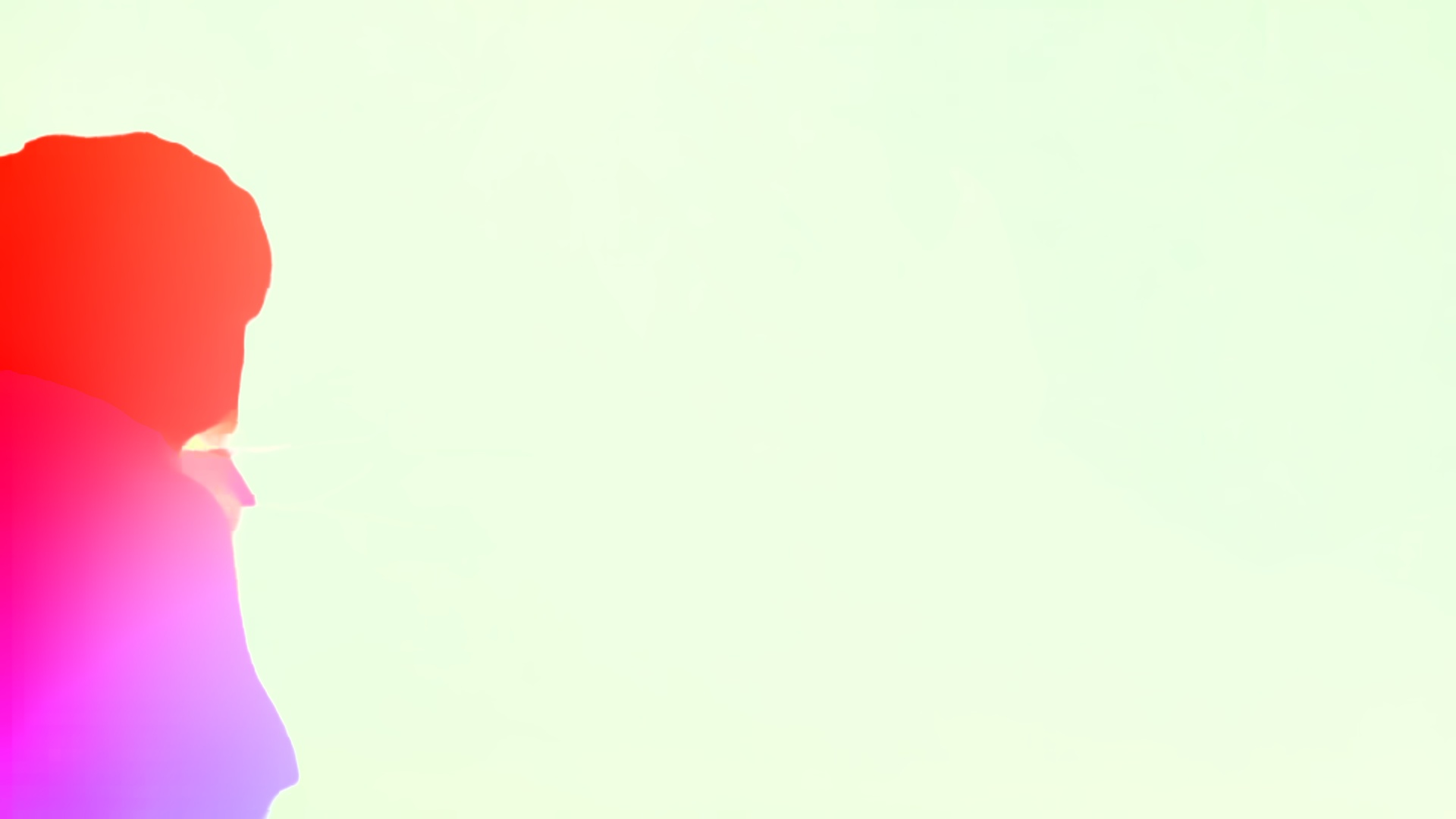} \put (3,48) {\textcolor{purple}{\textbf{\texttt{\small{EPE: 0.286}}}}}
    \end{overpic} &
     \begin{overpic}[width=0.315\textwidth, frame]{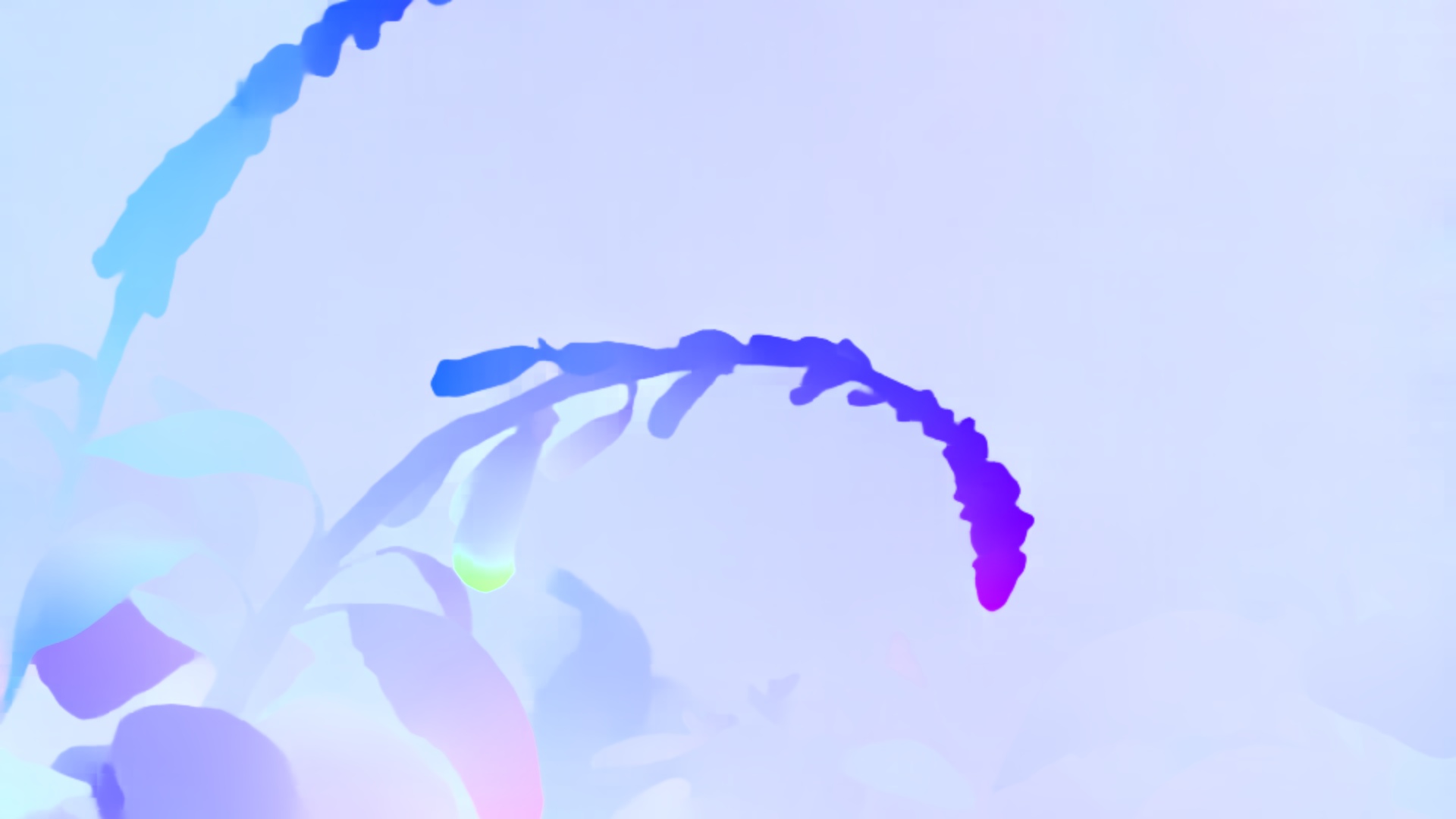} \put (50,48) {\textcolor{purple}{\textbf{\texttt{\small{EPE: 0.296}}}}}
     \end{overpic}
     &
     \begin{overpic}[width=0.315\textwidth, frame]{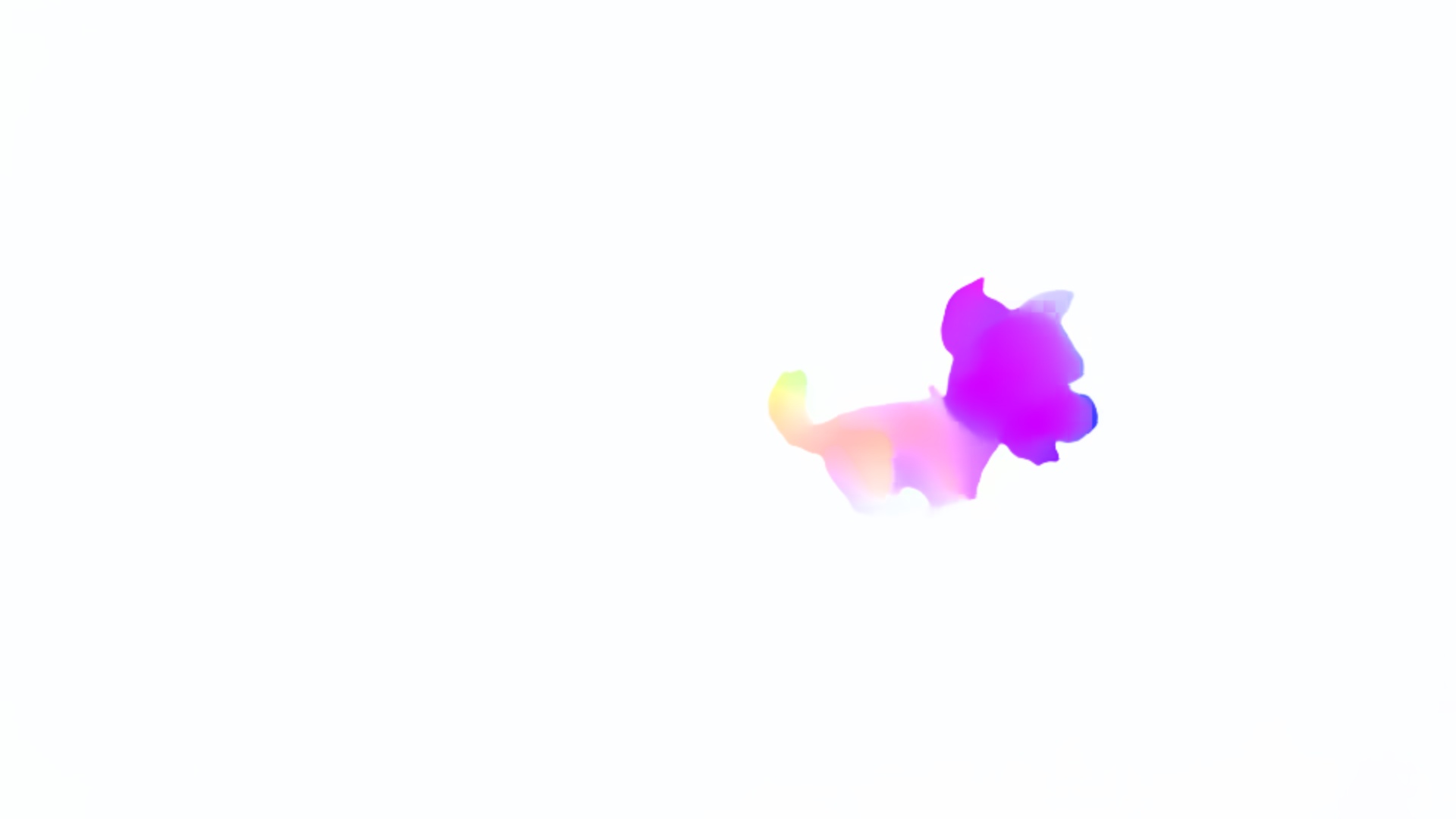} \put (3,48) {\textcolor{purple}{\textbf{\texttt{\small{EPE: 0.268}}}}}
     \end{overpic}
    
    \\
    \rotatebox[origin=l]{90}{\quad \scriptsize{\net{} (XL)}} &
    \begin{overpic}[width=0.315\textwidth, frame]{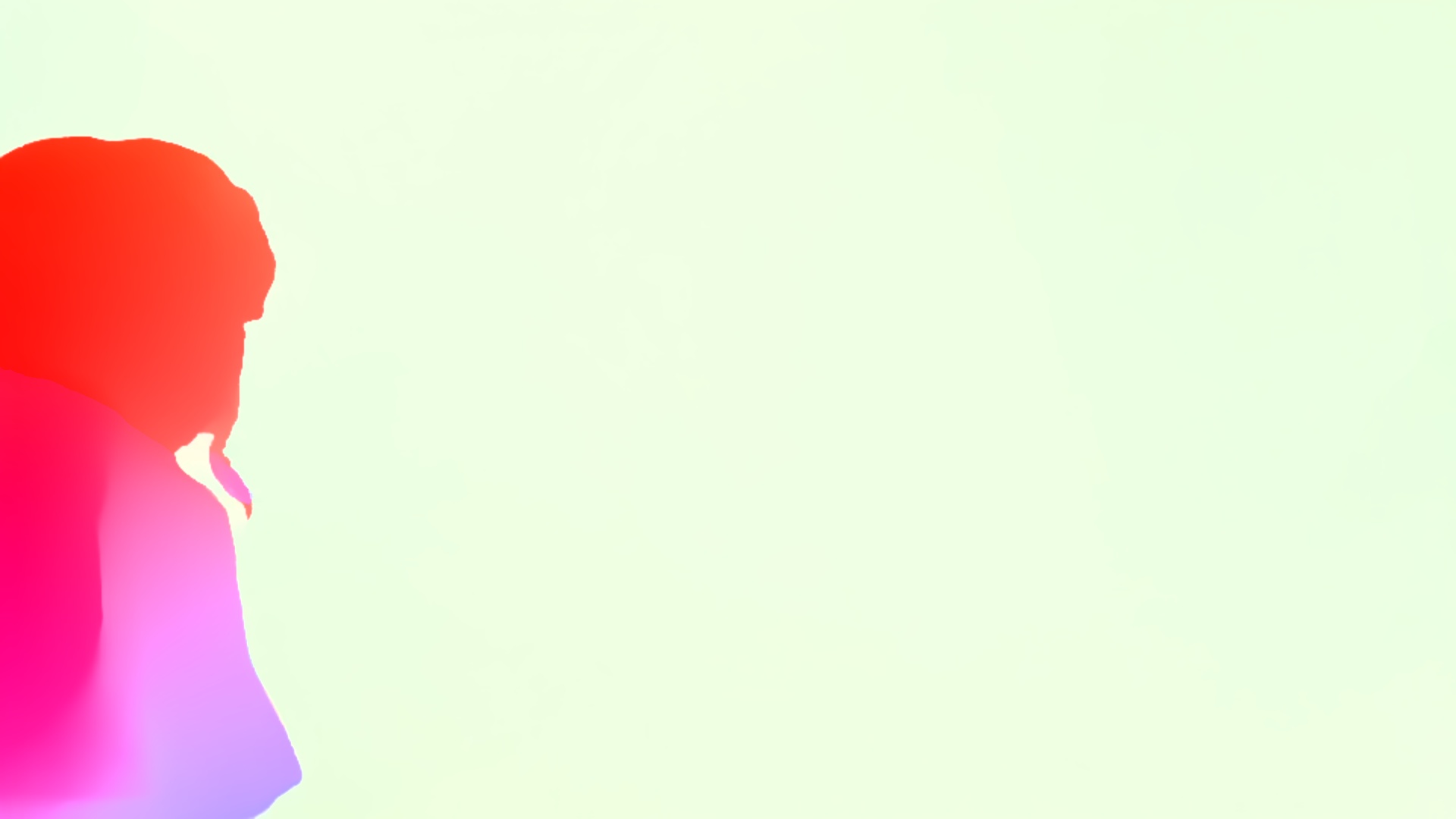} \put (3,48) {\textcolor{purple}{\textbf{\texttt{\small{EPE: 0.218}}}}} \end{overpic} &
     \begin{overpic}[width=0.315\textwidth, frame]{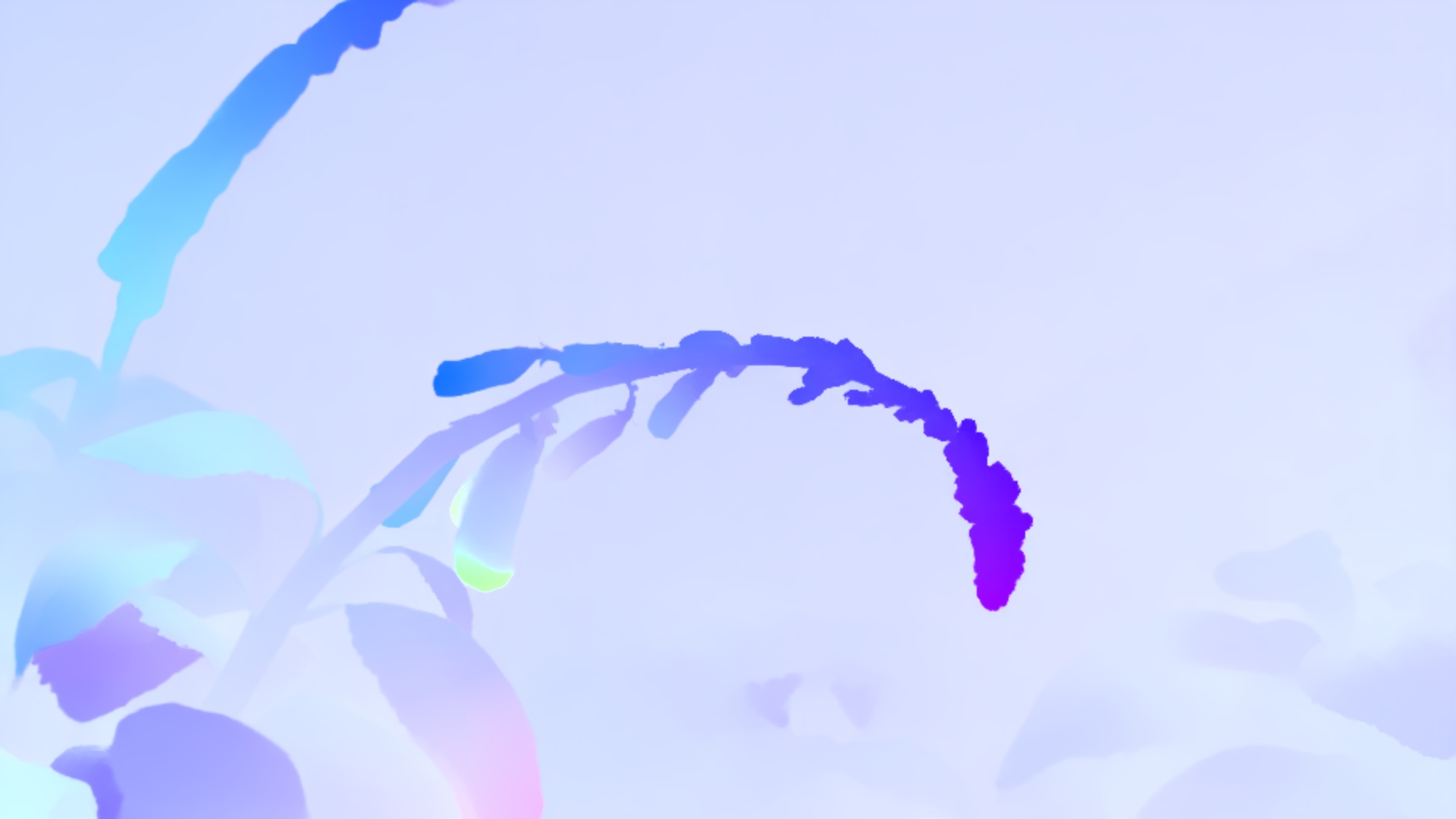}  \put (50,48) {\textcolor{purple}{\textbf{\texttt{\small{EPE: 0.254}}}}} \end{overpic}
     &
     \begin{overpic}[width=0.315\textwidth, frame]{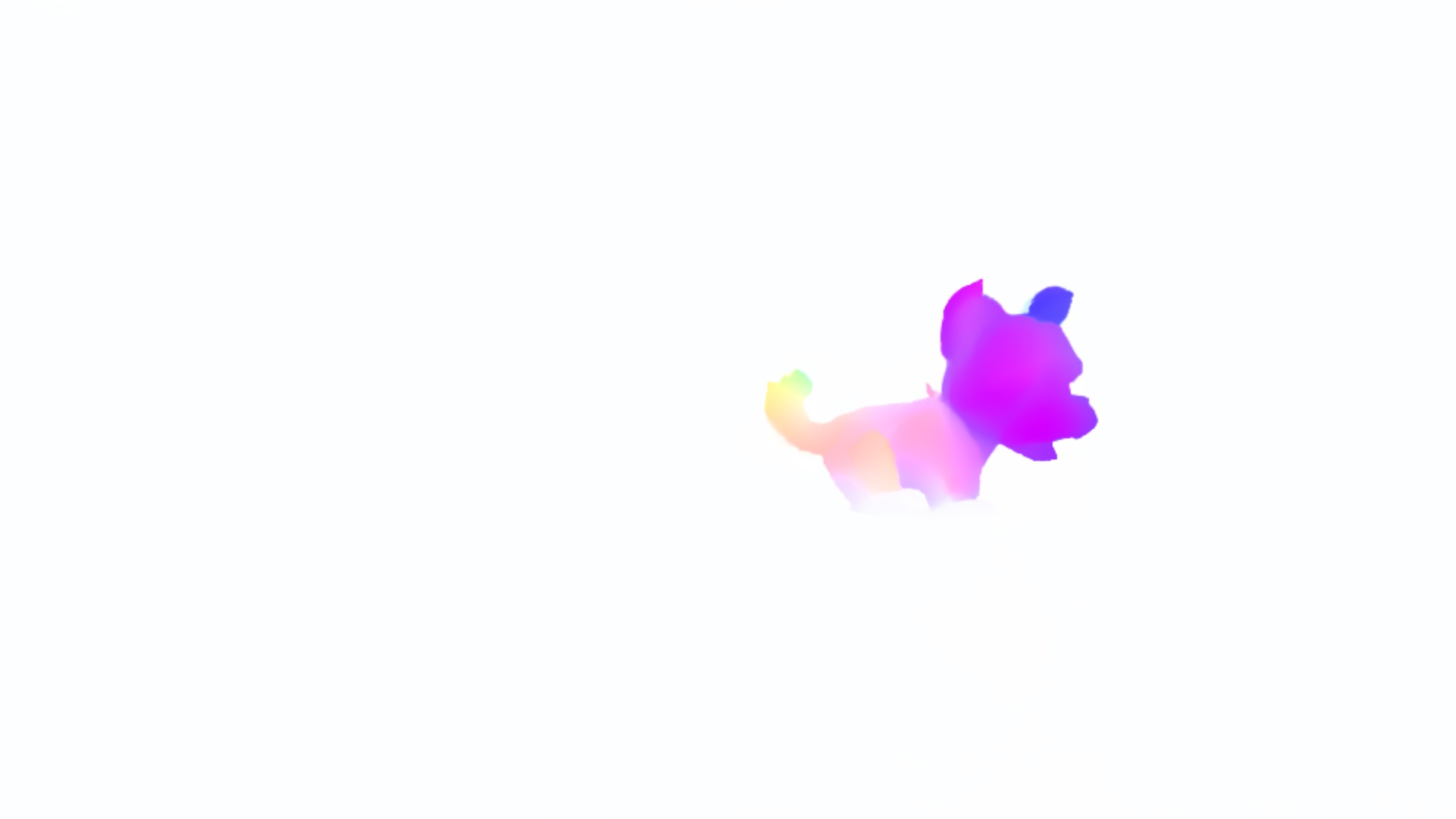}  \put (3,48) {\textcolor{purple}{\textbf{\texttt{\small{EPE: 0.215}}}}} \end{overpic}
    \\
    \rotatebox[origin=l]{90}{ \scriptsize{\quad Ground-Truth}} &
    \begin{overpic}[width=0.315\textwidth, frame]{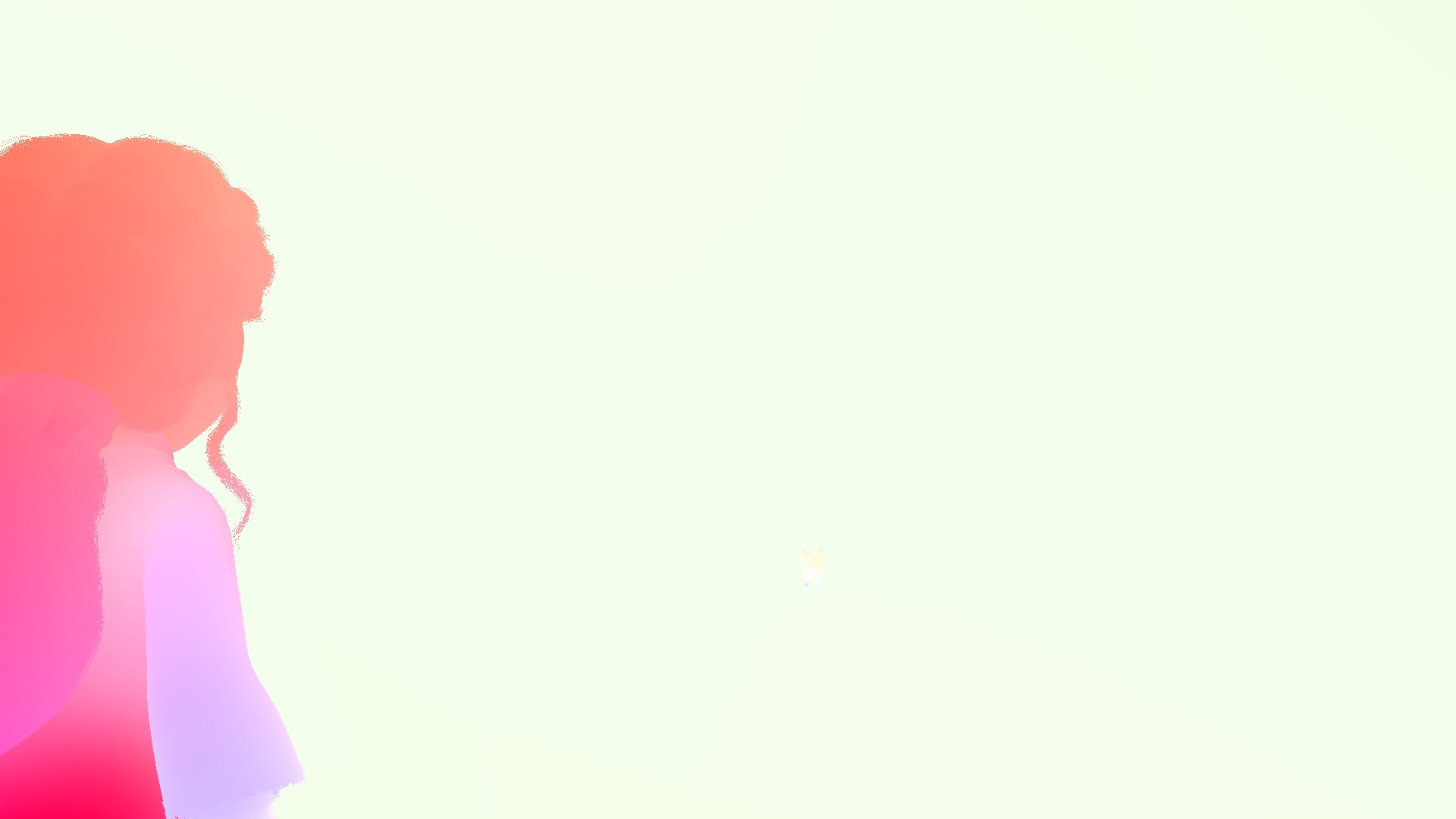} 
    \end{overpic} &
     \begin{overpic}[width=0.315\textwidth, frame]{supp_bmvc/images/spring/04357_gt.jpg}  
     \end{overpic}
     &
     \begin{overpic}[width=0.315\textwidth, frame]{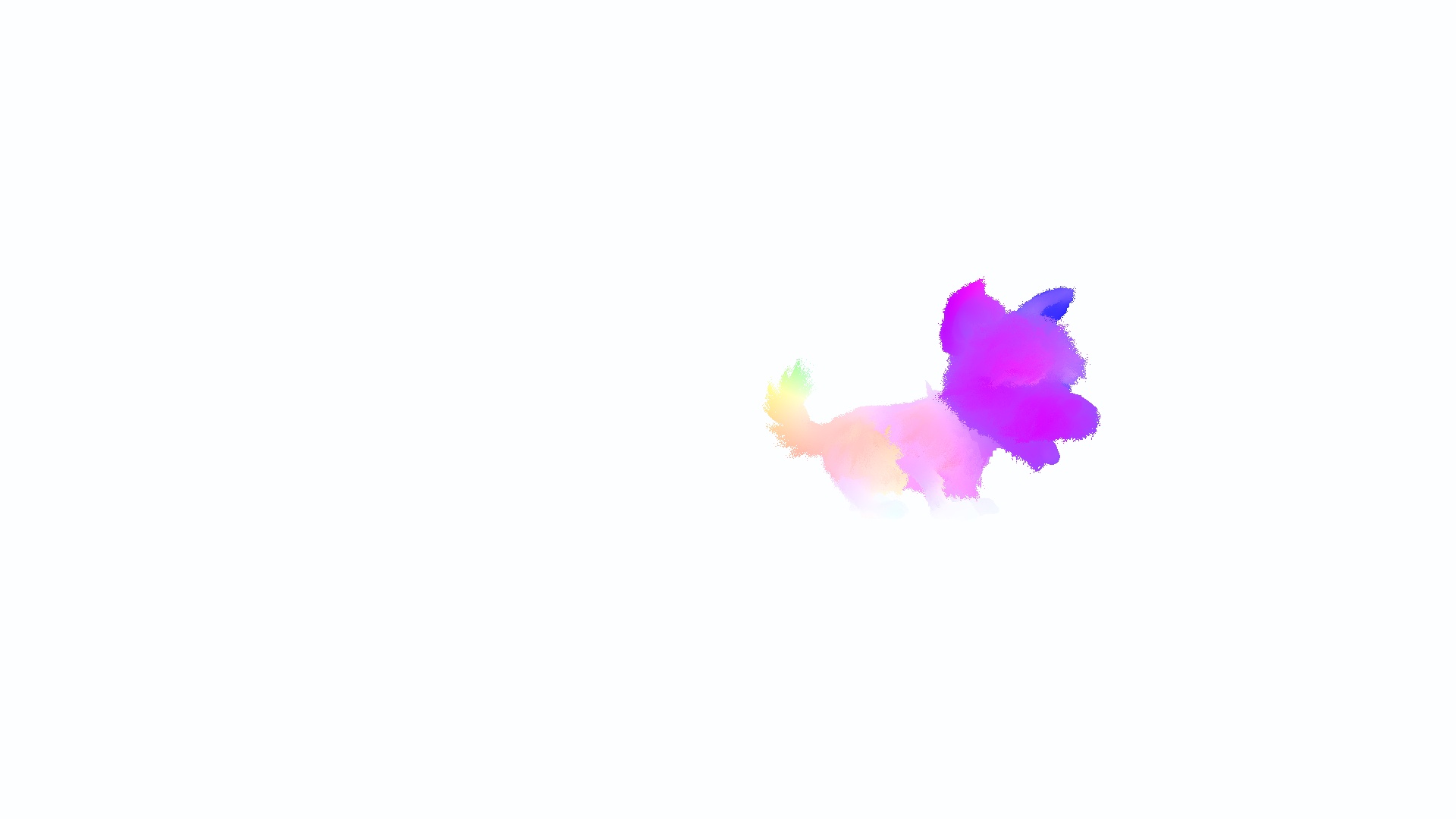} 
     \end{overpic}
    \end{tabular}
    \vspace{0.1cm}
    \caption{\textbf{Qualitative Results of Zero-Shot Generalization on Spring~\cite{mehl2023spring} Training Set.} From top to bottom: first frame, flow by  SEA-RAFT (L)~\cite{wang2024sea}, WAFT-DAv2-a1~\cite{wang2025waft}, FlowSeek (L)~\cite{Poggi_2025_ICCV}, \net{} (XL), and ground-truth.}
    \label{fig:supp_spring_new}
\end{figure*}
\setlength\tabcolsep{12pt}
\begin{table*}[t]
    \centering
    \resizebox{\linewidth}{!}{
    \begin{tabular}{llcccccc}
    \toprule
    \multirow{2}{*}[\multirowcenter]{Extra data} & \multirow{2}{*}[\multirowcenter]{Method} & \multicolumn{2}{c}{Sintel (test)} & \multicolumn{4}{c}{KITTI (test)}\\
    \cmidrule(l{0.5ex}r{0.5ex}){3-4}
    \cmidrule(l{0.5ex}r{0.5ex}){5-8}
    & & Clean$\downarrow$ & Final$\downarrow$ & Fl-all$\downarrow$ & Non-Occ$\downarrow$ & Fl-bg$\downarrow$ & Fl-fg$\downarrow$\\ 
        \midrule
        \multirow{6}{*}{Multi-frame}& SplatFlow~\cite{wang2024splatflow} & 1.12 & 2.07 & 4.61 & 2.96 & 4.26 & 6.34 \\
        & MemFlow \cite{dong2024memflow} & 1.05 & 1.91 & 4.10 & 2.56 & 3.67 & 6.27 \\ 
        & StreamFlow \cite{sun2024streamflow} & 1.04 & 1.87 & 4.24 & - & - &- \\
        & ARFlow \cite{liu2026arflow} & 0.96 & 1.79 & 2.85 & 1.91 & 2.48 & 4.69 \\
        & VideoFlow \cite{shi2023videoflow} & 0.99 & 1.65 & 3.65 & - & - & - \\
        & MEM-FOF \cite{Bargatin_2025_ICCV} & 0.93 & 1.89 & 2.94 & 1.97 & 2.60 & 4.66 \\
        \midrule
        \midrule
        & SpyNet~\cite{ranjan2017optical} & 6.64 & 8.36 &  35.07 & 26.71 & 33.36 & 43.62\\ 
        & FlowNet2~\cite{ilg2017flownet} & 4.16 & 5.74 & 10.41 & 6.94 & 10.75 & 8.75\\
        & GMFlow~\cite{xu2022gmflow} & 1.74 & 2.90 & 9.32 & 3.80 & 9.67 & 7.57\\
        & PWC-Net+~\cite{sun2019models} & 3.45 & 4.60 & 7.72 & 4.91 & 7.69 & 7.88\\
        & RAFT~\cite{teed2020raft} & 1.61 & 2.86 & 5.10 & 3.07 & 4.74 & 6.87\\
        & GMA~\cite{jiang2021learning} & 1.39 & 2.47 & 5.15 & - & - & - \\
        & DIP~\cite{zheng2022dip} &1.44 & 2.83 & 4.21 & 2.43 & 3.86 & 5.96 \\
        & GMFlowNet~\cite{zhao2022global} & 1.39 & 2.65 & 4.79 & 2.75 & 4.39 & 6.87\\
        & CRAFT~\cite{sui2022craft} & 1.45 & 2.42 & 4.79 & 3.02 & 4.58 & 5.85\\
        & FlowFormer~\cite{huang2022flowformer} & 1.16 &  \trd 2.09 & 4.68 & 2.69 & 4.37 & 6.18\\
        & SKFlow~\cite{sun2022skflow} & 1.28 & 2.23 & 4.85 & - & 4.55 & 6.39\\
        & GMFlow+~\cite{xu2023unifying} & 1.03 & 2.37 & 4.49 & 2.40 & 4.27 & 5.60\\
        & EMD-L~\cite{deng2023explicit} & 1.32 & 2.51 & 4.49 & - & 4.16 & 6.15\\
        & RPKNet~\cite{morimitsu2024recurrent} & 1.31 & 2.65 & 4.64 & 2.71 & 4.63 & \fst {4.69}\\ 
        & AnyFlow~\cite{jung2023anyflow} & 1.23 & 2.44 & 4.41 & 2.69 & 4.15 & 5.76 \\
        & SAMFlow~\cite{zhou2024samflow} & \trd 1.00 &  \snd 2.08 & 4.49 & - & - & -\\
        & DPFlow~\cite{morimitsu2025dpflow} &  1.04 & \fst 1.97 & \fst 3.56 & \snd 2.12 & \snd 3.29 & \snd 4.93\\
        & \textBF{\net{} (S)} & 1.09 & 2.44 & 4.49 & 2.53 & 4.30 & 5.42\\
         & \textBF{\net{} (M)} & 1.05 & 2.21 & 4.21 & 2.28 & 4.04 & \trd 5.11\\
        & \textBF{\net{} (L)} & \snd 0.95 & 2.32 & \trd 4.02 & \trd 2.21 & \trd 3.79 &  5.17\\
         & \textBF{\net{} (XL)} & \fst \textBF{0.90} &  2.12 & \snd 3.57 & \fst \textBF{1.88} & \fst 3.25 & 5.18\\
        \midrule
        \midrule
        VIPER~\cite{richter2017playing} & CCMR+~\cite{jahedi2024ccmr} & 1.07 & 2.10 & 3.86 & 2.07 & 3.39 & 6.21\\ 
        MegaDepth~\cite{li2018megadepth} 
         & MatchFlow(G)~\cite{dong2023rethinking} & 1.16 & 2.37 & 4.63 & 2.77 & 4.33 & 6.11\\
        YouTube-VOS~\cite{xu2018youtube} 
         & Flowformer++\cite{shi2023flowformer++} & 1.07 & \snd {1.94} & 4.52 & - & - & -\\ 
        CroCo-Pretrain 
        & CroCoFlow~\cite{weinzaepfel2023croco} & 1.09 & 2.44 & 3.64 & 2.40 & 3.18 & 5.94\\
        DDVM-Pretrain 
        & DDVM~\cite{saxena2023surprising} &1.75 &2.48 & \fst {3.26} & 2.24 & \fst {2.90} & \snd 5.05\\
        AF~\cite{sun2021autoflow} 
        & FlowDiffuser~\cite{luo2024flowdiffuser} & 1.02 &  2.03 & 4.17 & 2.82 & 3.68 & 6.64\\
        \hdashline[2pt/2pt]
        TartanAir~\cite{wang2020tartanair} & SEA-RAFT(L)~\cite{wang2024sea} &1.31 & 2.60 & 4.30 & - & 4.08 & 5.37\\
        TartanAir~\cite{wang2020tartanair} 
        & WAFT-DAv2-a1~\cite{wang2025waft} & 1.09 & 2.34 & \trd 3.42 & \trd 2.04 & - & -\\
        TartanAir~\cite{wang2020tartanair} 
        & WAFT-Twins-a2~\cite{wang2025waft} & 1.02 & 2.39 &  3.53 & 2.12 & 3.18 & 5.28\\
        TartanAir~\cite{wang2020tartanair} 
        & WAFT-DAv2-a2~\cite{wang2025waft} &  0.95 & 2.33 & \snd 3.31 & \snd 2.03 & \snd 2.98 & \fst 4.94\\
        TartanAir~\cite{wang2020tartanair} 
        & WAFT-DINOv3-a2~\cite{wang2025waft} & \trd 0.94 & \trd 2.02 & 3.56 & 2.13 & \trd 3.16 & 5.56\\
        \hdashline
        TartanAir~\cite{wang2020tartanair} 
        & \textBF{\net{} (S)} & 1.01 & 2.45 & 4.50 & 2.59 & 4.31 & 5.45\\
        TartanAir~\cite{wang2020tartanair} 
        & \textBF{\net{} (M)} &  0.95 & 2.10 & 4.41 & 2.35 & 3.88 & 5.43\\      
        TartanAir~\cite{wang2020tartanair} 
        & \textBF{\net{} (L)} & \snd 0.91 & 2.19 & 3.81 & 2.10 & 3.55 &  \trd 5.08\\   
        TartanAir~\cite{wang2020tartanair} 
        & \textBF{\net{} (XL)} & \fst \textBF{0.85} & \fst 1.84 & 3.59 & \fst \textBF{1.95} & 3.29 & 5.09\\
        
    \bottomrule
    \end{tabular}
    }\vspace{0.2cm}
    \caption{\textbf{Results on Sintel~\cite{sintel} and KITTI~\cite{kitti} benchmarks -- impact of extra data.} On the left-most column, we report extra data used at inference in general (i.e., multi-frame) or for training 2-frame methods (e.g., TartanAir). The top-three scores achieved by 2-frame methods trained with and without external data are highlighted separately.}
    \label{tab:supp_big_sintel_kitti_benchmark}       
\end{table*}

\clearpage
\bibliography{egbib}
\end{document}